\documentclass{article}

\usepackage[preprint]{neurips_2026}

\usepackage[utf8]{inputenc} % allow utf-8 input
\usepackage[T1]{fontenc}    % use 8-bit T1 fonts
\usepackage{hyperref}       % hyperlinks
\usepackage{url}            % simple URL typesetting
\usepackage{booktabs}       % professional-quality tables
\usepackage{amsfonts}       % blackboard math symbols
\usepackage{nicefrac}       % compact symbols for 1/2, etc.
\usepackage{microtype}      % microtypography
\usepackage{xcolor}         % colors

\usepackage{graphicx}
\usepackage{graphbox}
\usepackage{subcaption}
\usepackage{bbm}
\usepackage{xurl}
\usepackage{multirow}
\usepackage{color,colortbl}
\usepackage{wrapfig}

\title{Time Series as Language: A Universal Tokenizer for General-Purpose Time Series Foundation Models}

\author{%
Yunhao Zhang\\
Shanghai Jiao Tong University\\
\texttt{zhangyunhao@sjtu.edu.cn}
\And
Ruiying Qi\\
Shanghai Jiao Tong University\\
\texttt{qry-sylvia@sjtu.edu.cn}
\And Jiale Zheng\\
Huawei Noah’s Ark Lab\\
\texttt{zhengjiale2@huawei.com}
\And
Jianfeng Zhang\\
Huawei Noah’s Ark Lab\\
\texttt{zhangjianfeng3@huawei.com}
\And Lujia Pan\\
Huawei Noah’s Ark Lab\\
\texttt{panlujia@huawei.com}
\And
Junchi Yan\thanks{Junchi Yan is the corresponding author.}\\
Shanghai Jiao Tong University\\
\texttt{yanjunchi@sjtu.edu.cn}
}

\usepackage{amsmath}
\usepackage{amsfonts}
\usepackage{bm}

\newcommand{\bx}{\mathbf{x}}

\newcommand{\bz}{\mathbf{z}}

\newcommand{\bH}{\mathbf{H}}
\newcommand{\bh}{\mathbf{h}}
\newcommand{\bW}{\mathbf{W}}

\newcommand{\bp}{\mathbf{p}}

\newcommand{\ba}{\mathbf{a}}
\newcommand{\bd}{\mathbf{d}}

\newcommand{\bbR}{\mathbb{R}}

\makeatletter
\newcommand*\bigcdot{\mathpalette\bigcdot@{.5}}
\newcommand*\bigcdot@[2]{\mathbin{\vcenter{\hbox{\scalebox{#2}{$\m@th#1\bullet$}}}}}
\makeatother

\newcommand{\mycmidrule}[3]{%
  \noalign{\vskip -\aboverulesep}%
  \cmidrule[#1](#2){#3}%
  \noalign{\vskip -\belowrulesep}%
}

\begin{document}

\maketitle

\begin{abstract}
   While Next-Token Prediction (NTP) has unified LLM pretraining, its adaptation to unbounded, continuous time series (TS) remains open. To bridge the gap, we introduce \textbf{UniTok}, a universal tokenizer that transforms TS into discrete tokens, and \textbf{UniTok-FM}, a foundation model pretrained via NTP on these tokens. UniTok-FM is a general-purpose foundation model that supports zero-shot and prompt-boosted forecasting, as well as few-shot generation and classification via \textbf{training-free in-context inference}—a capability not achieved by prior works. Technically, UniTok is a vector-quantized autoencoder incorporating prefix normalization for scale stabilization, a progressive-resolution causal architecture for encoding and decoding, and a structure-preserving reconstruction loss for training. UniTok-FM adopts an off-the-shelf LLM architecture without TS-specific modifications. Instead of pretraining on isolated TS, it performs NTP on context windows formed by multiple series with similar patterns, aiming to capture their shared dynamics. Experiments on forecasting, generation, and classification show that a single unified UniTok-FM consistently outperforms statistical and supervised baselines, achieves competitive performance with task-specific foundation models, and uniquely enables training-free in-context inference across tasks.
\end{abstract}

\section{Introduction} \label{sec:intro}

In recent years, next-token prediction (NTP) has unified the pretraining of large language models (LLMs). Although time series (TS) are also sequential data, the pretraining paradigm of time series foundation models (TSFMs) remains highly fragmented. Existing TSFMs use different pretraining tasks, ranging from next-patch prediction~\cite{liu2024timer,cohen2025time} to mask-and-reconstruction~\cite{wang2025output,chen2025visionts} and fixed-horizon prediction~\cite{xiaoming2025time,liu2025sundial}. The backbone architectures are also diverse: ranging from xLSTMs~\cite{auer2025tirex} to Transformers equipped with TS-specific positional embeddings~\cite{woo2024unified}, attention mechanisms~\cite{sun2025xihe}, etc. Moreover, most TSFMs are narrowly tailored to forecasting, falling short of the multi-task generality that LLMs achieve.

A key challenge in extending NTP to TS is the unbounded, continuous nature of TS. Modeling complex distributions in continuous space is difficult: conventional regression-based objectives typically rely on rigid parametric assumptions~\cite{zhang2026mmpd}, whereas generative alternatives like diffusion introduce significant architectural complexity~\cite{li2024autoregressive}. In contrast, discretization enables flexible distribution modeling via a simple cross-entropy objective. And a discrete representation facilitates modeling multiple series within a unified context window, enabling in-context learning and generalizing the learned model beyond forecasting to broader tasks such as generation and classification.

While discretization offers a clear path, developing a tokenizer for TS is more complex than for images because of greater variability in series length and numerical scale. Consequently, most prior TS tokenizers are task-specific~\cite{shi2025kronos}, applicable to narrow datasets~\cite{lee2023vector,talukder2024totem}, or impose strict constraints on series length~\cite{tao2025values}, remaining inadequate for general-purpose TSFMs across domains and tasks. The work most relevant to ours is Chronos\footnote{Chronos should be distinguished from Chronos-Bolt~\cite{Ansari2024bolt} and Chronos-2~\cite{ansari2025chronos2}, which do not use discretization and NTP.}~\cite{ansari2024chronos}, which discretizes scaled TS via point-wise uniform binning and pretrains a TSFM using NTP. However, this simple binning strategy fails to capture rich temporal dependencies, and the model trained on isolated series is restricted to forecasting. As such, NTP's generalization potential has not been fully exploited.

To fill the gap, we propose a universal TS tokenizer, \textbf{UniTok}, and a general-purpose foundation model, \textbf{UniTok-FM}, pretrained via NTP using UniTok. Both are trained on large-scale datasets to enable cross-domain generalization. Beyond zero-shot forecasting supported by prior TSFMs, UniTok-FM unlocks three additional capabilities: 1) prompt-boosted forecasting, where TS with similar dynamics serve as prompts to guide prediction; 2) few-shot generation, producing high-fidelity samples from only a handful of example TS; 3) few-shot classification, classifying using limited labeled examples. All capabilities are realized via \textbf{training-free in-context inference}, without fine-tuning task-specific heads. To our best knowledge, no prior TSFM supports generation or classification in this manner.

Technically, UniTok builds on the VQ-VAE framework~\cite{van2017neural}, a commonly used approach for image tokenizers. To adapt it to TS with variable lengths and unbounded values, some key modifications are introduced: 1) incremental tokenization property that aligns tokenization with the NTP paradigm; 2) prefix normalization that stabilizes scale while preserving incremental property; 3) progressive-resolution causal autoencoder that assigns token resolution based on receptive fields; 4) structure-preserving reconstruction loss to faithfully capture temporal structures.

Built on UniTok, UniTok-FM is pretrained via NTP using an off-the-shelf LLM architecture, without TS-specific modification. Instead of pretraining on isolated series, UniTok-FM performs NTP on context windows comprising multiple series with similar patterns. During pretraining, similarity is enforced by extracting segments from the same long series; for inference, it generalizes beyond this construction to align with each task's requirements. UniTok-FM supports general-purpose, training-free inference: zero-shot/prompt-boosted forecasting and few-shot generation are performed through autoregressive (AR) token generation under different prompt contexts, while few-shot classification is achieved by evaluating the conditional likelihood of the query series. \textbf{The highlights are:}

1. We propose UniTok, a universal TS tokenizer that operates across domains and tasks, transforming continuous TS into discrete tokens suitable for NTP pretraining.

2. We pretrain UniTok-FM, a general-purpose TSFM via in-context NTP, supporting training-free zero-shot and prompt-boosted forecasting, as well as few-shot generation and classification, which prior TSFMs do not support.

3. Although UniTok-FM does not surpass task-specific SOTA models in every setting, a single unified model, using training-free in-context inference with only a handful of prompt examples, matches: 1) forecasting performance of forecasting-specific TSFM \textbf{Chronos-Bolt}~\cite{Ansari2024bolt}; 2) generation quality of \textbf{Diffusion-TS}~\cite{yuan2024diffusionts}, despite it is trained on one thousand samples; 3) few-shot classification accuracy of \textbf{MOMENT}~\cite{goswami2024moment}, despite it relies on a downstream classifier.

\section{Related Works}

\textbf{Time Series Foundation Models}
TSFMs are pretrained on large-scale datasets and generalize to new scenarios in a zero-shot setting. Unlike the unified NTP paradigm in LLMs, TSFM pretraining remains fragmented. Recent works explore pretraining tasks such as next-patch prediction~\cite{liu2024timer}, mask-and-reconstruction~\cite{wang2025output} and fixed-horizon forecasting. The latter spans point regression~\cite{xiaoming2025time}, mixture distribution modeling~\cite{cohen2025time}, quantile regression~\cite{ansari2025chronos2} and flow matching~\cite{liu2025sundial}. Backbone designs are also diverse: while most adopt Transformer variants, alternatives such as xLSTM are also competitive~\cite{auer2025tirex}. Many TSFMs further introduce task-specific modifications, including customized positional embeddings~\cite{woo2024unified}, attention mechanisms~\cite{sun2025xihe}, or horizon-specific prediction heads~\cite{liu2025sundial}. Moreover, most TSFMs support only zero-shot forecasting, whereas general-purpose TSFMs serve as feature extractors and require training task-specific models on these features~\cite{gao2024units,goswami2024moment}.
 
The work most relevant to ours is Chronos~\cite{ansari2024chronos}. It normalizes series with mean scaling and discretizes each point using uniform binning with fixed bin edges to enable NTP with a T5~\cite{raffel2020exploring} backbone. However, its point-wise binning limits the modeling of rich temporal structure, and pretraining on isolated series restricts Chronos to only zero-shot forecasting.

\textbf{Image Tokenizers} AR image generation relies on discrete tokenizers. Efforts have evolved from pixel-level discretization~\cite{van2016conditional} to vector quantized variational autoencoders (VQ-VAE)~\cite{van2017neural} and its extensions~\cite{razavi2019generating,esser2021taming,lee2022autoregressive}. To address the instability of VQ, lookup-free alternatives have been proposed~\cite{yu2023language}. In particular, Finite Scalar Quantization (FSQ)~\cite{mentzer2024finite} stabilizes training via simple low-dimensional projection and scalar quantization; we therefore adopt FSQ in UniTok. Others also explore multi-scale generation~\cite{tian2024visual} and efficient compression~\cite{yu2024image}. Readers can refer to \cite{jia2025principles} for an overview. Applying these techniques to TS is non-trivial due to the variable lengths and unbounded value ranges, unlike fixed-size images (e.g., $512\times512$) with bounded pixel values (e.g., $0\sim255$).

% Image tokenizers are used for discrete image representation which was established by VQ-VAE \cite{van2017neural}. VQ-VAE quantizes continuous feature maps by mapping them to the nearest entries in a learnable codebook, and then it was further scaled by VQ-VAE-2 \cite{razavi2019generating} using hierarchical latent codes. VQGAN \cite{esser2021taming} significantly enhanced generation fidelity by introducing adversarial and perceptual losses, combining Transformer and CNN architectures. Subsequent works focused on improving efficiency and generative qualities : Residual Quantization (RQ-VAE and RQ-Transformer)\cite{lee2022autoregressive} uses a fixed size codebook to refine coarse codes recursively, while MagViT-v2 \cite{yu2023language} and FSQ \cite{mentzer2024finite} simplified the process by removing codebook lookups in favor of lookup-free quantization. Recently, VAR \cite{tian2024visual} redefined the autoregressive paradigm via introducing "next-scale prediction" to generate tokens from coarse to fine resolutions, diverging from the standard raster scan, while TiTok \cite{yu2024image} achieved compressing images into compact 1D latent sequences, yielding substantially more efficient and effective
% representations. However, all the tokenizers mentioned above are tailored for the image or video tasks, which means that directly applying them to sequential data is often suboptimal, necessitating significant architectural adaptations to effectively capture continuous temporal dynamics.

\begin{figure*}[tb!] 
\centering
    \subcaptionbox{}{\includegraphics[align=t,width=0.24\textwidth,height=0.28\textwidth]{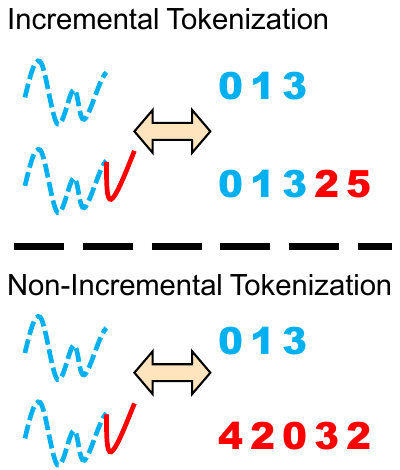}\vspace{-5pt}}
    \subcaptionbox{}{\includegraphics[width=0.68\textwidth,height=0.28\textwidth]{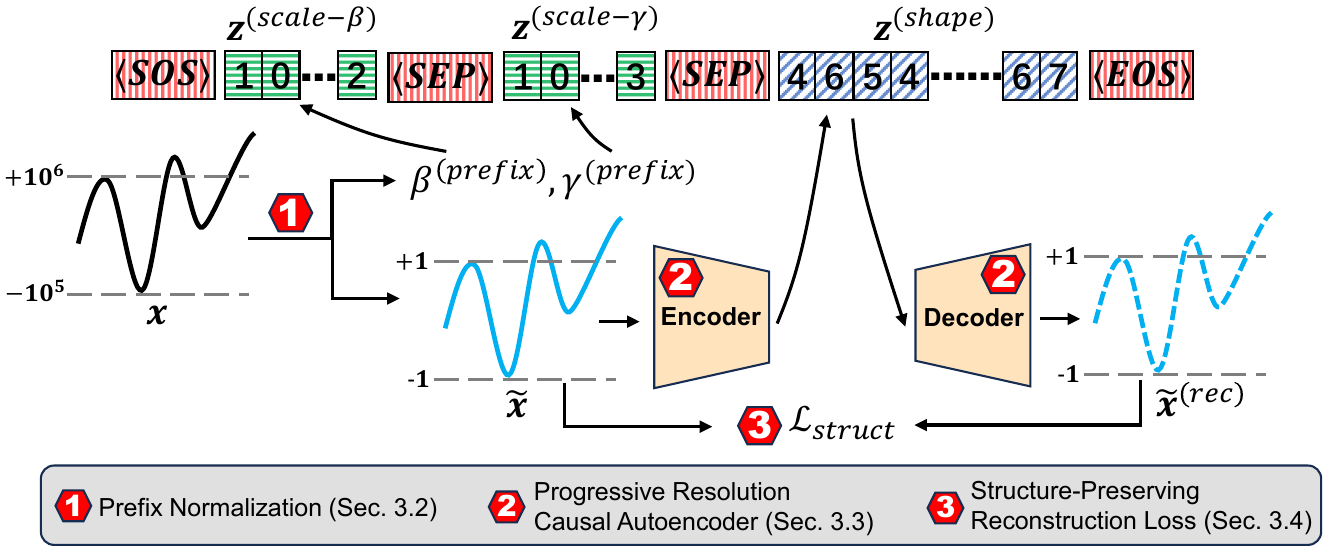}\vspace{-5pt}} 
    \vspace{-8pt}
    \caption{(a) \textbf{Incremental vs. non-incremental tokenization.} Incremental tokenization makes prefix tokens independent of future observations, so appending data extends the token sequence, aligning with the NTP paradigm. Otherwise, incompatible tokens for a prefix and its extension limit generalization from long to short series. (b) \textbf{Overview of UniTok.} The raw TS is decomposed into scale statistics and a normalized series via prefix normalization (Sec.~\ref{sec:prefix_norm}). Scale statistics are discretized in hex of Float32, while normalized series is encoded by a progressive-resolution causal autoencoder (Sec.~\ref{sec:auto_encoder}), trained with a structure-preserving reconstruction loss (Sec.~\ref{sec:struct_loss}).}
    \label{fig:framework}
    \vspace{-15pt}
\end{figure*}

\textbf{Time Series Tokenizers} Recent efforts adapt VQ for TS generation~\cite{lee2023vector}, forecasting~\cite{feng2025hdt}, and classification~\cite{wen2024abstracted}. Some align TS with texts via reprogramming~\cite{jin2023time} or VQ-VAE~\cite{tao2025values}. Domain-specific variants, such as K-line tokenizers for finance~\cite{shi2025kronos}, also exist. Nevertheless, most existing TS tokenizers are task- or dataset-specific and lack reusability. Although \cite{talukder2024totem} explores extending VQ-VAE beyond a single dataset, it remains limited to a small number of datasets, falling short of a universal tokenizer for general-purpose TSFMs.

\section{UniTok: A Universal Time Series Tokenizer} \label{sec:unitok}

As illustrated in Fig.~\ref{fig:framework}(b), UniTok converts a TS into a sequence of discrete tokens while satisfying the incremental tokenization property required for NTP (Sec.~\ref{sec:incremental}). Specifically, the input series is decomposed into scale statistics and a normalized series via prefix normalization (Sec.~\ref{sec:prefix_norm}). The scale statistics are tokenized using a hexadecimal representation, while the normalized series is encoded by the progressive-resolution causal autoencoder (Sec.~\ref{sec:auto_encoder}), trained with the structure-preserving reconstruction loss (Sec.~\ref{sec:struct_loss}). The tokenized sequence takes the form:
\begin{equation}
\langle\text{\small SOS}\rangle
\bz^{(scale-\beta)}
\langle\text{\small SEP}\rangle
\bz^{(scale-\gamma)}
\langle\text{\small SEP}\rangle
\bz^{(shape)}
\langle\text{\small EOS}\rangle
\end{equation}
where $\langle\text{SOS}\rangle/\langle\text{EOS}\rangle$ denotes start/end of a series. $\langle\text{SEP}\rangle$ is for seperation. $\bz^{(scale-\beta)}, \mathbf{z}^{(scale-\gamma)}$ are tokenized scale statistics. $\mathbf{z}^{(shape)}$ corresponds to normalized series, capturing intrinsic series shape.

\subsection{Incremental Tokenization Property} \label{sec:incremental}
Given a TS\footnote{We focus on univariate time series and adopt the widely used channel-independent technique for multivariate data~\cite{nie2023time}.} $\bx = [x_1, \dots, x_T], x_t \in \bbR$, our goal is to transform it into discrete tokens $\bz = [z_1, \dots, z_L], z_i \in \mathcal{C}, \mathcal{C}=\{1, \dots, C\}$ such that $\bz$ is suitable for NTP. Unlike fixed-size images, UniTok handles TS of variable length. We require it to satisfy the incremental tokenization property:
\begin{equation} \label{eq:incremental}
    \text{Enc}(\bx_{\le t}) = \text{Enc}(\bx)_{\le \phi(t)} \quad \text{Dec}(\bz_{\le l}) = \text{Dec}(\bz)_{\le \phi^{-1}(l)}  
\end{equation}
where $\text{Enc}(\bigcdot), \text{Dec}(\bigcdot)$ denote encoding and decoding. $\phi(\bigcdot)$ maps a TS length to the corresponding token sequence length, and $\phi^{-1}(\bigcdot)$ denotes the inverse mapping. As illustrated in Fig.~\ref{fig:framework}(a), this property states that encoding or decoding any prefix is independent of the remaining part. This ensures that a TSFM trained on length $T$ seamlessly generalizes to any shorter series.

\subsection{Prefix Normalization} \label{sec:prefix_norm}

Unlike images with bounded pixel values (i.e., $0\sim255$), TS exhibit widely varying scales across domains, making normalization essential. Given $\bx$, conventional normalization applies $\widetilde{\bx}=(\bx - \beta)/\gamma$, where $\beta, \gamma$ are extracted scale statistics (e.g., mean–std or min–max) from $\bx$. The normalized $\widetilde{\bx}$ has a more stable scale and is used as the network input. However, conventional normalization violates the incremental property: statistics extracted from a prefix differ from those of the whole series, such that $\text{Norm}(\bx_{\le t}) \ne \text{Norm}(\bx)_{\le t}$. We therefore propose prefix normalization that computes statistics from a fixed-length prefix. Assuming all the TS we process are longer than $P$, we perform:
\begin{equation}
    \widetilde{\bx} = \text{Prefix-Norm}(\bx) = \frac{\bx - \beta^{(prefix)}}{\gamma^{(prefix)}} \quad
    (\beta^{(prefix)}, \gamma^{(prefix)}) = f_{scale}(\bx_{\le P})
\end{equation}
where $f_{scale}(\bx_{\le P})$ extracts statistics from the length-$P$ prefix. In this work, we adopt min–max normalization, although other choices are possible. Prefix normalization preserves the incremental property as $\text{Prefix-Norm}(\bx_{\le t}) = \text{Prefix-Norm}(\bx)_{\le t}, \forall t \ge P$. To relax the requirement that all series must be longer than $P$, we introduce two prefix lengths $P_1 < P_2$ and perform: 
{\small
\begin{equation}
(\beta^{(prefix)}, \gamma^{(prefix)}) =
\begin{cases}
 f_{scale}(\bx) & T<P_1, \\
 f_{scale}(\bx_{\le P_1}) & P1\le T < P_2, \\
 f_{scale}(\bx_{\le P_2}) & T \ge P_2, \\
\end{cases}
\end{equation}
}
We set $P_1=8$ and $P_2=128$, ensuring that the incremental property holds within each range. For extremely short series ($T < 8$), we fall back to conventional normalization.

\textbf{Scale Statistics Tokenization.} Prior works typically discard scale statistics and operate only on the normalized series, which is sufficient for forecasting. However, for tasks such as classification, the absolute scale often carries semantic information and should be preserved. For a tokenizer that works across various domains, the range of $\beta^{(prefix)},\gamma^{(prefix)}$ can be as broad as a Float32 number ($-3.4\times10^{38}\sim+3.4\times10^{38}$). This range is far too vast for traditional neural network embeddings. To address this, we access their 32-bit computer storage representation and group every 4 bits into one token. This encodes each of $\beta^{(prefix)},\gamma^{(prefix)}$ into 8 hexadecimal tokens in the range $0 \sim F$. For example, $3.14159$ is tokenized as $[4,0,4,9,0,F,D,0]$. This process is lossless and reversible. Since only two statistics are stored per series, the overhead is affordable.

\subsection{Progressive-Resolution Causal Autoencoder} \label{sec:auto_encoder}

We adopt an autoencoder with FSQ to tokenize the normalized series $\widetilde{\bx}$. The encoder follows a standard image tokenizer design, consisting of $S$ blocks with convolutional and self-attention layers, each followed by a downsampling operation that halves the resolution. This yields intermediate representations $\bH^{s} \in \bbR^{\frac{T}{2^s} \times d_{model}}$ at the $s$-th block. FSQ~\cite{mentzer2024finite} is used to quantize final block representations $\bH^{S}$ into $\frac{T}{2^S}$ tokens. A symmetric decoder with mirrored upsampling layers reconstructs $\widetilde{\bx}$. While effective for images, some modifications are required for TS.

\textbf{Causal Structure} To preserve the incremental property, we replace all non-causal components with causal counterparts so that each token depends only on the observed prefix. Specifically, we adopt causal convolutions and self-attention with Layer Normalization.

\begin{wrapfigure}{l}{0.5\textwidth}
    \centering
    {\includegraphics[width=0.99\linewidth, height=0.55\linewidth]{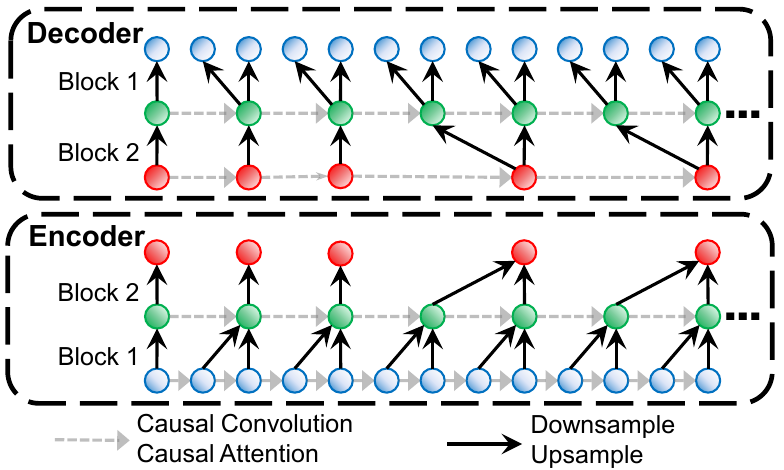}} 
    \vspace{-15pt}
     \caption{\textbf{Progressive-Resolution Causal Autoencoder.} Each block applies causal convolution and attention, allowing each latent vector to attend only to the past. At block $s$, the first $2^s-1$ vectors are preserved, while the remaining are downsampled/upsampled, yielding a progressive-resolution architecture in which earlier tokens with smaller receptive fields receive finer representations.}
    \label{fig:autoencoder}
    \vspace{-10pt}
\end{wrapfigure}

\textbf{Progressive-Resolution Autoencoder} Causal structure induces information asymmetry. Specifically,
\begin{equation}
    \bz^{(shape)}_l = \text{Enc}\left(\widetilde{\bx}_{\le l*2^S}\right)_l 
\end{equation}

This implies that earlier tokens (small $l$) are computed from limited context $\widetilde{\bx}_{\le l*2^S}$, creating an information bottleneck. In the extreme, the first token is derived solely from the first patch of length $2^S$, correspondingly, this patch must be reconstructed from a single token, limiting the reconstruction to only $C$ candidate patches. To mitigate this asymmetry, a progressive-resolution architecture that allocates higher resolution to earlier positions is devised, illustrated in Fig.~\ref{fig:autoencoder}. Specifically, we replace the uniform downsampling with progressive downsampling:
\begin{equation}
    \bH^{s} = [\widetilde{\bH}^{s}_{\le 2^s - 1}, \text{Downsample}\left(\widetilde{\bH}^{s}_{\ge 2^s}\right)]
\end{equation}
where $[\bigcdot,\bigcdot]$ denotes concatenation, $\widetilde{\bH}^{s}$ denotes representation before downsampling at block $s$. The first $2^s-1$ vectors are kept, while downsampling is applied to the remaining suffix. As a result, resolution decreases progressively along the sequence: early tokens represent finer spans, while later tokens aggregate larger spans, with tail tokens covering $2^S$ time points, matching the standard downsampling rate. A symmetric progressive upsampling is applied to the decoder.

\subsection{Structure-Preserving Reconstruction Loss} \label{sec:struct_loss}

Obtaining ground truth and reconstructed normalized series $\widetilde{\bx}, \widetilde{\bx}^{(rec)}$, a reconstruction loss is required for autoencoder training. Image tokenizers typically employ a composite loss with three components~\cite{esser2021taming,tian2024visual}: 1) an $L1$ loss for point-wise fidelity; 2) a perceptual loss to preserve semantic patterns; and 3) an adversarial loss for global distribution alignment. We construct our loss based on this.

\textbf{Surrogate Perceptual Loss} Perceptual loss typically aligns ground-truth and reconstruction in the latent space of a feature extractor pretrained on large-scale datasets~\cite{zhang2018unreasonable}. Unlike images, TS lacks a widely adopted pretrained network. To address this, we reuse the discriminator from the adversarial loss as the feature extractor to construct a surrogate perceptual loss. This complements the adversarial loss: while the adversarial loss enforces global distributional consistency, this term encourages each reconstructed series to match its corresponding ground truth in the discriminator's latent space.
 
\textbf{High-Frequency Wavelet Loss} $L1$ loss tends to produce overly smooth reconstructions~\cite{ledig2017photo}. To explicitly preserve fine-grained structures, we introduce a loss that aligns high-frequency coefficients of the Discrete Wavelet Transformation (DWT):
\begin{equation} \label{eq:dwt}
    \begin{split}
        &\left( \ba_J,\{\bd_j\}_{j=1}^J \right) = \mathcal{W}(\widetilde{\bx}) \quad \left(\ba^{(rec)}_J,\{\bd^{(rec)}_j\}_{j=1}^J\right) = \mathcal{W}(\widetilde{\bx}^{(rec)}) \\
        &\alpha = \text{Threshold}(\bd_1) \quad \mathcal{L}_{hf} = \sum_{j=1}^{J'} \left\Vert (\bd_j - \bd^{(rec)}_j) \odot \mathbbm{1}[\bd_j \ge \alpha] \right\Vert_1
    \end{split}
\end{equation}
where $\mathcal{W}(\bigcdot)$ denotes the DWT, producing approximation coefficients $\ba_J$ and detail coefficients $\{\bd_j\}_{j=1}^J$ with smaller $j$ corresponding to higher frequency. $\text{Threshold($\bigcdot$)}$ computes a threshold to distinguish salient coefficients from noise~\cite{donoho1994ideal}. Only $J'=2$ finest scales are included in the loss.

The final structure-preserving loss is computed by:
\begin{equation} \label{eq:loss_all}
    \mathcal{L}_{struct} = \mathcal{L}_{L1} + \lambda_{adv}\mathcal{L}_{adv} + \lambda_{sp}\mathcal{L}_{sp} + \lambda_{hf}\mathcal{L}_{hf}
\end{equation}
The four terms denote $L1$, adversarial, surrogate perceptual and high-frequency wavelet losses. Weights are dynamically adjusted using the adaptive weighting strategy proposed in \cite{esser2021taming}. Details of this strategy, along with other referenced techniques (i.e., FSQ, DWT Threshold), are in Appendix~\ref{sec:appendix_techniques}.

\section{UniTok-FM: A General-Purpose Time Series Foundation Model} \label{sec:unitok-fm}

\begin{figure*}[tb!] 
\centering
     \includegraphics[width=0.99\textwidth]{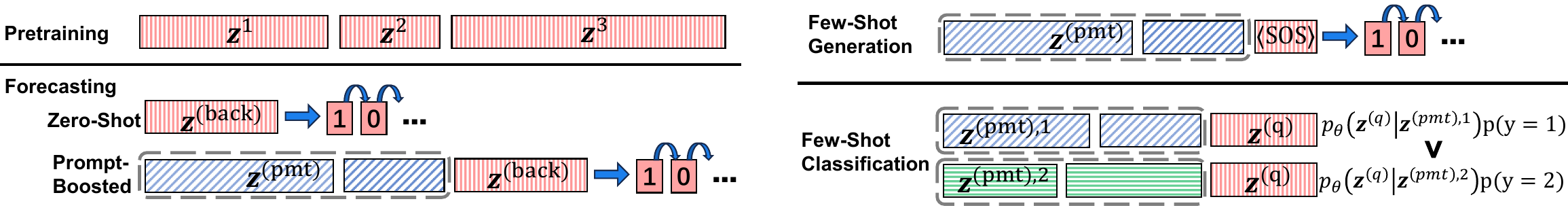}\vspace{-5pt}
    \caption{\textbf{Token arrangement for in-context NTP pretraining and training-free in-context inference.} In pretraining, multiple series with similar patterns are concatenated into a context window. In zero-shot forecasting, lookback tokens condition AR generation of future tokens. In prompt-boosted forecasting and few-shot generation, similar-pattern series are prepended as contextual prompts. In few-shot classification, the query series is conditioned on class-specific prompts, with labels determined by comparing likelihoods.}
    \label{fig:token_arrange}
    \vspace{-15pt}
\end{figure*}

As illustrated in Fig.~\ref{fig:token_arrange}, built on UniTok, UniTok-FM is pretrained on context windows formed by multiple similar-pattern series (Sec.~\ref{sec:pretrain}). Training-free in-context inference is achieved by AR token generation or likelihood evaluation with different prompts (Sec.~\ref{sec:inference}).

\subsection{In-Context Next-Token Prediction} \label{sec:pretrain}

With UniTok, each TS is transformed into a discrete token sequence for NTP. UniTok-FM adopts modern LLM architectures without modification. Instead of performing NTP on isolated series, we aggregate multiple series with similar temporal patterns into a context window and conduct NTP over it, allowing the model to capture shared dynamics. Formally, given a set of series with similar patterns $\{\bx^{i}\}_{i=1}^N, \bx^{i} \in \bbR^{T_i}$, the pretraining is conducted as:

\begin{equation}
    \bz^{i} = \text{Enc}(\bx^{i}) \quad \bz^{(ctx)} = [\bz^{1}, \dots, \bz^{N}] \quad \mathcal{L}_{NTP} = \sum_{l=1}^{L_{ctx}-1}\text{CE}\left(z^{(ctx)}_{l+1}, p_{\theta}(z^{(ctx)}_{l+1}|\bz^{(ctx)}_{\le l})\right)
\end{equation}

where $L_{ctx}$ denotes the length of the context window, $\text{CE}$ is the cross-entropy loss and $p_{\theta}(\bigcdot)$ is a LLM backbone. Each series $\bx^{i}$ is encoded into a token sequence $\bz^{i}$. These sequences are then concatenated to form a context window $\bz^{(ctx)}$ on which NTP is performed. In pretraining, $\{\bx^{i}\}_{i=1}^N$ is obtained by extracting non-overlapping segments from the same long series. There can be gaps between segments, and segments are arranged in time order to prevent future information leakage. In inference, it generalizes beyond this construction to meet the requirements of each task (Sec.~\ref{sec:inference}).

\subsection{Training-Free In-Context Inference} \label{sec:inference}

\textbf{Zero-Shot Forecasting} Given lookback window $\bx^{(back)}\in\bbR^{T_{back}}$, next $\tau$ points is predicted as:
\begin{equation} \label{eq:zero-shot}
    \begin{split}
        &\bz^{(back)} = \text{Enc}(\bx^{(back)}) \quad L_{pred} = \phi(T_{back}+\tau)-\phi(T_{back}) \\
        &z^{(pred)}_{l+1} \sim p_{\theta}(z^{(pred)}_{l+1}|[\bz^{(back)}, \bz^{(pred)}_{\le l}]) \text{ for } l = 0, \dots,  L_{pred}-1\\
        &\bx^{(pred)} = \text{Dec}([\bz^{(back)}, \bz^{(pred)}])_{T_{back}+1:T_{back}+\tau}
    \end{split}
\end{equation}
$\bx^{(back)}$ is encoded into $\bz^{(back)}$, after which $L_{pred}$ tokens are generated autoregressively. The decoded suffix corresponding to the future horizon is taken as the prediction. As a probabilistic model, multiple trajectories are sampled to estimate the future distribution following~\cite{ansari2024chronos}.

\textbf{Prompt-Boosted Forecasting} Series exhibiting similar patterns can serve as contextual prompts to guide predictions. Such prompt series may be extracted from the target series’ own historical records (e.g., weather records from previous years) or from other entities (e.g., observations from nearby stations).  Given prompts $\{\bp^{i}\}_{i=1}^N$, we extend Eq.~\ref{eq:zero-shot} as:
\begin{equation} \label{eq:prompt-forecast}
    \begin{split}
        \bz^{(pmt)} = [\text{Enc}(\bp^{1}), \dots, \text{Enc}(\bp^{N})] \quad z^{(pred)}_{l+1} \sim p_{\theta}(z^{(pred)}_{l+1}|[\bz^{(pmt)}, \bz^{(past)}, \bz^{(pred)}_{\le l}])
    \end{split}
\end{equation}
Each prompt series is encoded and concatenated to form the prompt sequence $\bz^{(pmt)}$. It is then prepended to the tokenized lookback window $\bz^{(past)}$ for AR generation.

\textbf{Few-Shot Generation} Given prompt $\{\bp^{i}\}_{i=1}^N$, a length-$\tau$ series with similar dynamic is generated:
\begin{equation} \label{eq:generation}
    \begin{split}
        &\bz^{(pmt)} = \text{Enc}(\{\bp^{i}\}_{i=1}^N), \quad z^{(gen)}_1 = \langle \text{SOS}\rangle\ \\
        & z^{(gen)}_{l+1} \sim p_{\theta}(z^{(gen)}_{l+1}|[\bz^{(pmt)}, \bz^{(gen)}_{\le l}]) \text{ for } l = 1, \dots, \phi(\tau)-1\\
        &\bx^{(gen)} = \text{Dec}(\bz^{(gen)})
    \end{split}
\end{equation}
$\bz^{(pmt)}$ is constructed same as in prompt-boosted forecasting. Generation is initialized with a $\langle\text{SOS}\rangle$.

\textbf{Few-Shot Classification} By tokenizing TS, UniTok-FM can evaluate the conditional likelihood of a series, enabling in-context few-shot classification. Given $M$ classes each with $N_m$ examples per class, $\{\{\bp^{m, i}\}_{i=1}^{N_m}\}_{m=1}^M$, the class label $y$ of a query series $\bx^{(q)}$ is inferred as:
\begin{equation} \label{eq:classification}
    \begin{split}
        &\bz^{(q)} = \text{Enc}(\bx^{(q)}) \quad \forall m : \bz^{(pmt),m} = \text{Enc}(\{\bp^{m, i}\}_{i=1}^{N_m}) \\ &p(\bx^{(q)}|y=m) \approx p_{\theta}(\bz^{(q)}|\bz^{(pmt),m}) \quad p(y=m|\bx^{(q)}) \propto p(\bx^{(q)}|y=m)p(y=m)
    \end{split}
\end{equation}
The query series is encoded into $\bz^{(q)}$, while examples from each class form a class-specific prompt $\bz^{(pmt),m}$. UniTok-FM approximates the class-conditional likelihood $p(\bx^{(q)}|y=m)$ using the likelihood of tokenized query conditioned on corresponding prompt, $p_{\theta}(z^{(q)}|\bz^{(pmt),m})$, which has a closed-form solution under AR factorization. Posterior class probability $p(y=m|\bx^{(q)})$ is obtained by Bayes’ rule. Intuitively, this procedure evaluates which class is most likely to generate the query, sharing the spirit of using generative models as classifiers~\cite{li2023your}.

\begin{table*}[tb!]
    \centering
    \caption{\textbf{Forecasting performance on the GIFT-Eval benchmark.} Forecasting TSFM denotes forecasting-specific TSFMs, while General TSFM denotes general-purpose TSFMs. Bold indicates the best model. Chronos is underlined as the most closely related baseline. Our methods are highlighted in gray. Full results on each dataset are in Tab.~\ref{tab:crps_full}-\ref{tab:mase_full} of Appendix~\ref{sec:full_results}.}
    \vspace{-8pt}
    \label{tab:main_gift}
\resizebox{\textwidth}{!}{
    \begin{tabular}{c|c|ccc||c|c|ccc}
    \toprule[1.5pt]
    Method Type & Method & CRPS$\downarrow$ & MAPE$\downarrow$ & MASE$\downarrow$ & Method Type & Method & CRPS$\downarrow$ & MAPE$\downarrow$ & MASE$\downarrow$ \\
    \midrule[1pt]
    \multirow{4}{*}{Statistical} & Naive & 1.591 & 1.055 & 1.270 & \multirow{7}{*}{Forecasting TSFM} & TiRex~\cite{auer2025tirex} & 0.488 & 0.677 & 0.716 \\
     & Seasonal-Naive & 1.000 & 1.000 & 1.000 &  & Sundial~\cite{liu2025sundial} & 0.559 & 0.777 & 0.750 \\
     & Auto-Theta & 1.244 & 1.126 & 1.090 &  & Chronos-Bolt~\cite{Ansari2024bolt} & 0.574 & 0.775 & 0.808 \\
     & Auto-Arima & 0.912 & 1.033 & 1.074 &  & Moirai~\cite{woo2024unified} & 0.610 & 0.825 & 0.901 \\
     \mycmidrule{0.5pt}{r}{1-5}
    \multirow{4}{*}{Supervised} & Crossformer~\cite{zhang2023crossformer} & 1.637 & 1.024 & 2.574 &  & Chronos~\cite{ansari2024chronos} & \underline{0.652} & \underline{0.802} & \underline{0.876} \\
     & DLinear~\cite{zeng2023transformers} & 0.846 & 1.086 & 1.061 &  & VisionTS~\cite{chen2025visionts} & 0.755 & 0.925 & 0.863 \\
     & PatchTST~\cite{nie2023time} & 0.587 & 0.788 & 0.849 &  & Lag-Llama~\cite{rasul2023lagllama} & 0.880 & 1.115 & 1.228 \\
    \mycmidrule{0.5pt}{}{6-10}
     & iTransformer~\cite{liu2024itransformer} & 0.620 & 0.846 & 0.893 & \multirow{2}{*}{General TSFM} & \cellcolor{gray!30} UniTok-FM(ZeroShot) & \cellcolor{gray!30} 0.591 & \cellcolor{gray!30} 0.798 & \cellcolor{gray!30} 0.851 \\
     \mycmidrule{0.5pt}{r}{1-5}
    Forecasting TSFM & Chronos-2~\cite{ansari2025chronos2} & \textbf{0.485} & \textbf{0.666} & \textbf{0.698} & & \cellcolor{gray!30} UniTok-FM(Prompt) & \cellcolor{gray!30} 0.573 & \cellcolor{gray!30} 0.761 & \cellcolor{gray!30} 0.824\\
    \bottomrule[1.5pt]
    \end{tabular}
}
\vspace{-10pt}
\end{table*}

\begin{table*}[tb!]
    \centering
    \caption{\textbf{Generation performance on Stocks, ETTh, Energy and fMRI.} \#Train denotes the number of training examples for baseline models, while \#Prompt denotes the number of in-context prompt examples for UniTok-FM. Pred Score and Disc Score indicate the average predictive and discriminative score across four datasets. Full results are in Tab.~\ref{tab:generation_full} of Appendix~\ref{sec:full_results}.}
    \vspace{-8pt}
    \label{tab:main_generate}
\resizebox{0.99\textwidth}{!}{
    \begin{tabular}{c|ccc|ccc|ccc|ccc|ccc|>{\columncolor{gray!30}}c}
    \toprule[1.5pt]
    Methods & \multicolumn{3}{c|}{Diffusion-TS~\cite{yuan2024diffusionts}} & \multicolumn{3}{c|}{SDFormer~\cite{chen2024sdformer}} & \multicolumn{3}{c|}{TimeGAN~\cite{yoon2019time}} & \multicolumn{3}{c|}{TimeVAE~\cite{desai2021timevae}} & \multicolumn{3}{c|}{Cot-GAN~\cite{xu2020cot}} & UniTok-FM \\
    \midrule[1pt]
    \#Train / \#Prompt & 5 & 200 & 1000 & 5 & 200 & 1000 & 5 & 200 & 1000 & 5 & 200 & 1000 & 5 & 200 & 1000 & 5 \\
    \midrule[1pt]
    Pred Score$\uparrow$ & 0.293 & 0.596 & 0.582 & 0.239 & 0.337 & 0.573 & -0.068 & 0.020 & 0.051 & 0.157 & 0.594 & \textbf{0.612} & 0.520 & 0.573 & 0.602 & 0.601 \\
    Disc Score$\downarrow$ & 0.497 & 0.311 & 0.351 & 0.487 & 0.147 & \textbf{0.096} & 0.500 & 0.500 & 0.500 & 0.483 & 0.365 & 0.423 & 0.451 & 0.399 & 0.486 & 0.420 \\
    \bottomrule[1.5pt]
    \end{tabular}
}
\vspace{-15pt}
\end{table*}

\section{Experiments} \label{sec:experiments}
\subsection{Pretraining Protocols} \label{sec:exp-detail}
\textbf{Datasets} We pretrain UniTok and UniTok-FM on the union of the GIFT-Pretrain~\cite{aksu2024gift} and Chronos-Dataset~\cite{ansari2024chronos}. Benchmark datasets are strictly filtered out from the pretraining corpus.

\textbf{UniTok} We pretrain a 113M-parameter UniTok with $S=4$ blocks and a codebook size of $C=1,940$. The maximum supported series length is 2048. Training runs 200K steps using AdamW optimizer with a global batch size of 512. The learning rate is linearly warmed up to $1\times10^{-4}$ over the first 2,000 steps and then cosine-decayed to $1\times10^{-5}$.

\textbf{UniTok-FM} During UniTok-FM training, the pretrained UniTok is frozen. We adopt the Qwen3~\cite{qwen3} as the backbone and train a 129M-parameter model from scratch. Other LLM architectures and model sizes are evaluated in Sec.~\ref{sec:model_analysis}. One context window supports up to 800 tokens, corresponding to 5 series of length 2048 or 8 series of length 1024. Training runs for 100K steps with a global batch size of 384, using the same optimizer and learning-rate schedule as UniTok. Both UniTok and UniTok-FM are trained on 4 NVIDIA A100 GPUs, with details shwon in Appendices~\ref{sec:appendix_pretrain} and~\ref{sec:appendix_benchmark}.

\subsection{Main Results}
\textbf{Zero-Shot\&Prompt-Boosted Forecasting} We evaluate forecasting on GIFT-Eval~\cite{aksu2024gift}, which comprises 97 tasks with different datasets and prediction horizons. Probabilistic forecasting is evaluated using Continuous Ranked Probability Score (CRPS), while point forecasting is evaluated by Mean Absolute Percentage Error (MAPE) and Mean Absolute Seasonal Error (MASE). Three categories of methods are compared: statistical methods, supervised models and forecasting-specific TSFMs. Two UniTok-FM variants are evaluated: \textbf{1) UniTok-FM(ZeroShot)} performs standard zero-shot forecasting; \textbf{2) UniTok-FM(Prompt)} enables prompt-boosted forecasting by first retrieving prompt series from the target’s earlier history and, if the context window is not full, from the training split of other entities within the same dataset.

Tab.~\ref{tab:main_gift} shows that UniTok-FM(ZeroShot) matches the strongest supervised baseline, PatchTST, while outperforming other statistical and supervised methods, with prompt boosting yielding consistent gains. Although UniTok-FM does not exceed SOTA models such as Chronos-2 and TiRex, it consistently outperforms Moirai, Chronos, VisionTS, and Lag-Llama, and is competitive with Chronos-Bolt. Notably, most TSFMs rely on forecasting-specific designs (e.g., multi-horizon heads and quantile objectives), whereas UniTok-FM adopts standard LLM-style NTP pretraining and AR inference.  Moreover, UniTok-FM substantially outperforms Chronos, which employs pointwise binning tokenization, highlighting the importance of an expressive tokenizer for NTP.

\begin{figure*}[tb!]
\centering
    \includegraphics[align=t, width=0.18\textwidth, height=0.1\textwidth]{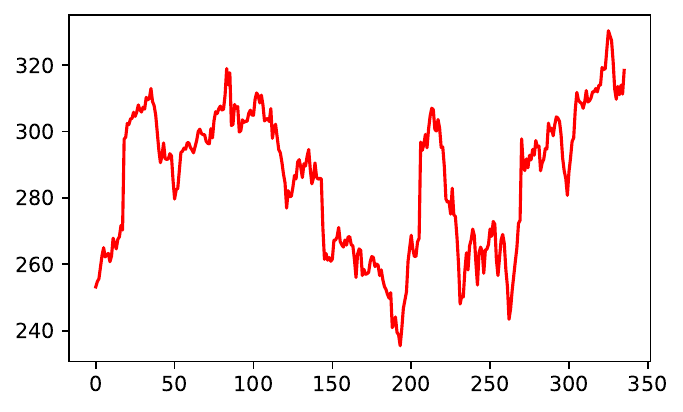}
    \includegraphics[align=t, width=0.18\textwidth, height=0.1\textwidth]{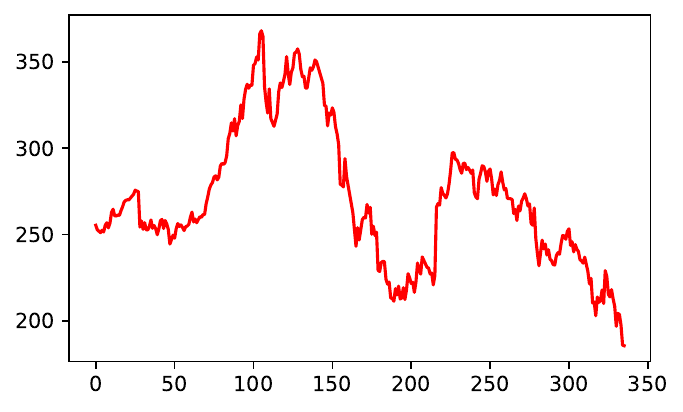}
    \includegraphics[align=t, width=0.18\textwidth, height=0.1\textwidth]{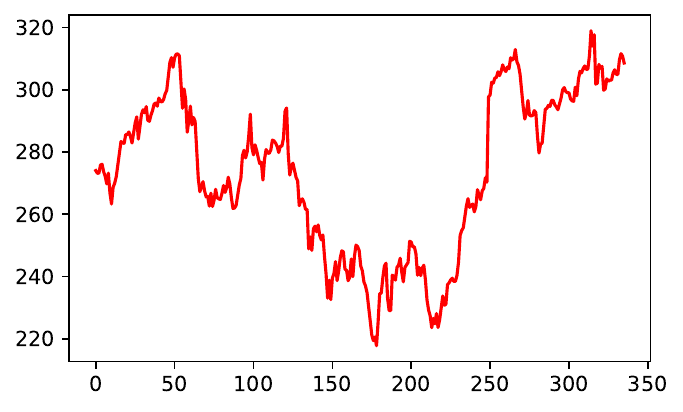}
    \includegraphics[align=t, width=0.18\textwidth, height=0.1\textwidth]{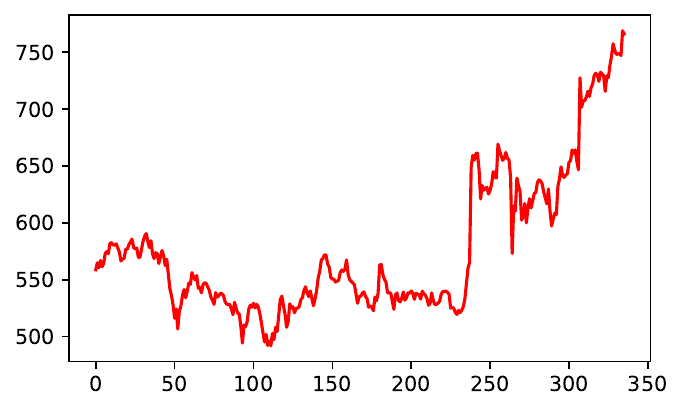}
    \includegraphics[align=t, width=0.18\textwidth, height=0.1\textwidth]{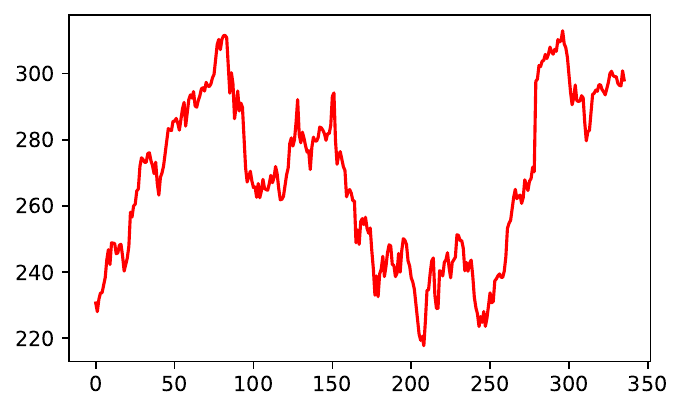} \vspace{-2pt} \\
    \includegraphics[align=t, width=0.18\textwidth, height=0.1\textwidth]{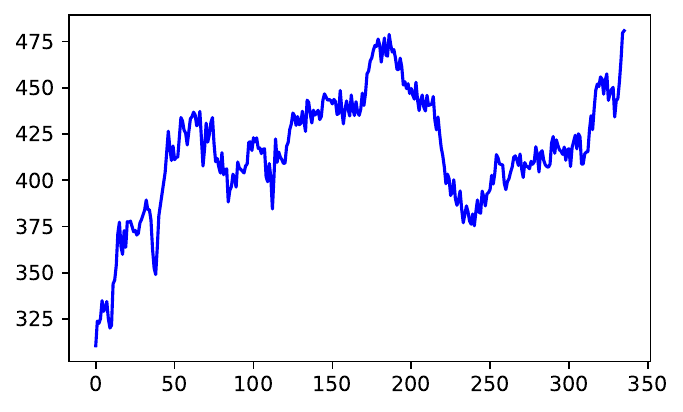}
    \includegraphics[align=t, width=0.18\textwidth, height=0.1\textwidth]{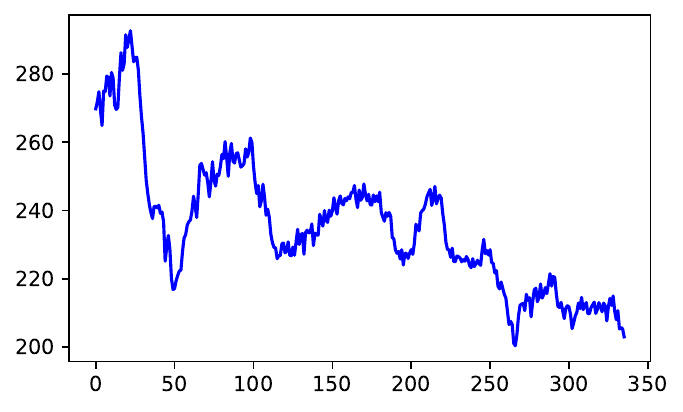}
    \includegraphics[align=t, width=0.18\textwidth, height=0.1\textwidth]{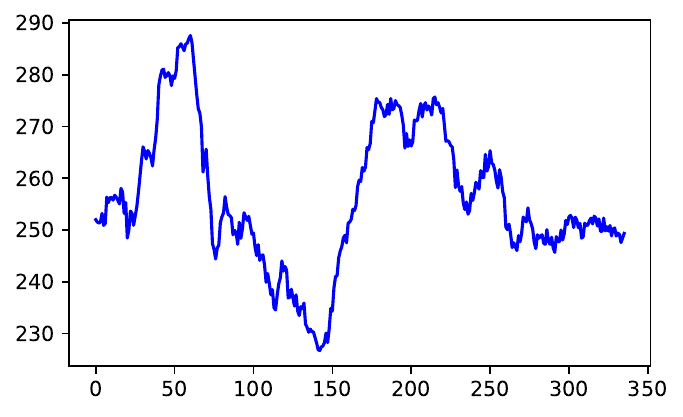}
    \includegraphics[align=t, width=0.18\textwidth, height=0.1\textwidth]{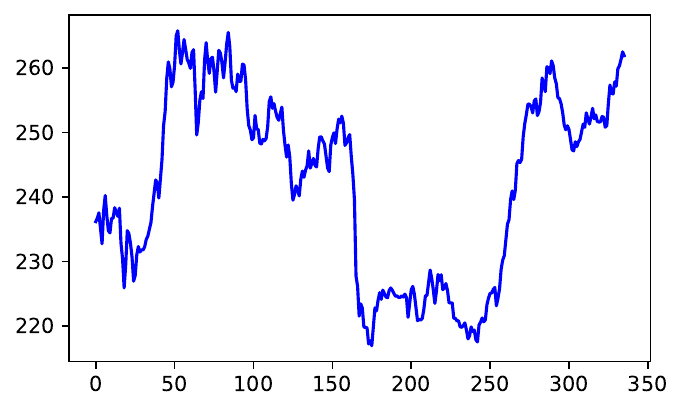}
    \includegraphics[align=t, width=0.18\textwidth, height=0.1\textwidth]{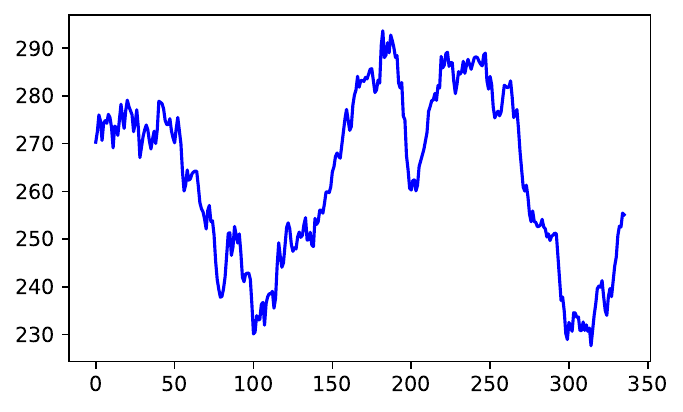}
    \vspace{-8pt}
    \caption{All prompt examples (red) and sampled generations (blue) of UniTok-FM on Stocks.} 
    \label{fig:generated_samples}
    \vspace{-12pt}
\end{figure*}

\begin{figure*}[tb!]
\centering
    \subcaptionbox{}{\includegraphics[align=t, width=0.3\textwidth, height=0.18\textwidth]{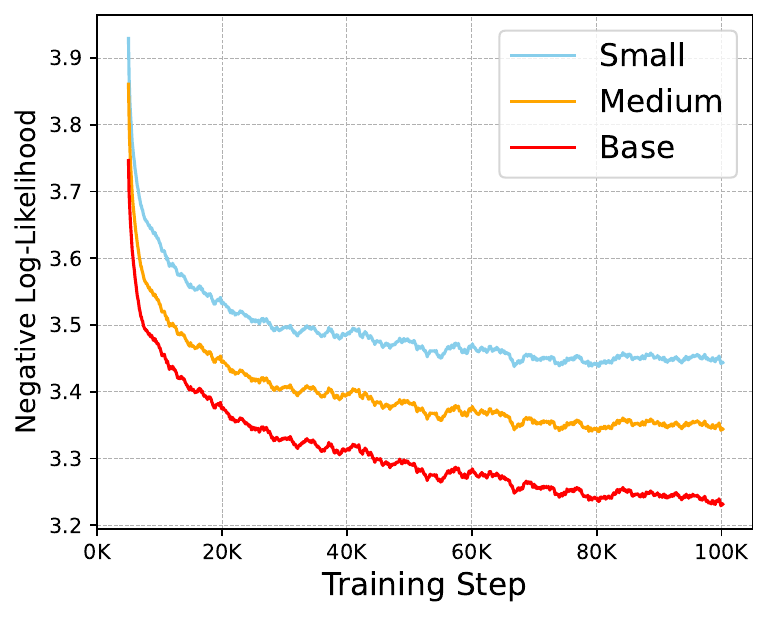}\vspace{-8pt}}
    \subcaptionbox{}{\includegraphics[align=t, width=0.25\textwidth, height=0.18\textwidth]{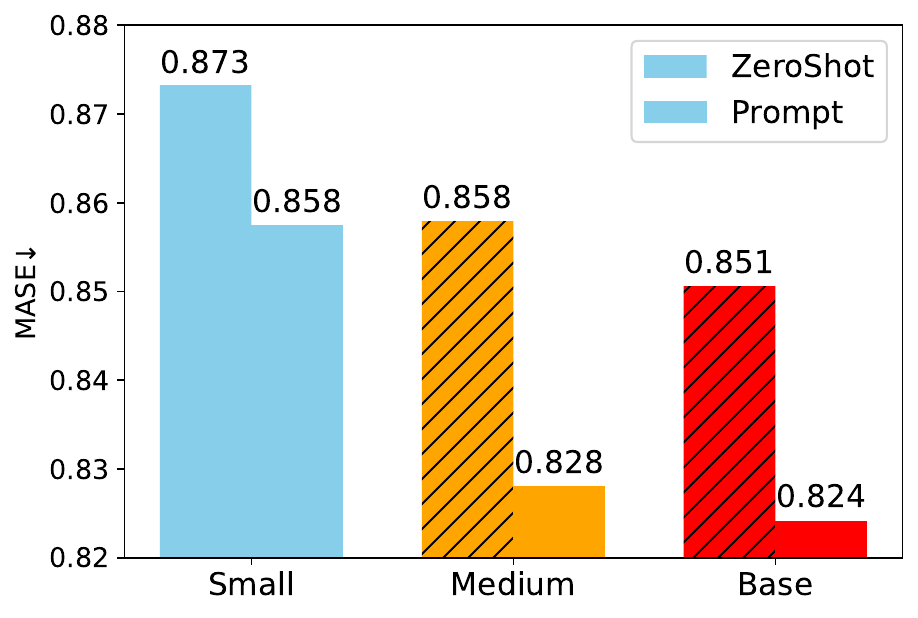}\vspace{-8pt}}
    \subcaptionbox{}{\includegraphics[align=t, width=0.2\textwidth, height=0.18\textwidth]{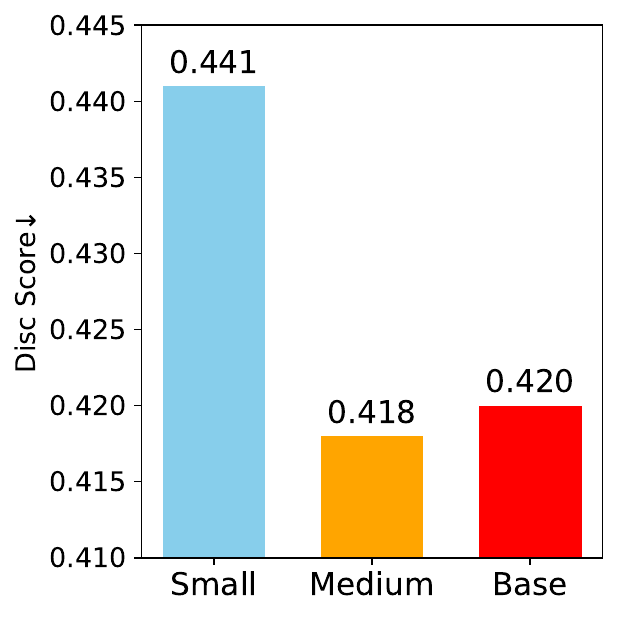}\vspace{-8pt}}
    \subcaptionbox{}{\includegraphics[align=t, width=0.2\textwidth, height=0.18\textwidth]{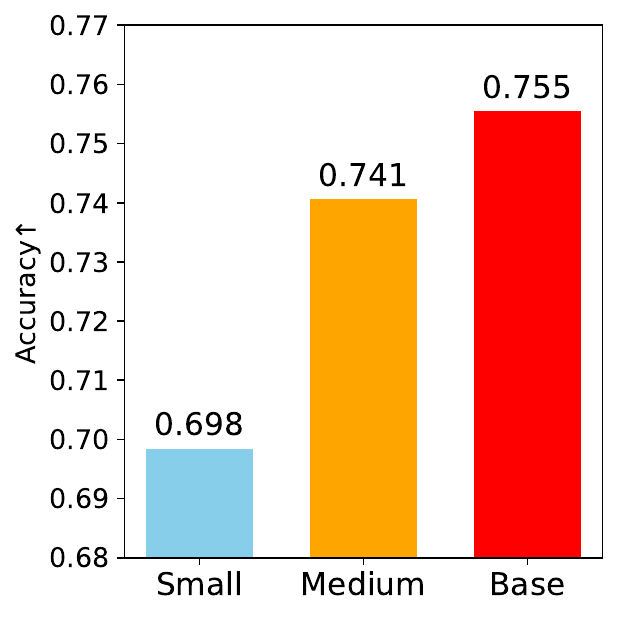}\vspace{-8pt}}
    \vspace{-8pt}
    \caption{\textbf{Scaling behavior across LLM backbone sizes.} Qwen3 backbones of three sizes are evaluated: Small (14M), Medium (26M), and Base (129M). (a) Training loss. (b) Forecasting MASE. (c) Generation discriminative score. (d) Classification accuracy.} 
    \label{fig:scale}
    \vspace{-20pt}
\end{figure*}

\begin{wrapfigure}{l}{0.5\textwidth}
    \vspace{-10pt}
    \centering
    \makeatletter\def\@captype{table}\makeatother\caption{\textbf{Classification performance on the UCR-FewShot.} Acc denotes the average accuracy over 53 datasets, and \#Win counts the number of datasets on which a method achieves the best performance (including ties). Full results are in Tab.~\ref{tab:classify_full} of Appendix~\ref{sec:full_results}.} 
    \vspace{-8pt}
   \label{tab:main_classify}
    \resizebox{\linewidth}{!}{
    \begin{tabular}{c|c|cc}
    \toprule[1.5pt]
    Method Type & Method & Acc$\uparrow$ & \#Win$\uparrow$ \\
    \midrule[1pt]
    \multirow{3}{*}{Statistical} & KNN & 0.672 & 5 \\
    & TStree~\cite{deng2013time} & 0.628 & 2 \\
     & RDST~\cite{guillaume2022random} & 0.673 & 0 \\
    \midrule[0.5pt]
    \multirow{3}{*}{Supervised} & FCN~\cite{wang2017time} & 0.357 & 0 \\
     & LITE~\cite{ismail2025look} & 0.472 & 3 \\
     & Inception~\cite{ismail2020inceptiontime} & 0.491 & 6 \\
     \midrule[0.5pt]
    Classfication TSFM & Mantis~\cite{feofanov2025mantis} & \textbf{0.840} & \textbf{27} \\
     \midrule[0.5pt]
    General TSFM & UniTS~\cite{gao2024units} & 0.697 & 1 \\
    (Downstream Classfier) & Moment~\cite{goswami2024moment} & 0.778 & 9 \\
     \midrule[0.5pt]
     \rowcolor{gray!30}
    General TSFM (In-context) & UniTok-FM & 0.755 & 10 \\
    \bottomrule[1.5pt]
    \end{tabular}
    }
    \vspace{-10pt}
\end{wrapfigure}

\textbf{Few-Shot Generation} Following \cite{yuan2024diffusionts}, we evaluate few-shot generation on four datasets (Stocks, ETTh, Energy, and fMRI) using two metrics: 1) predictive score, measuring the forecasting accuracy of a predictor trained on generated data; 2) discriminative score, measuring how well a discriminator distinguishes generated data from real data. UniTok-FM conducts training-free in-context inference using five examples. As conventional generative models cannot be trained with so few samples, we train each baseline on progressively larger sets of 5, 200, and 1000 samples, where larger sets subsume smaller ones. The five-sample setting matches the examples provided to UniTok-FM.

Tab.~\ref{tab:main_generate} shows that UniTok-FM, using only five prompt examples, achieves a predictive score comparable to the best generative models trained on 1K samples, demonstrating its ability to capture underlying dynamics via in-context inference. UniTok-FM also consistently outperforms baselines in discriminative score under the same sample budget. Since a discriminative score of 0.5 indicates perfect distinguishability between generated and real samples, UniTok-FM’s substantially lower score (0.420) suggests that its generated series are non-trivial. See Fig.~\ref{fig:generated_samples} for qualitative evaluation.

\textbf{Few-Shot Classification} We evaluate few-shot classification on the UCR Archive~\cite{UCRArchive2018}. The original archive contains 128 datasets, and we select those with at most 20 training instances per class, yielding a subset of 53 datasets, denoted as UCR-FewShot. The scarcity of labeled data makes this it suitable for few-shot classification. Besides the three categories compared in forecasting, we include general-purpose TSFMs that rely on downstream classifiers. These models extract features using a pretrained model and then train a classifier on them. This paradigm fundamentally differs from UniTok-FM, which uses training-free in-context inference. Notably, all TSFMs except ours are pretrained on corpora including UCR training splits, giving them prior exposure to the benchmark distribution.

Tab.~\ref{tab:main_classify} shows that UniTok-FM outperforms statistical and supervised baselines. Among TSFMs, it consistently surpasses UniTS and is comparable to MOMENT, despite requiring no dataset-specific fine-tuning and never accessing UCR data during pretraining. Although UniTok-FM is pretrained only on forecasting-oriented corpora and NTP is often viewed as suboptimal for discriminative tasks~\cite{feofanov2025mantis}, its competitive performance demonstrates NTP's strong generalizability for TS understanding.

\subsection{Model Analysis} \label{sec:model_analysis}

\textbf{Scaling Behavior across LLM Backbone Size} The model with 129M-parameter Qwen3 backbone is denoted as \emph{Base}. Halving the hidden dimension creates \emph{Medium} (26M), and further halving layers yields \emph{Small} (14M). Fig.~\ref{fig:scale}(a) shows that larger backbones consistently achieve lower training loss. Figs.~\ref{fig:scale}(b–d) demonstrate a clear scaling trend on downstream tasks where performance generally improves with increasing backbone size. For efficiency, unless otherwise stated, subsequent analyses in this section are conducted on the Medium.

\begin{figure*}[tb!]
\centering
    \subcaptionbox{}{\includegraphics[align=t, width=0.19\textwidth]{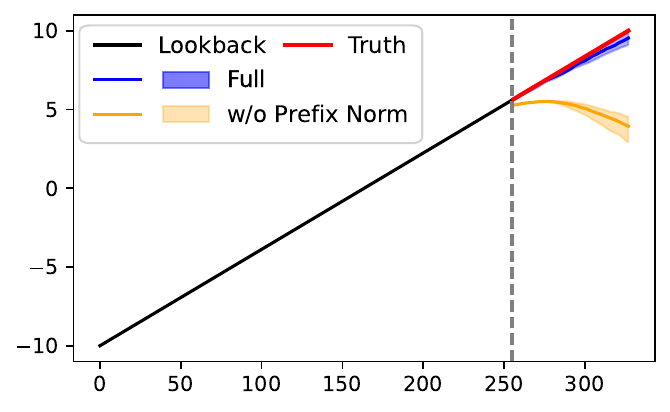}\vspace{-8pt}}
    \raisebox{-8pt}{\rule{0.5pt}{0.12\textwidth}}   
    \subcaptionbox{}{\includegraphics[align=t, width=0.19\textwidth]{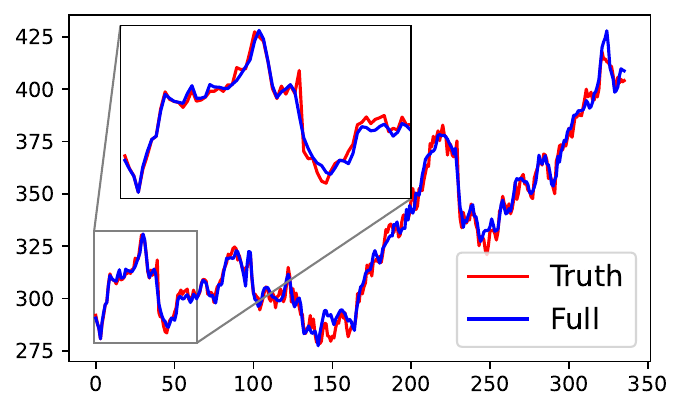}\hspace{-5pt}
    \includegraphics[align=t, width=0.19\textwidth]{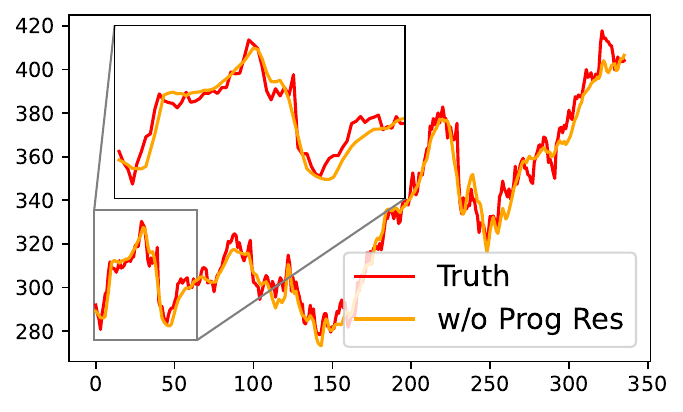}\vspace{-8pt}}
    \raisebox{-8pt}{\rule{0.5pt}{0.12\textwidth}}
    \subcaptionbox{}{\includegraphics[align=t, width=0.19\textwidth]{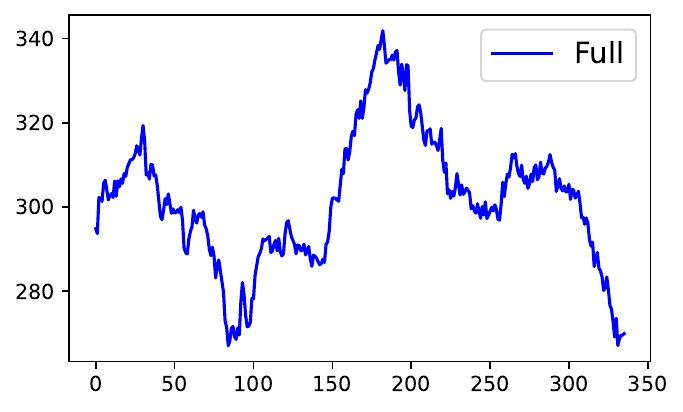}\hspace{-5pt}
    \includegraphics[align=t, width=0.19\textwidth]{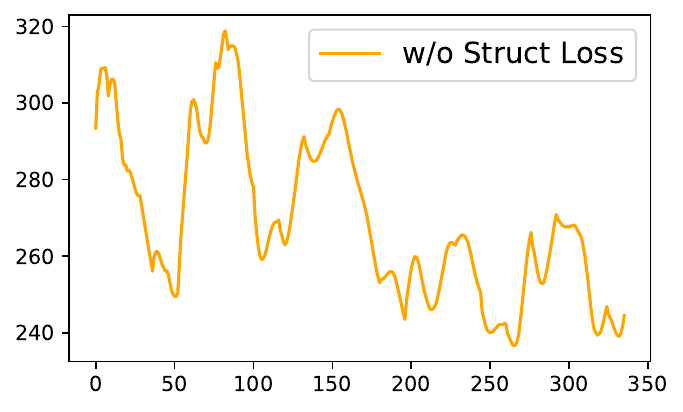}\vspace{-8pt}}
    \vspace{-8pt}
    \caption{\textbf{Qualitative comparison between the full UniTok and ablated variants.} (a) Zero-shot forecasting: full v.s. prefix normalization ablated. (b) Series reconstruction: full v.s. progressive-resolution causal autoencoder ablated. (c) Series generation: full v.s. structure-preserving reconstruction loss ablated, using same prompts as Fig.~\ref{fig:generated_samples}. Blue: full UniTok; Orange: ablated variants.} 
    \label{fig:ablation}
    \vspace{-20pt}
\end{figure*}

\begin{wrapfigure}{l}{0.5\textwidth}
    \centering
    \vspace{-10pt}\makeatletter\def\@captype{table}\makeatother\caption{\textbf{Performance of UniTok-FM across downstream tasks with different LLM architectures.} ZS: ZeroShot; PMT: Prompt; Disc: discriminative score.} 
    \vspace{-8pt}
   \label{tab:llm_arch}
    \resizebox{\linewidth}{!}{
    \begin{tabular}{c|c|ccc>{\columncolor{gray!30}}c}
    \toprule[1.5pt]
    \multirow{2}{*}{Task} & \multirow{2}{*}{Metric} & GPT2 & LLama2 & Gemma2 & Qwen3 \\
     & & \cite{radford2019language} & \cite{touvron2023llama} & \cite{team2024gemma} & \cite{qwen3} \\
    \midrule[1pt]
    Forecast (ZS) & MASE$\downarrow$ & 0.862 & 0.861 & 0.866 & \textbf{0.858} \\
    Forecast (PMT) & MASE$\downarrow$ & 0.907 & 0.840 & 0.842 & \textbf{0.828} \\
    Generation & Disc$\downarrow$ & 0.439 & 0.425 & 0.445 & \textbf{0.418} \\
    Classification & Acc$\uparrow$ & 0.659 & 0.739 & 0.694 & \textbf{0.741} \\
    \bottomrule[1.5pt]
    \end{tabular}
    }
    \vspace{5pt}\makeatletter\def\@captype{table}\makeatother\caption{\textbf{Ablation study of UniTok components.} \textit{w/o Prefix Norm} replaces prefix normalization with whole-series norm. \textit{w/o Prog Res} replaces the progressive-resolution up/downsample with uniform ones. \textit{w/o Struct Loss} replaces the structure-preserving reconstruction loss with $L_1$.} 
    \vspace{-8pt}
    \label{tab:ablation}
    \resizebox{\linewidth}{!}{
    \begin{tabular}{c|c|>{\columncolor{gray!30}}cccc}
    \toprule[1.5pt]
    \multirow{2}{*}{Task} & \multirow{2}{*}{Metric} &  & w/o Prefix & w/o Prog & w/o Struct \\
     & & \multirow{-2}{*}{Full} & Norm & Res & Loss \\
    \midrule[1pt]
    Forecast (ZS) & MASE$\downarrow$ & 0.858 & 0.870 & 0.884 & \textbf{0.840} \\
    Forecast (PMT) & MASE$\downarrow$ & 0.828 & 0.845 & 0.866 & \textbf{0.816} \\
    Generation & Disc.$\downarrow$ & 0.418 & \textbf{0.407} & 0.438 & 0.470 \\
    Classification & Acc$\uparrow$ & \textbf{0.741} & 0.740 & 0.726 & 0.731 \\
    \bottomrule[1.5pt]
    \end{tabular}
    }
    \vspace{-15pt}
\end{wrapfigure}

\textbf{Generality across LLM Architectures} Beyond Qwen3, we adapt other LLM architectures as the AR backbone of UniTok-FM, including GPT2~\cite{radford2019language}, Llama2~\cite{touvron2023llama} and Gemma2~\cite{team2024gemma}. All models are scaled to Medium (26M). As shown in Tab.~\ref{tab:llm_arch}, UniTok-FM generalizes well across architectures and benefits from advances in LLM design, with more recent models outperforming the early-stage GPT2 on most tasks. Notably, thanks to the well-established LLM community, we can swap backbones by changing configuration files while reusing all interfaces.

\textbf{Ablation Study} Tab.~\ref{tab:ablation} shows that: \textbf{1)} Removing prefix normalization substantially degrades forecasting, primarily due to failures on non-stationary series where statistics estimated on the lookback window fail to generalize to the future (Fig.~\ref{fig:ablation}(a)). \textbf{2)} Removing progressive-resolution downsample/upsample degrades performance across all tasks by harming reconstruction quality, particularly severe at early positions with limited causal receptive field (Fig.~\ref{fig:ablation}(b)). \textbf{3)} Replacing the structure-preserving reconstruction loss with an $L1$ loss yields a slight gain in forecasting, as it mainly depends on low-frequency trends~\cite{xu2024fits}, but severely degrades generation, producing overly smooth samples (Fig.~\ref{fig:ablation}(c)). Overall, rather than optimizing for forecasting alone, UniTok integrates these components to form a universal tokenizer that generalizes across tasks.

\begin{wrapfigure}{l}{0.5\textwidth}
    \vspace{-10pt}
    \centering
    {\subcaptionbox{}{\includegraphics[align=t, width=0.155\textwidth, height=0.2\textwidth]{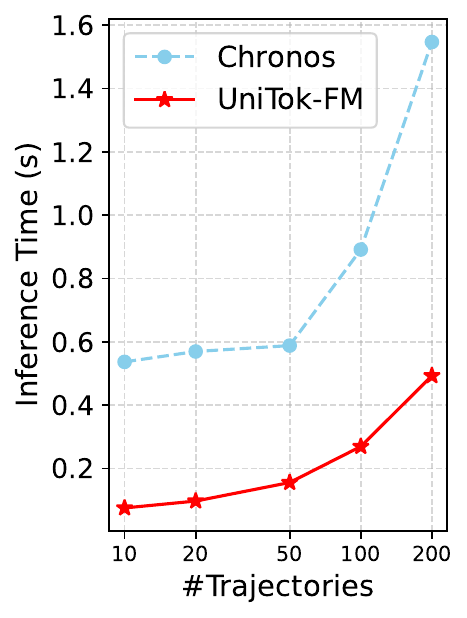}\vspace{-8pt}}
    \subcaptionbox{}{\includegraphics[align=t, width=0.155\textwidth, height=0.2\textwidth]{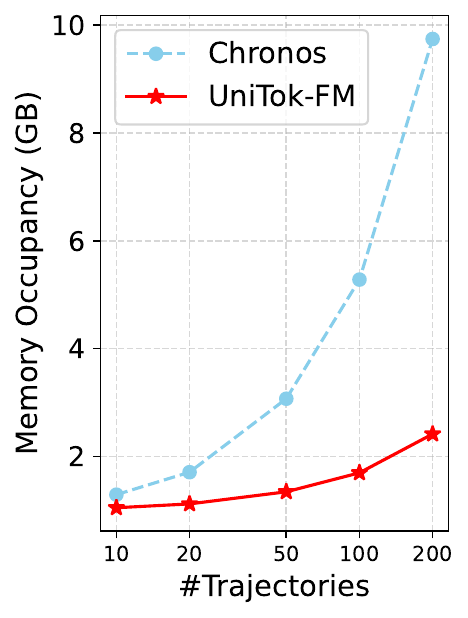}\vspace{-8pt}}
    \subcaptionbox{}{\includegraphics[align=t, width=0.155\textwidth, height=0.2\textwidth]{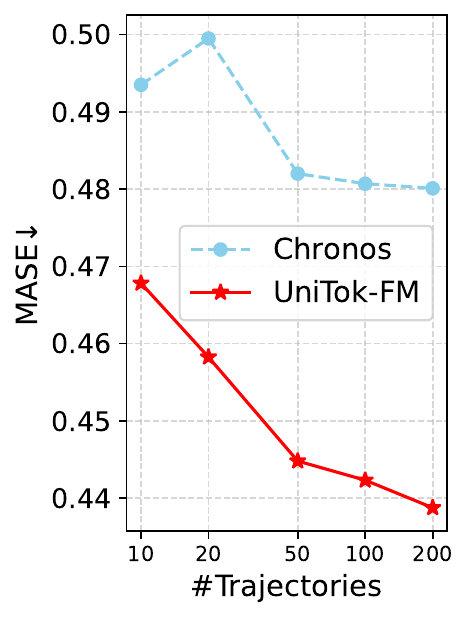}\vspace{-8pt}}}
    \vspace{-5pt}
    \caption{\textbf{Zero-shot forecasting efficiency comparison between Chronos and UniTok-FM on Jena Weather.} (a) Inference time per instance w.r.t number of sampling trajectories. (b) Memory occupancy. (c) Forecasting performance (MASE).}
    \label{fig:efficiency}
    \vspace{-15pt}
\end{wrapfigure}

\textbf{Inference Efficiency} UniTok produces much shorter token sequences than point-wise binning of Chronos, resulting in improved LLM inference efficiency. We compare Chronos (Base) and UniTok-FM (Base) under varying numbers of sampled trajectories on the Jena Weather (10min) from GIFT-Eval, using a single NVIDIA A100 GPU. The lookback window is 512 and the prediction horizon is 48, matching Chronos’ setup. UniTok-FM is evaluated in the zero-shot setting for a fair comparison. Fig.~\ref{fig:efficiency} shows that UniTok-FM not only achieves better forecasting performance but also consistently reduces inference time and memory usage.

\section{Conclusion, Limitation and Outlook} \label{sec:future-work}
We propose UniTok, a universal tokenizer that converts TS into discrete tokens, enabling next-token prediction on TS. Built on UniTok, we pretrain UniTok-FM, a general-purpose foundation model that supports zero-shot and prompt-boosted forecasting, as well as few-shot generation and classification via training-free in-context inference.

Despite the generality, a performance gap remains compared to task-specific SOTA models that benefit from carefully designed inductive biases. Moreover, this work focuses on univariate TS and relies on channel independence for multivariate data. Extending to multivariate settings requires careful modeling of cross-channel dependencies, which we leave for future work.

\bibliography{citations}

@string{cvpr    = {IEEE / CVF Computer Vision and Pattern Recognition Conference (CVPR)}}

@string{icml    = {International Conference on Machine Learning (ICML)}}

@string{neurips  = {Conference on Neural Information Processing Systems (NeurIPS)}}

@string{iclr    = {International Conference on Learning Representations (ICLR)}}

@string{aaai    = {AAAI Conference on Artificial Intelligence (AAAI)}}

@string{jmlr    = {Journal of Machine Learning Research (JMLR)}}

@string{aistats = {International Conference on Artificial Intelligence and Statistics (AISTATS)}}

@string{tmlr = {Transactions on Machine Learning Research (TMLR)}}

@string{ijcnn = {International Joint Conference on Neural Networks (IJCNN)}}

@inproceedings{van2017neural,
  title={Neural discrete representation learning},
  author={Van Den Oord, Aaron and Vinyals, Oriol and others},
  booktitle=neurips,
  year={2017}
}

@inproceedings{esser2021taming,
  title={Taming transformers for high-resolution image synthesis},
  author={Esser, Patrick and Rombach, Robin and Ommer, Bjorn},
  booktitle=cvpr,
  year={2021}
}

@inproceedings{
mentzer2024finite,
title={Finite Scalar Quantization: {VQ}-{VAE} Made Simple},
author={Fabian Mentzer and David Minnen and Eirikur Agustsson and Michael Tschannen},
booktitle=iclr,
year={2024},
}

@article{donoho1994ideal,
  title={Ideal spatial adaptation by wavelet shrinkage},
  author={Donoho, David L and Johnstone, Iain M},
  journal={biometrika},
  year={1994},
}

@inproceedings{ledig2017photo,
  title={Photo-realistic single image super-resolution using a generative adversarial network},
  author={Ledig, Christian and Theis, Lucas and Husz{\'a}r, Ferenc and Caballero, Jose and Cunningham, Andrew and Acosta, Alejandro and Aitken, Andrew and Tejani, Alykhan and Totz, Johannes and Wang, Zehan and others},
  booktitle=cvpr,
  year={2017}
}

@article{aksu2024gift,
  title={Gift-eval: A benchmark for general time series forecasting model evaluation},
  author={Aksu, Taha and Woo, Gerald and Liu, Juncheng and Liu, Xu and Liu, Chenghao and Savarese, Silvio and Xiong, Caiming and Sahoo, Doyen},
  journal={arXiv preprint arXiv:2410.10393},
  year={2024}
}

@article{ansari2024chronos,
  title={Chronos: Learning the language of time series},
  author={Ansari, Abdul Fatir and Stella, Lorenzo and Turkmen, Caner and Zhang, Xiyuan and Mercado, Pedro and Shen, Huibin and Shchur, Oleksandr and Rangapuram, Syama Sundar and Arango, Sebastian Pineda and Kapoor, Shubham and others},
  journal=tmlr,
  year={2024}
}

@article{qwen3,
    title={Qwen3 Technical Report}, 
    author={An Yang and Anfeng Li and Baosong Yang and Beichen Zhang and Binyuan Hui and Bo Zheng and Bowen Yu and Chang Gao and Chengen Huang and Chenxu Lv and Chujie Zheng and Dayiheng Liu and Fan Zhou and Fei Huang and Feng Hu and Hao Ge and Haoran Wei and Huan Lin and Jialong Tang and Jian Yang and Jianhong Tu and Jianwei Zhang and Jianxin Yang and Jiaxi Yang and Jing Zhou and Jingren Zhou and Junyang Lin and Kai Dang and Keqin Bao and Kexin Yang and Le Yu and Lianghao Deng and Mei Li and Mingfeng Xue and Mingze Li and Pei Zhang and Peng Wang and Qin Zhu and Rui Men and Ruize Gao and Shixuan Liu and Shuang Luo and Tianhao Li and Tianyi Tang and Wenbiao Yin and Xingzhang Ren and Xinyu Wang and Xinyu Zhang and Xuancheng Ren and Yang Fan and Yang Su and Yichang Zhang and Yinger Zhang and Yu Wan and Yuqiong Liu and Zekun Wang and Zeyu Cui and Zhenru Zhang and Zhipeng Zhou and Zihan Qiu},
    journal = {arXiv preprint arXiv:2505.09388},
    year={2025}
}

@inproceedings{zhang2023crossformer,
    title={Crossformer: Transformer Utilizing Cross-Dimension Dependency for Multivariate Time Series Forecasting},
    author={Yunhao Zhang and Junchi Yan},
    booktitle=iclr,
    year={2023}
}

@inproceedings{zeng2023transformers,
  title={Are transformers effective for time series forecasting?},
  author={Zeng, Ailing and Chen, Muxi and Zhang, Lei and Xu, Qiang},
  booktitle=aaai,
  year={2023}
}

@inproceedings{nie2023time,
  title     = {A Time Series is Worth 64 Words: Long-term Forecasting with Transformers},
  author    = {Nie, Yuqi and
               H. Nguyen, Nam and
               Sinthong, Phanwadee and 
               Kalagnanam, Jayant},
  booktitle = iclr,
  year      = {2023}
}

@inproceedings{liu2024itransformer,
  title={itransformer: Inverted transformers are effective for time series forecasting},
  author={Liu, Yong and Hu, Tengge and Zhang, Haoran and Wu, Haixu and Wang, Shiyu and Ma, Lintao and Long, Mingsheng},
  booktitle=iclr,
  year={2024}
}

@article{ansari2025chronos2,
  title        = {Chronos-2: From Univariate to Universal Forecasting},
  author       = {Abdul Fatir Ansari and Oleksandr Shchur and Jaris Küken and Andreas Auer and Boran Han and Pedro Mercado and Syama Sundar Rangapuram and Huibin Shen and Lorenzo Stella and Xiyuan Zhang and Mononito Goswami and Shubham Kapoor and Danielle C. Maddix and Pablo Guerron and Tony Hu and Junming Yin and Nick Erickson and Prateek Mutalik Desai and Hao Wang and Huzefa Rangwala and George Karypis and Yuyang Wang and Michael Bohlke-Schneider},
  journal      = {arXiv preprint arXiv:2510.15821},
  year         = {2025},
}

@inproceedings{
auer2025tirex,
title={TiRex: Zero-Shot Forecasting Across Long and Short Horizons with Enhanced In-Context Learning},
author={Andreas Auer and Patrick Podest and Daniel Klotz and Sebastian B{\"o}ck and G{\"u}nter Klambauer and Sepp Hochreiter},
booktitle=neurips,
year={2025}
}

@inproceedings{
liu2025sundial,
title={Sundial: A Family of Highly Capable Time Series Foundation Models},
author={Yong Liu and Guo Qin and Zhiyuan Shi and Zhi Chen and Caiyin Yang and Xiangdong Huang and Jianmin Wang and Mingsheng Long},
booktitle=icml,
year={2025}
}

@article{Ansari2024bolt,
  author       = {Abdul Fatir Ansari and Caner Turkmen and Oleksandr Shchur and Lorenzo Stella},
  title        = {Fast and accurate zero-shot forecasting with Chronos-Bolt and AutoGluon},
  journal = {\url{https://aws.amazon.com/blogs/machine-learning/fast-and-accurate-zero-shot-forecasting-with-chronos-bolt-and-autogluon}},
  year         = {2024}
}

@inproceedings{woo2024unified,
  title={Unified training of universal time series forecasting transformers},
  author={Woo, Gerald and Liu, Chenghao and Kumar, Akshat and Xiong, Caiming and Savarese, Silvio and Sahoo, Doyen},
  booktitle=icml,
  year={2024}
}

@inproceedings{
chen2025visionts,
title={Vision{TS}: Visual Masked Autoencoders Are Free-Lunch Zero-Shot Time Series Forecasters},
author={Mouxiang Chen and Lefei Shen and Zhuo Li and Xiaoyun Joy Wang and Jianling Sun and Chenghao Liu},
booktitle=icml,
year={2025}
}

@inproceedings{
rasul2023lagllama,
title={Lag-Llama: Towards Foundation Models for Time Series Forecasting},
author={Kashif Rasul and Arjun Ashok and Andrew Robert Williams and Arian Khorasani and George Adamopoulos and Rishika Bhagwatkar and Marin Bilo{\v{s}} and Hena Ghonia and Nadhir Hassen and Anderson Schneider and Sahil Garg and Alexandre Drouin and Nicolas Chapados and Yuriy Nevmyvaka and Irina Rish},
booktitle={NeurIPS Workshop R0-FoMo:Robustness of Few-shot and Zero-shot Learning in Large Foundation Models},
year={2023},
}

@article{UCRArchive2018,
    title = {The UCR Time Series Classification Archive},
    author = {Dau, Hoang Anh and Keogh, Eamonn and Kamgar, Kaveh and Yeh, Chin-Chia Michael and Zhu, Yan 
              and Gharghabi, Shaghayegh and Ratanamahatana, Chotirat Ann and Yanping and Hu, Bing 
              and Begum, Nurjahan and Bagnall, Anthony and Mueen, Abdullah and Batista, Gustavo and Hexagon-ML},
    year = {2018},
    journal = {\url{https://www.cs.ucr.edu/~eamonn/time_series_data_2018}}
}

@article{deng2013time,
  title={A time series forest for classification and feature extraction},
  author={Deng, Houtao and Runger, George and Tuv, Eugene and Vladimir, Martyanov},
  journal={Information Sciences},
  year={2013},
}

@inproceedings{guillaume2022random,
  title={Random dilated shapelet transform: A new approach for time series shapelets},
  author={Guillaume, Antoine and Vrain, Christel and Elloumi, Wael},
  booktitle={International Conference on Pattern Recognition and Artificial Intelligence (ICPRAI)},
  year={2022},
}

@inproceedings{wang2017time,
  title={Time series classification from scratch with deep neural networks: A strong baseline},
  author={Wang, Zhiguang and Yan, Weizhong and Oates, Tim},
  booktitle=ijcnn,
  year={2017},
}

@article{ismail2025look,
  title={Look into the lite in deep learning for time series classification},
  author={Ismail-Fawaz, Ali and Devanne, Maxime and Berretti, Stefano and Weber, Jonathan and Forestier, Germain},
  journal={International Journal of Data Science and Analytics},
  year={2025}
}

@article{ismail2020inceptiontime,
  title={Inceptiontime: Finding alexnet for time series classification},
  author={Ismail Fawaz, Hassan and Lucas, Benjamin and Forestier, Germain and Pelletier, Charlotte and Schmidt, Daniel F and Weber, Jonathan and Webb, Geoffrey I and Idoumghar, Lhassane and Muller, Pierre-Alain and Petitjean, Fran{\c{c}}ois},
  journal={Data Mining and Knowledge Discovery},
  year={2020},
}

@inproceedings{goswami2024moment,
  title={MOMENT: a family of open time-series foundation models},
  author={Goswami, Mononito and Szafer, Konrad and Choudhry, Arjun and Cai, Yifu and Li, Shuo and Dubrawski, Artur},
  booktitle=icml,
  year={2024}
}

@inproceedings{gao2024units,
  title={Units: A unified multi-task time series model},
  author={Gao, Shanghua and Koker, Teddy and Queen, Owen and Hartvigsen, Tom and Tsiligkaridis, Theodoros and Zitnik, Marinka},
  booktitle=neurips,
  year={2024}
}

@article{feofanov2025mantis,
  title={Mantis: Lightweight calibrated foundation model for user-friendly time series classification},
  author={Feofanov, Vasilii and Wen, Songkang and Alonso, Marius and Ilbert, Romain and Guo, Hongbo and Tiomoko, Malik and Pan, Lujia and Zhang, Jianfeng and Redko, Ievgen},
  journal={1st ICML Workshop on Foundation Models for Structured Data},
  year={2025}
}

@inproceedings{xiaoming2025time,
  title={Time-MoE: Billion-Scale Time Series Foundation Models with Mixture of Experts},
  author={Shi, Xiaoming and Wang, Shiyu and Nie, Yuqi and Li, Dianqi and Ye, Zhou and Wen, Qingsong and Jin, Ming},
  booktitle=iclr,
  year={2025}
}

@inproceedings{liu2024timer,
  title={Timer: Generative Pre-trained Transformers Are Large Time Series Models},
  author={Liu, Yong and Zhang, Haoran and Li, Chenyu and Huang, Xiangdong and Wang, Jianmin and Long, Mingsheng},
  booktitle=icml,
  year={2024}
}

@article{cohen2025time,
  title={This Time is Different: An Observability Perspective on Time Series Foundation Models},
  author={Cohen, Ben and Khwaja, Emaad and Doubli, Youssef and Lemaachi, Salahidine and Lettieri, Chris and Masson, Charles and Miccinilli, Hugo and Ram{\'e}, Elise and Ren, Qiqi and Rostamizadeh, Afshin and others},
  journal={arXiv preprint arXiv:2505.14766},
  year={2025}
}

@inproceedings{wen2024abstracted,
  title={Abstracted shapes as tokens-a generalizable and interpretable model for time-series classification},
  author={Wen, Yunshi and Ma, Tengfei and Weng, Lily and Nguyen, Lam and Julius, Anak Agung},
  booktitle=neurips,
  year={2024}
}

@inproceedings{lee2023vector,
  title={Vector quantized time series generation with a bidirectional prior model},
  author={Lee, Daesoo and Malacarne, Sara and Aune, Erlend},
  booktitle=aistats,
  year={2023}
}

@article{tao2025values,
  title={From values to tokens: An llm-driven framework for context-aware time series forecasting via symbolic discretization},
  author={Tao, Xiaoyu and Zhang, Shilong and Cheng, Mingyue and Wang, Daoyu and Pan, Tingyue and Pan, Bokai and Zhang, Changqing and Wang, Shijin},
  journal={arXiv preprint arXiv:2508.09191},
  year={2025}
}

@article{shi2025kronos,
  title={Kronos: A foundation model for the language of financial markets},
  author={Shi, Yu and Fu, Zongliang and Chen, Shuo and Zhao, Bohan and Xu, Wei and Zhang, Changshui and Li, Jian},
  journal={arXiv preprint arXiv:2508.02739},
  year={2025}
}

@article{talukder2024totem,
  title={Totem: Tokenized time series embeddings for general time series analysis},
  author={Talukder, Sabera and Yue, Yisong and Gkioxari, Georgia},
  journal=tmlr,
  year={2024}
}

@inproceedings{
yuan2024diffusionts,
title={Diffusion-{TS}: Interpretable Diffusion for General Time Series Generation},
author={Xinyu Yuan and Yan Qiao},
booktitle=iclr,
year={2024},
}

@inproceedings{yoon2019time,
  title={Time-series generative adversarial networks},
  author={Yoon, Jinsung and Jarrett, Daniel and Van der Schaar, Mihaela},
  booktitle=neurips,
  year={2019}
}

@article{desai2021timevae,
  title={Timevae: A variational auto-encoder for multivariate time series generation},
  author={Desai, Abhyuday and Freeman, Cynthia and Wang, Zuhui and Beaver, Ian},
  journal={arXiv preprint arXiv:2111.08095},
  year={2021}
}

@inproceedings{xu2020cot,
  title={Cot-gan: Generating sequential data via causal optimal transport},
  author={Xu, Tianlin and Wenliang, Li Kevin and Munn, Michael and Acciaio, Beatrice},
  booktitle=neurips,
  year={2020}
}

@inproceedings{
li2023your,
title={Your Diffusion Model is Secretly a Zero-Shot Classifier},
author={Alexander Cong Li and Mihir Prabhudesai and Shivam Duggal and Ellis Langham Brown and Deepak Pathak},
booktitle={ICML 2023 Workshop on Structured Probabilistic Inference {\&} Generative Modeling},
year={2023},
}

@inproceedings{razavi2019generating,
  title={Generating diverse high-fidelity images with vq-vae-2},
  author={Razavi, Ali and Van den Oord, Aaron and Vinyals, Oriol},
  booktitle=neurips,
  year={2019}
}

@inproceedings{tian2024visual,
  title={Visual autoregressive modeling: Scalable image generation via next-scale prediction},
  author={Tian, Keyu and Jiang, Yi and Yuan, Zehuan and Peng, Bingyue and Wang, Liwei},
  booktitle=neurips,
  year={2024}
}

@inproceedings{yu2024image,
  title={An image is worth 32 tokens for reconstruction and generation},
  author={Yu, Qihang and Weber, Mark and Deng, Xueqing and Shen, Xiaohui and Cremers, Daniel and Chen, Liang-Chieh},
  booktitle=neurips,
  year={2024}
}

@article{yu2023language,
  title={Language Model Beats Diffusion--Tokenizer is Key to Visual Generation},
  author={Yu, Lijun and Lezama, Jos{\'e} and Gundavarapu, Nitesh B and Versari, Luca and Sohn, Kihyuk and Minnen, David and Cheng, Yong and Birodkar, Vighnesh and Gupta, Agrim and Gu, Xiuye and others},
  journal={arXiv preprint arXiv:2310.05737},
  year={2023}
}

@inproceedings{lee2022autoregressive,
  title={Autoregressive image generation using residual quantization},
  author={Lee, Doyup and Kim, Chiheon and Kim, Saehoon and Cho, Minsu and Han, Wook-Shin},
  booktitle=cvpr,
  year={2022}
}

@inproceedings{jin2023time,
  title={Time-llm: Time series forecasting by reprogramming large language models},
  author={Jin, Ming and Wang, Shiyu and Ma, Lintao and Chu, Zhixuan and Zhang, James Y and Shi, Xiaoming and Chen, Pin-Yu and Liang, Yuxuan and Li, Yuan-Fang and Pan, Shirui and others},
  booktitle=iclr,
  year={2024}
}

@article{radford2019language,
  title={Language models are unsupervised multitask learners},
  author={Radford, Alec and Wu, Jeffrey and Child, Rewon and Luan, David and Amodei, Dario and Sutskever, Ilya and others},
  journal={OpenAI blog},
  year={2019}
}

@article{touvron2023llama,
  title={Llama 2: Open foundation and fine-tuned chat models},
  author={Touvron, Hugo and Martin, Louis and Stone, Kevin and Albert, Peter and Almahairi, Amjad and Babaei, Yasmine and Bashlykov, Nikolay and Batra, Soumya and Bhargava, Prajjwal and Bhosale, Shruti and others},
  journal={arXiv preprint arXiv:2307.09288},
  year={2023}
}

@article{team2024gemma,
  title={Gemma 2: Improving open language models at a practical size},
  author={Team, Gemma and Riviere, Morgane and Pathak, Shreya and Sessa, Pier Giuseppe and Hardin, Cassidy and Bhupatiraju, Surya and Hussenot, L{\'e}onard and Mesnard, Thomas and Shahriari, Bobak and Ram{\'e}, Alexandre and others},
  journal={arXiv preprint arXiv:2408.00118},
  year={2024}
}

@inproceedings{
xu2024fits,
title={{FITS}: Modeling Time Series with \$10k\$ Parameters},
author={Zhijian Xu and Ailing Zeng and Qiang Xu},
booktitle=iclr,
year={2024}
}

@article{wang2025output,
      title={Output Scaling: YingLong-Delayed Chain of Thought in a Large Pretrained Time Series Forecasting Model}, 
      author={Xue Wang and Tian Zhou and Jinyang Gao and Bolin Ding and Jingren Zhou},
      year={2025},
      journal={arXiv preprint arXiv:2506.11029},
}

@article{sun2025xihe,
  title={Xihe: Scalable Zero-Shot Time Series Learner Via Hierarchical Interleaved Block Attention},
  author={Sun, Yinbo and Fang, Yuchen and Zhu, Zhibo and Li, Jia and Liu, Yu and Deng, Qiwen and Zhou, Jun and Yu, Hang and Lu, Xingyu and Ma, Lintao},
  journal={arXiv preprint arXiv:2510.21795},
  year={2025}
}

@article{raffel2020exploring,
  title={Exploring the limits of transfer learning with a unified text-to-text transformer},
  author={Raffel, Colin and Shazeer, Noam and Roberts, Adam and Lee, Katherine and Narang, Sharan and Matena, Michael and Zhou, Yanqi and Li, Wei and Liu, Peter J},
  journal=jmlr,
  year={2020}
}

@inproceedings{van2016conditional,
  title={Conditional image generation with pixelcnn decoders},
  author={Van den Oord, Aaron and Kalchbrenner, Nal and Espeholt, Lasse and Vinyals, Oriol and Graves, Alex and others},
  booktitle=neurips,
  year={2016}
}

@article{jia2025principles,
  title={From principles to applications: A comprehensive survey of discrete tokenizers in generation, comprehension, recommendation, and information retrieval},
  author={Jia, Jian and Gao, Jingtong and Xue, Ben and Wang, Junhao and Cai, Qingpeng and Chen, Quan and Zhao, Xiangyu and Jiang, Peng and Gai, Kun},
  journal={arXiv preprint arXiv:2502.12448},
  year={2025}
}

@inproceedings{feng2025hdt,
  title={Hdt: Hierarchical discrete transformer for multivariate time series forecasting},
  author={Feng, Shibo and Zhao, Peilin and Liu, Liu and Wu, Pengcheng and Shen, Zhiqi},
  booktitle=aaai,
  year={2025}
}

@inproceedings{zhang2018unreasonable,
  title={The unreasonable effectiveness of deep features as a perceptual metric},
  author={Zhang, Richard and Isola, Phillip and Efros, Alexei A and Shechtman, Eli and Wang, Oliver},
  booktitle=cvpr,
  year={2018}
}

@inproceedings{chen2024sdformer,
  title={Sdformer: Similarity-driven discrete transformer for time series generation},
  author={Chen, Zhicheng and SHIBO, FENG and Zhang, Zhong and Xiao, Xi and Gao, Xingyu and Zhao, Peilin},
  booktitle=neurips,
  year={2024}
}

@inproceedings{zhang2026mmpd,
title={{MMPD}: Diverse Time Series Forecasting via Multi-Mode Patch Diffusion Loss},
author={Yunhao Zhang and Wenyao Hu and Jiale Zheng and Lujia Pan and Junchi Yan},
booktitle=iclr,
year={2026},
}

@inproceedings{li2024autoregressive,
  title={Autoregressive Image Generation without Vector Quantization},
  author={Li, Tianhong and Tian, Yonglong and Li, He and Deng, Mingyang and He, Kaiming},
  booktitle=neurips,
  year={2024}
}
\bibliographystyle{plain}

%%%%%%%%%%%%%%%%%%%%%%%%%%%%%%%%%%%%%%%%%%%%%%%%%%%%%%%%%%%%
\newpage

\appendix

\section{Design Details and Referenced Methods in UniTok} \label{sec:appendix_techniques}

\subsection{Length Mapping Functions}
Accounting for the $4$ special tokens, $2 \times 8$ scale statistic tokens and the progressive-resolution autoencoder, the length mapping function $\phi(t)$ and its inverse $\phi^{-1}(l)$ in Eq.~\ref{eq:incremental} are:
\begin{equation}
\phi(t) =
\begin{cases}
 21 & t=1, \\
 21 + \lceil(t-1)/2\rceil & 1 < t \le 5, \\
 23 + \lceil(t-5)/4\rceil & 5 < t \le 21, \\
 27 + \lceil(t-21)/8\rceil & 21 < t \le 85, \\
 35 + \lceil(t-85)/16\rceil & t > 85, \\
\end{cases}
\quad
\phi^{-1}(l) =
\begin{cases}
 1 & l=21, \\
 1 + 2\times(l-21) & 21 < l \le 23, \\
 5 + 4\times(l-23) & 23 < l \le 27 \\
 21 + 8\times(l-27) & 27 < l \le 35 \\
 85 + 16\times(l-35) & l > 35 \\
\end{cases}
\end{equation}
Lengths that do not lie exactly on the grids are padded to the nearest valid grid point, which results in the ceiling operations in $\phi(t)$. The hierarchical mapping structure arises naturally from the autoencoder's progressive-resolution structure.

\subsection{Finite Scalar Quantization}

Finite Scalar Quantization (FSQ), proposed by \cite{mentzer2024finite}, is designed as a replacement for Vector Quantization (VQ) in VQ-VAE. Unlike VQ, FSQ eliminates explicit codebook lookup operations, leading to more stable training and effectively avoiding codebook collapse.

At a high level, FSQ projects the autoencoder latent representation into a low-dimensional space and independently quantizes each dimension to a fixed set of discrete values, forming an implicit codebook. Let the encoder output be $\bh \in \bbR^{d_{model}}$, FSQ performs:
\begin{equation}
    \begin{split}
        &\bz = \bW^{(down)}\bh, \bW^{(down)} \in \bbR^{d_{FSQ}\times d_{model}} \\
        &\bz^{(pre)} = \lfloor L/2 \rfloor \text{Tanh}(\bz) \\
        &\bz^{(post)} = \text{Round}(\bz^{(pre)}) \\
        &\widetilde{\bh} = \bW^{(up)}\bz^{(post)}, \bW^{(up)} \in \bbR^{d_{model} \times d_{FSQ}}
    \end{split}
\end{equation}
The latent vector $\bh$ is first projected to $\bz$ with $d_{FSQ}$ dimensions, typically with $d_{FSQ} \le 10$. Each dimension of $\bz$ is bounded to a finite range $(-\lfloor L/2 \rfloor, +\lfloor L/2 \rfloor)$, producing $\bz^{(pre)}$. After that, $\bz^{(pre)}$ is rounded to $\bz^{(post)}$, whose elements are integers in $\{-\lfloor L/2 \rfloor, \dots, +\lfloor L/2 \rfloor\}$. Finally, $\bz^{(post)}$ is projected back the original latent dimension.

Since rounding is non-differentiable, gradients are propagated via the straight-through estimator:
\begin{equation}
    \bz^{(post)}=\bz^{(pre)} + \text{StopGrad}(\bz^{(post)} - \bz^{(pre)})
\end{equation}

Each dimension of $\bz^{(\text{post})}$ admits $L$ possible integer values, resulting in an implicit codebook of size $L^{d_{\text{FSQ}}}$. In practice, different dimensions may use different quantization levels $L_i$, yielding a flexible codebook size of $\prod_{i=1}^{d_{FSQ}}L_i$.

\subsection{Threshold for Discrete Wavelet Transformation Coefficients}
Although high-frequency components of the DWT capture fine-grained structures, they are also more susceptible to noise. We use the classical universal thresholding method proposed by \cite{donoho1994ideal} to distinguish salient coefficients. Given coefficients from the highest-frequency $\bd_1$, threshold in Eq.~\ref{eq:dwt} is computed by:
\begin{equation}
\begin{split}
    &\text{MAD} = \text{median}\left(|\bd_1-\text{median}(\bd_1)|\right) \\
    &\alpha = \frac{\text{MAD}}{0.6745}\sqrt{2\log N}
\end{split}
\end{equation}
where $\text{MAD}$ is the median absolute deviation at the finest scale. The constant $0.6745$ ensures consistency with the standard deviation under a Gaussian noise assumption, and $N$ denotes the total number of wavelet coefficients across all scales.

\subsection{Loss Weights Adjustment}
We use the widely adopted adaptive loss weighting strategy for image tokenizers~\cite{esser2021taming} to balance loss terms in Eq.~\ref{eq:loss_all}. Taking the adversarial loss $\mathcal{L}_{adv}$ as an example, its weight is computed by a comparison with the base $L1$ loss:
\begin{equation}
    \lambda_{adv} = \frac{\Vert\nabla_{Dec}(\mathcal{L}_{L1})\Vert_2}{\Vert\nabla_{Dec}(\mathcal{L}_{adv})\Vert_2+\delta}
\end{equation}
where $\nabla_{Dec}(\bigcdot)$ denotes the gradient to the last decoder layer in the autoencoder. $\delta=10^{-6}$ is a small constant for numerical stability. Weights of the other two terms $\lambda_{sp}, \lambda_{hf}$ are computed in the same way.

\section{Details of Pretraining Protocols} \label{sec:appendix_pretrain}
\subsection{Datasets}
We pretrain UniTok-FM on the union of two large-scale time series corpora: \emph{GIFT-Pretrain} and the \emph{Chronos-Dataset}. Below, we describe each dataset and clarify how they are combined while strictly avoiding test-set leakage.

\textbf{1) GIFT-Pretrain}~\cite{aksu2024gift}: This is large-scale corpus released alongside the GIFT-Eval benchmark. A strict split-checking procedure is applied to ensure that no test data from GIFT-Eval appears in the pretraining set, guaranteeing a fully zero-shot evaluation for TSFMs trained on it. The dataset consists of 88 sub-datasets spanning 7 domains and 13 sampling frequencies. The total number of time points in GIFT-Pretrain is 230B. This dataset is publicly available at \url{https://huggingface.co/datasets/Salesforce/GiftEvalPretrain}.

\textbf{2) Chronos-Dataset}~\cite{ansari2024chronos}: This is the dataset for training and evaluation of Chronos. The original dataset contains 67 subsets spanning 8 domains and is publicly available at \url{https://huggingface.co/datasets/autogluon/chronos_datasets}.

These two datasets partially overlap. We carefully construct their union and avoid test-set leakage by adding the following subsets from Chronos-Dataset to GIFT-Pretrain: dominick, ercot, exchange\_rate, mexico\_city\_bikes, training\_corpus(kernel\_synth\_1m, tsmixup\_10m), ushcn\_daily, weatherbench(hourly, daily, weekly).

\subsection{Hyperparameters in UniTok}
The encoder in the UniTok autoencoder has $S=4$ blocks. Each block contains 4 sub-blocks, and each sub-block consists of a 1D causal convolution layer followed by a causal multi-head self-attention layer. For the convolution layers, we use a kernel size of 9, which matches the commonly used $3\times3$ kernels in 2D image convolution. For attention layers, we employ 8 attention heads. The hidden dimensions of the four encoder blocks are set to $[128, 256, 256, 512]$, respectively. The decoder mirrors the encoder architecture in a symmetric manner.

For FSQ, we use quantization levels of $[8, 8, 6, 5]$, corresponding to a codebook size of 1,920 for the normalized series. In addition, we include four special tokens: $\langle\text{SOS}\rangle,\langle\text{EOS}\rangle,\langle\text{SEP}\rangle,\langle\text{PAD}\rangle$ and 16 tokens representing statistic scales. The final codebook size for UniTok is 1,940.

Overall, UniTok contains approximately 113M parameters and supports TS with a maximum length of 2,048.

\subsection{Hyperparameters in UniTok-FM}
We directly implement the Qwen3 LLM backbone using the Hugging Face Transformers library(\url{https://huggingface.co/docs/transformers/model_doc/qwen3}), with hyperparameters specified in Tab.~\ref{tab:qwen}, resulting in a 129M-parameter Qwen3 backbone.

During pretraining, the context window is truncated to a maximum length of 800 tokens. This corresponds approximately to concatenating 5 series of length 2,048 or 8 series of length 1,024.

\begin{table}[ht]
\centering
\caption{\textbf{Hyperparameters of the Qwen3 backbone used in UniTok-FM.} All unspecified hyperparameters follow the default Qwen3 configuration.} \label{tab:qwen}
\resizebox{0.35\textwidth}{!}{
\begin{tabular}{lc}
\bottomrule[1.5pt]
Hyper-parameter & Value \\
\midrule[1pt]
vocab\_size & 1,940 \\
hidden\_size & 1,024 \\
intermediate\_size & 3,072 \\
num\_hidden\_layers & 8 \\
num\_attention\_heads & 16 \\
num\_key\_value\_heads & 8 \\
max\_position\_embeddings & 40,960 \\
tie\_word\_embeddings & TRUE \\
\bottomrule[1.5pt]
\end{tabular}
}
\end{table}

\section{Details of Benchmark Protocols} \label{sec:appendix_benchmark}
\subsection{Zero-Shot \& Prompt-Boosted Forecasting}
\textbf{Datasets} Forecasting performance is evaluated on GIFT-Eval~\cite{aksu2024gift}, which comprises 23 datasets across multiple sampling frequencies and prediction horizons (short-, medium-, and long-term). Each evaluation task follows the format ``Dataset/Frequency/Prediction Term'' (e.g., \texttt{bitbrains\_fast\_storage/5T/long}), resulting in 97 tasks in total.

For each task, multiple target windows are generated using non-overlapping rolling windows. When predicting a given target window, the entire historical series preceding it is provided to the model; this history can be extremely long (e.g., exceeding 100K points for \texttt{Electricity/15T/short}). Evaluated models are allowed to select an arbitrary lookback length based on their own design.

\textbf{Metric Aggregation} After computing metrics for each individual task, GIFT-Eval aggregates results across the 97 tasks using the following procedure:

\textit{1) Normalization by Seasonal-Naive:} For each task $i$, the raw score $s_i$ is normalized by the corresponding score of Seasonal-Naive $s_i^{(season)}$: 
\begin{equation}
    \widetilde{s}_i=\frac{s_i}{s_i^{(season)}}
\end{equation}
This normalization reflects the relative performance of the evaluated model compared to the Seasonal-Naive baseline.

\textit{2) Geometric Mean Aggregation:} The final aggregated score is computed as the geometric mean of the normalized scores across all $N=97$ tasks: 
\begin{equation}
    s_{agg} = (\prod_{i=1}^N\widetilde{s}_i)^{1/N}
\end{equation}

\subsection{Few-Shot Generation}
\textbf{Datasets} We adopt four real-world datasets from Diffusion-TS~\cite{yuan2024diffusionts} (i.e., Stocks, ETTh, Energy, fMRI) for generation evaluation. For each dataset, only the first channel is used. A sliding window of length 336 with a stride of 1 is applied to construct a sample pool. Depending on the data budget, we randomly select 5, 200, or 1,000 samples from this pool for model training or in-context inference. Regardless of the budget, each model is required to generate the same number of samples as the full sample pool.

\textbf{Metrics} We evaluate generation quality using the predictive score and discriminative score.

\textit{Predictive Score:} An RNN predictor is trained on the generated samples, where the first 335 points are used to predict the last point. The trained predictor is then evaluated on real data. To account for scale differences across datasets, we report the $R^2$ (coefficient of determination). Given real–prediction pairs ${(y_i, \widetilde{y}i)}{i=1}^N$, $R^2$ is computed as
\begin{equation}
R^2 = 1 - \frac{\sum_i (y_i - \widetilde{y}_i)^2}{\sum_i (y_i - \bar{y})^2}, \bar{y} = \frac{1}{N} \sum_{i=1}^N y_i
\end{equation}
Higher values indicate better performance. Intuitively, this metric reflects how well the generator captures the underlying dynamics: if the generated data preserves the true dynamics, a predictor trained on them should generalize well to real samples.

\textit{Discriminative Score:} 80\% of both real and generated data are used to construct a training set. Then an RNN discriminator is trained on it to distinguish generated data from real data. The discriminator is then evaluated on the remaining 20\% held-out samples. The discriminative score is defined as
\begin{equation}
\text{Disc} = |\text{Accuracy}-0.5|
\end{equation}
Lower values indicate better performance: a score of 0 means the discriminator cannot distinguish generated samples from real ones, while 0.5 indicates perfect separability. Intuitively, this metric measures the realism of the generated data: an ideal generator produces samples that are indistinguishable from real data.

These two metrics were introduced in 2019~\cite{yoon2019time}, with the original implementation based on TensorFlow 1, which is no longer maintained. As a result, the original training procedure is often unstable and may fail to converge. We therefore reimplement both metrics in PyTorch 2 and adopt several modern practices: 1) using AdamW as the optimizer; 2) applying gradient clipping; 3) extending training to 100K optimization steps while selecting the best checkpoint for evaluation. These modifications stabilize the training process and ensure reliable convergence, leading to a more robust evaluation.

\subsection{Few-Shot Classification}

The UCR Archive~\cite{UCRArchive2018} is publicly available at \url{https://www.cs.ucr.edu/%7Eeamonn/time_series_data_2018/}, and contains 128 univariate datasets, each with predefined training and test splits. We compute the number of training samples per class for each dataset and select those with at most 20 instances per class, forming a subset of 53 datasets referred to as UCR-FewShot. The selected datasets are listed in Tab.~\ref{tab:classify_full}.

For baseline methods, training splits are used to train or fine-tune the models. In contrast, UniTok-FM uses the training samples as in-context prompts. Classification accuracy is evaluated on the test split of each dataset, and the average accuracy across the 53 datasets is reported as the final metric.

\section{Full Results} \label{sec:full_results}

Full forecasting results for each dataset are reported in Tab.~\ref{tab:crps_full} (CRPS), Tab.~\ref{tab:mape_full} (MAPE), and Tab.~\ref{tab:mase_full} (MASE). Full generation results are presented in Tab.~\ref{tab:generation_full}, and full classification results are shown in Tab.~\ref{tab:classify_full}.

\begin{table}[tb!]
\centering
\caption{\textbf{Full forecasting CRPS (lower is better) on GIFT-Eval.} The Task column follows the format “Dataset / Frequency / Prediction Term”. The best-performing model on each dataset is highlighted in bold. Results besides UniTok-FM(ZeroShot) and UniTok-FM(Prompt) are from leaderboard of GIFT-Eval: \url{https://huggingface.co/spaces/Salesforce/GIFT-Eval}.} \label{tab:crps_full}
\resizebox{!}{0.65\textwidth}{
\begin{tabular}{c|cccc|cccc|cccccccc|>{\columncolor{gray!30}}c>{\columncolor{gray!30}}c}
\toprule[1.5pt]
Task & \rotatebox{90}{Naive} & \rotatebox{90}{Seasonal-Naive} & \rotatebox{90}{Auto-Theta} & \rotatebox{90}{Auto-Arima} & \rotatebox{90}{Crossformer} & \rotatebox{90}{DLinear} & \rotatebox{90}{PatchTST} & \rotatebox{90}{iTransformer} & \rotatebox{90}{Chronos-2} & \rotatebox{90}{TiRex} & \rotatebox{90}{Sundial} & \rotatebox{90}{Chronos-Bolt} & \rotatebox{90}{Moirai} & \rotatebox{90}{Chronos} & \rotatebox{90}{VisionTS} & \rotatebox{90}{Lag-Llama} & \rotatebox{90}{UniTok-FM(ZeroShot)} & \rotatebox{90}{UniTok-FM(Prompt)} \\
\midrule[1pt]
bitbrains\_fast\_storage/5T/long & 1.803 & 1.000 & 1.155 & 1.096 & 0.900 & 1.130 & 0.568 & 0.586 & 0.597 & 0.571 & 0.689 & 0.635 & 0.622 & 0.604 & 0.822 & 0.849 & 0.545 & \textbf{0.544} \\
bitbrains\_fast\_storage/5T/medium & 1.711 & 1.000 & 1.210 & 1.060 & 0.851 & 0.807 & 0.536 & 0.590 & 0.520 & 0.533 & 0.607 & 0.631 & 0.553 & 0.671 & 0.749 & 0.784 & \textbf{0.506} & 0.509 \\
bitbrains\_fast\_storage/5T/short & 0.738 & 1.000 & 0.604 & 1.000 & 0.549 & 0.477 & 0.389 & 0.379 & 0.323 & \textbf{0.314} & 0.381 & 0.375 & 0.341 & 0.383 & 0.511 & 0.503 & 0.391 & 0.388 \\
bitbrains\_fast\_storage/H/short & 1.180 & 1.000 & 1.125 & 0.826 & 0.898 & 0.786 & \textbf{0.537} & 0.542 & 0.649 & 0.685 & 0.748 & 0.757 & 0.600 & 0.608 & 1.516 & 0.753 & 0.606 & 0.604 \\
bitbrains\_rnd/5T/long & 2.009 & 1.000 & 1.361 & 1.098 & 0.970 & 1.106 & 0.565 & 0.594 & 0.743 & \textbf{0.537} & 0.609 & 0.643 & 0.566 & 0.893 & 0.826 & 0.665 & 0.585 & 0.630 \\
bitbrains\_rnd/5T/medium & 1.717 & 1.000 & 1.257 & 1.078 & 1.001 & 0.864 & 0.530 & 0.566 & 0.859 & \textbf{0.516} & 0.625 & 0.517 & 0.527 & 0.551 & 0.777 & 0.668 & 0.568 & 0.577 \\
bitbrains\_rnd/5T/short & 0.788 & 1.000 & 0.673 & 0.999 & 0.584 & 0.518 & 0.430 & 0.457 & 0.377 & \textbf{0.367} & 0.393 & 0.398 & 0.405 & 0.460 & 0.590 & 0.581 & 0.414 & 0.414 \\
bitbrains\_rnd/H/short & 1.174 & 1.000 & 1.110 & 0.703 & 0.837 & 0.861 & 0.485 & 0.479 & 0.646 & 0.491 & 0.584 & 0.502 & \textbf{0.467} & 0.535 & 0.828 & 0.669 & 0.613 & 0.616 \\
bizitobs\_application/10S/long & 2.129 & 1.000 & \textbf{0.776} & 21.277 & 1.675 & 1.533 & 1.187 & 1.187 & 0.993 & 1.133 & 1.344 & 2.383 & 2.624 & 2.025 & 1.410 & 1.711 & 1.409 & 1.170 \\
bizitobs\_application/10S/medium & 2.074 & 1.000 & \textbf{0.562} & 0.982 & 1.141 & 1.319 & 1.108 & 1.052 & 0.601 & 0.891 & 1.069 & 2.425 & 2.436 & 2.741 & 1.216 & 1.799 & 1.004 & 0.651 \\
bizitobs\_application/10S/short & 1.097 & 1.000 & 0.283 & 0.999 & 1.030 & 2.273 & 0.631 & 0.591 & \textbf{0.275} & 0.329 & 0.468 & 1.549 & 0.941 & 0.890 & 1.200 & 2.016 & 0.443 & 0.380 \\
bizitobs\_l2c/5T/long & 0.913 & 1.000 & 0.975 & 1.039 & 0.452 & 1.007 & 0.500 & \textbf{0.382} & 0.459 & 0.415 & 0.478 & 1.138 & 0.854 & 1.113 & 0.535 & 1.240 & 0.909 & 0.771 \\
bizitobs\_l2c/5T/medium & 0.690 & 1.000 & 0.797 & 1.018 & 0.632 & 0.942 & 0.638 & 0.611 & 0.475 & 0.482 & \textbf{0.450} & 0.856 & 0.730 & 0.930 & 0.661 & 1.194 & 0.644 & 0.541 \\
bizitobs\_l2c/5T/short & 0.306 & 1.000 & 0.305 & 1.000 & 0.569 & 0.307 & 0.283 & 0.293 & 0.264 & 0.289 & \textbf{0.256} & 0.284 & 0.297 & 0.320 & 0.427 & 0.731 & 0.351 & 0.307 \\
bizitobs\_l2c/H/long & 0.869 & 1.000 & 0.870 & 0.836 & 0.370 & 0.448 & 0.309 & 1.029 & \textbf{0.284} & 0.285 & 0.345 & 0.295 & 0.526 & 0.784 & 0.394 & 0.449 & 0.348 & 0.346 \\
bizitobs\_l2c/H/medium & 0.955 & 1.000 & 0.987 & 0.899 & 0.342 & 0.440 & 0.291 & 0.294 & \textbf{0.261} & 0.279 & 0.306 & 0.281 & 0.761 & 0.877 & 0.334 & 0.425 & 0.295 & 0.305 \\
bizitobs\_l2c/H/short & 0.933 & 1.000 & 0.973 & 1.050 & 0.547 & 0.645 & 0.416 & 0.368 & \textbf{0.339} & 0.407 & 0.428 & 0.363 & 0.946 & 0.900 & 0.633 & 0.596 & 0.382 & 0.395 \\
bizitobs\_service/10S/long & 2.222 & 1.000 & 0.967 & 1.046 & 1.272 & 1.261 & 1.064 & 1.016 & \textbf{0.956} & 0.993 & 1.063 & 2.111 & 2.151 & 1.751 & 1.126 & 1.794 & 1.513 & 1.059 \\
bizitobs\_service/10S/medium & 2.003 & 1.000 & 0.576 & 1.022 & 1.234 & 1.110 & 0.948 & 0.883 & \textbf{0.465} & 0.485 & 0.925 & 2.022 & 1.892 & 1.527 & 1.024 & 1.921 & 1.092 & 0.661 \\
bizitobs\_service/10S/short & 0.915 & 1.000 & 0.318 & 1.000 & 1.083 & 0.790 & 0.630 & 0.590 & \textbf{0.253} & 0.293 & 0.408 & 1.267 & 1.053 & 0.680 & 1.020 & 2.019 & 0.422 & 0.361 \\
car\_parts/M/short & 1.452 & 1.000 & 0.778 & 0.749 & 8.712 & 0.732 & 0.581 & 0.581 & \textbf{0.561} & 0.578 & 0.690 & 0.578 & 0.580 & 0.621 & 0.854 & 0.609 & 0.573 & 0.574 \\
covid\_deaths/D/short & 1.000 & 1.000 & 0.746 & \textbf{0.235} & 55.396 & 0.497 & 0.531 & 0.463 & 0.274 & 0.253 & 1.034 & 0.371 & 0.346 & 0.357 & 0.458 & 2.740 & 0.621 & 0.614 \\
electricity/15T/long & 4.247 & 1.000 & 3.563 & 1.146 & 2.177 & 1.146 & 0.722 & 0.767 & \textbf{0.637} & 0.696 & 0.730 & 0.748 & 1.022 & 0.840 & 0.933 & 2.542 & 0.903 & 0.862 \\
electricity/15T/medium & 3.812 & 1.000 & 2.909 & 1.100 & 2.199 & 1.259 & 0.761 & 0.772 & \textbf{0.633} & 0.696 & 0.729 & 0.734 & 0.940 & 0.844 & 0.922 & 2.412 & 0.859 & 0.851 \\
electricity/15T/short & 1.338 & 1.000 & 0.849 & 1.001 & 20.377 & 1.073 & 0.813 & 0.752 & \textbf{0.475} & 0.551 & 0.507 & 0.496 & 0.728 & 0.561 & 1.019 & 1.501 & 0.605 & 0.602 \\
electricity/D/short & 1.000 & 1.000 & 0.850 & 0.800 & 46.319 & 1.624 & 0.796 & 0.737 & 0.558 & \textbf{0.523} & 0.613 & 0.531 & 0.583 & 0.587 & 0.876 & 0.980 & 0.651 & 0.633 \\
electricity/H/long & 4.354 & 1.000 & 1.955 & 1.238 & 0.939 & 1.323 & 0.678 & 0.636 & 0.572 & 0.612 & 0.606 & 0.637 & \textbf{0.557} & 0.684 & 0.828 & 0.768 & 0.948 & 0.849 \\
electricity/H/medium & 4.390 & 1.000 & 1.994 & 1.225 & 1.696 & 1.617 & 0.636 & 0.672 & \textbf{0.596} & 0.617 & 0.627 & 0.636 & 0.645 & 0.685 & 0.879 & 0.778 & 0.969 & 0.861 \\
electricity/H/short & 2.786 & 1.000 & 1.675 & 1.032 & 25.552 & 1.060 & 0.743 & 0.692 & 0.643 & 0.604 & 0.657 & \textbf{0.602} & 0.711 & 0.609 & 0.884 & 0.724 & 0.961 & 0.963 \\
electricity/W/short & 1.000 & 1.000 & 1.017 & 1.004 & 221.544 & 1.118 & 0.961 & 1.098 & 0.563 & 0.484 & 0.725 & \textbf{0.477} & 0.775 & 0.497 & 1.118 & 1.982 & 1.033 & 1.010 \\
ett1/15T/long & 2.104 & 1.000 & 4.086 & 1.164 & 1.005 & 1.008 & 0.726 & 0.720 & 0.709 & \textbf{0.689} & 0.743 & 0.875 & 0.803 & 1.176 & 0.941 & 1.301 & 0.809 & 0.767 \\
ett1/15T/medium & 1.920 & 1.000 & 3.514 & 1.095 & 1.029 & 1.079 & 0.777 & 0.781 & \textbf{0.725} & 0.738 & 0.809 & 0.874 & 1.008 & 1.179 & 1.151 & 1.397 & 0.838 & 0.767 \\
ett1/15T/short & 1.594 & 1.000 & 1.699 & 0.998 & 1.081 & 0.965 & 0.791 & 0.787 & 0.685 & 0.667 & 0.735 & \textbf{0.654} & 0.800 & 0.820 & 1.040 & 1.717 & 0.702 & 0.690 \\
ett1/D/short & 1.000 & 1.000 & 0.835 & 0.683 & 0.935 & 0.921 & 0.744 & 0.842 & \textbf{0.671} & 0.678 & 0.913 & 0.703 & 0.737 & 0.948 & 0.982 & 0.757 & 0.777 & 0.785 \\
ett1/H/long & 2.095 & 1.000 & 4.118 & 0.913 & 1.210 & 0.771 & 0.630 & 0.630 & 0.584 & \textbf{0.547} & 0.600 & 0.660 & 0.609 & 0.743 & 0.773 & 0.683 & 0.584 & 0.686 \\
ett1/H/medium & 1.935 & 1.000 & 3.794 & 0.883 & 0.846 & 1.046 & 0.628 & 0.651 & 0.599 & \textbf{0.579} & 0.619 & 0.696 & 0.648 & 0.759 & 0.855 & 0.700 & 0.635 & 0.648 \\
ett1/H/short & 1.692 & 1.000 & 2.779 & 0.928 & 1.186 & 1.065 & 0.790 & 0.807 & \textbf{0.710} & 0.732 & 0.791 & 0.753 & 0.820 & 0.807 & 1.003 & 0.907 & 0.849 & 0.844 \\
ett1/W/short & 1.000 & 1.000 & 1.023 & 0.978 & 1.825 & 1.433 & 1.036 & 0.940 & 0.870 & 0.891 & 1.295 & 0.949 & \textbf{0.837} & 1.001 & 1.267 & 1.467 & 1.019 & 1.076 \\
ett2/15T/long & 2.132 & 1.000 & 1.272 & 1.242 & 1.077 & 1.062 & 0.739 & 0.760 & 0.699 & \textbf{0.691} & 0.735 & 0.839 & 1.031 & 1.009 & 1.001 & 0.885 & 0.890 & 0.849 \\
ett2/15T/medium & 1.950 & 1.000 & 1.208 & 1.152 & 1.071 & 1.216 & 0.754 & 0.755 & \textbf{0.704} & 0.716 & 0.771 & 0.889 & 0.878 & 0.983 & 1.241 & 0.923 & 0.816 & 0.789 \\
ett2/15T/short & 1.256 & 1.000 & 0.800 & 1.000 & 1.016 & 1.058 & 0.787 & 0.731 & \textbf{0.645} & 0.690 & 0.712 & 0.691 & 0.805 & 0.742 & 1.058 & 0.901 & 0.706 & 0.701 \\
ett2/D/short & 1.000 & 1.000 & 1.070 & 0.815 & 0.965 & 1.422 & 0.854 & 0.841 & 0.614 & 0.611 & 0.669 & 0.610 & 0.618 & \textbf{0.603} & 0.894 & 1.101 & 0.706 & 0.718 \\
ett2/H/long & 1.680 & 1.000 & 1.616 & 1.308 & 0.794 & 0.794 & 0.625 & 0.601 & 0.505 & 0.549 & 0.562 & 0.562 & 0.529 & 0.654 & 0.779 & 0.694 & 0.525 & \textbf{0.503} \\
ett2/H/medium & 1.583 & 1.000 & 1.525 & 1.316 & 0.773 & 0.891 & 0.671 & 0.634 & 0.586 & 0.575 & 0.610 & 0.616 & \textbf{0.537} & 0.709 & 0.859 & 0.696 & 0.571 & 0.574 \\
ett2/H/short & 1.268 & 1.000 & 1.147 & 1.003 & 1.203 & 0.986 & 0.832 & 0.845 & 0.722 & 0.716 & 0.810 & \textbf{0.713} & 0.807 & 0.803 & 0.986 & 0.871 & 0.794 & 0.791 \\
ett2/W/short & 1.000 & 1.000 & 1.197 & 1.017 & 1.578 & 1.451 & 1.062 & 6.269 & 0.671 & 0.648 & 0.735 & 0.660 & 0.652 & \textbf{0.573} & 1.369 & 1.822 & 0.980 & 0.926 \\
hierarchical\_sales/D/short & 1.000 & 1.000 & 0.557 & 0.423 & 2.730 & 0.470 & 0.340 & 0.344 & 0.333 & \textbf{0.328} & 0.374 & 0.332 & 0.331 & 0.346 & 0.476 & 0.358 & 0.364 & 0.363 \\
hierarchical\_sales/W/short & 1.000 & 1.000 & 0.570 & 0.583 & 19.706 & 0.699 & 0.430 & 0.447 & \textbf{0.409} & 0.418 & 0.469 & 0.424 & 0.429 & 0.441 & 0.612 & 0.553 & 0.472 & 0.471 \\
hospital/M/short & 1.382 & 1.000 & 0.885 & 0.954 & 289.655 & 1.224 & 1.019 & 0.832 & 0.816 & 0.812 & 0.975 & 0.913 & \textbf{0.810} & 0.895 & 1.125 & 2.157 & 0.935 & 0.926 \\
jena\_weather/10T/long & 4.512 & 1.000 & 1.788 & 1.282 & 0.352 & 0.391 & 0.279 & 0.269 & 0.217 & \textbf{0.207} & 0.236 & 0.269 & 0.297 & 0.336 & 0.279 & 0.351 & 0.229 & 0.223 \\
jena\_weather/10T/medium & 4.163 & 1.000 & 1.652 & 1.308 & 0.436 & 0.464 & 0.306 & 0.297 & 0.234 & \textbf{0.229} & 0.256 & 0.270 & 0.320 & 0.359 & 0.319 & 0.377 & 0.256 & 0.247 \\
jena\_weather/10T/short & 1.854 & 1.000 & 0.838 & 0.999 & 8.183 & 0.831 & 0.412 & 0.255 & 0.195 & \textbf{0.173} & 0.199 & 0.210 & 0.345 & 0.281 & 0.391 & 0.358 & 0.203 & 0.199 \\
jena\_weather/D/short & 1.000 & 1.000 & 0.388 & 0.380 & 0.336 & 0.347 & 0.250 & 0.446 & 0.222 & \textbf{0.211} & 0.227 & 0.214 & 0.236 & 0.233 & 0.349 & 0.453 & 0.243 & 0.247 \\
jena\_weather/H/long & 2.722 & 1.000 & 3.078 & 0.549 & 0.274 & 0.332 & 0.181 & 0.181 & \textbf{0.140} & 0.140 & 0.157 & 0.147 & 0.156 & 0.178 & 0.193 & 0.187 & 0.149 & 0.152 \\
jena\_weather/H/medium & 2.705 & 1.000 & 2.427 & 0.616 & 0.242 & 0.272 & 0.200 & 0.176 & \textbf{0.147} & 0.154 & 0.170 & 0.158 & 0.166 & 0.205 & 0.198 & 0.212 & 0.156 & 0.162 \\
jena\_weather/H/short & 1.996 & 1.000 & 1.919 & 0.927 & 0.444 & 0.558 & 0.326 & 0.344 & 0.272 & \textbf{0.265} & 0.322 & 0.274 & 0.284 & 0.300 & 0.374 & 0.326 & 0.293 & 0.289 \\
kdd\_cup\_2018/D/short & 1.000 & 1.000 & 0.680 & 0.583 & 5.159 & 0.715 & 0.595 & 0.595 & \textbf{0.543} & 0.557 & 0.587 & 0.552 & 0.557 & 0.746 & 0.817 & 0.674 & 0.602 & 0.604 \\
kdd\_cup\_2018/H/long & 1.331 & 1.000 & 1.036 & 1.121 & \textbf{0.126} & 0.623 & 0.509 & 0.512 & 0.476 & 0.471 & 0.400 & 0.320 & 0.446 & 0.666 & 0.617 & 0.567 & 0.511 & 0.497 \\
kdd\_cup\_2018/H/medium & 1.317 & 1.000 & 1.042 & 1.122 & \textbf{0.312} & 0.722 & 0.583 & 0.576 & 0.550 & 0.555 & 0.497 & 0.397 & 0.581 & 0.875 & 0.692 & 0.657 & 0.618 & 0.611 \\
kdd\_cup\_2018/H/short & 1.026 & 1.000 & 0.970 & 1.021 & 5.607 & 1.061 & 0.835 & 0.844 & 0.682 & 0.691 & 0.642 & \textbf{0.450} & 0.710 & 0.838 & 1.112 & 0.808 & 0.792 & 0.790 \\
loop\_seattle/5T/long & 2.835 & 1.000 & 1.817 & 1.077 & \textbf{0.168} & 1.030 & 0.750 & 0.730 & 0.625 & 0.663 & 0.664 & 1.015 & 0.407 & 1.125 & 0.944 & 1.002 & 0.716 & 0.620 \\
loop\_seattle/5T/medium & 2.787 & 1.000 & 2.046 & 1.049 & 0.520 & 1.074 & 0.807 & 0.813 & 0.632 & 0.665 & 0.660 & 0.989 & \textbf{0.386} & 1.501 & 1.288 & 1.096 & 0.720 & 0.632 \\
loop\_seattle/5T/short & 1.431 & 1.000 & 1.008 & 1.000 & 5.641 & 1.221 & 0.812 & 0.812 & 0.572 & 0.593 & 0.614 & 0.675 & \textbf{0.565} & 0.861 & 1.076 & 1.021 & 0.646 & 0.656 \\
loop\_seattle/D/short & 1.000 & 1.000 & 0.695 & 0.753 & 3.516 & 0.518 & 0.447 & 0.442 & 0.411 & \textbf{0.408} & 0.457 & 0.424 & 0.428 & 0.432 & 0.544 & 0.816 & 0.459 & 0.458 \\
loop\_seattle/H/long & 2.974 & 1.000 & 2.501 & 1.032 & \textbf{0.056} & 0.469 & 0.366 & 0.368 & 0.317 & 0.321 & 0.386 & 0.406 & 0.425 & 0.437 & 0.471 & 0.472 & 0.370 & 0.365 \\
loop\_seattle/H/medium & 2.824 & 1.000 & 2.406 & 0.950 & \textbf{0.086} & 0.642 & 0.437 & 0.420 & 0.388 & 0.395 & 0.465 & 0.470 & 0.492 & 0.518 & 0.557 & 0.524 & 0.453 & 0.447 \\
loop\_seattle/H/short & 1.747 & 1.000 & 1.583 & 1.036 & 1.497 & 0.947 & 0.733 & 0.716 & \textbf{0.553} & 0.556 & 0.639 & 0.624 & 0.709 & 0.633 & 0.929 & 0.744 & 0.668 & 0.669 \\
m4\_daily/D/short & 1.000 & 1.000 & 0.965 & 0.948 & 300.130 & 1.182 & 0.936 & 0.920 & 0.942 & \textbf{0.862} & 1.094 & 0.865 & 1.646 & 0.907 & 1.363 & 2.215 & 1.072 & 1.088 \\
m4\_hourly/H/short & 3.633 & 1.000 & 1.102 & 0.897 & 112.050 & 1.472 & 1.038 & 1.094 & 0.618 & \textbf{0.536} & 0.612 & 0.674 & 0.591 & 0.649 & 1.049 & 1.646 & 1.302 & 1.326 \\
m4\_monthly/M/short & 1.061 & 1.000 & 0.804 & 0.803 & 42.241 & 1.058 & 0.837 & 0.935 & \textbf{0.737} & 0.750 & 0.954 & 0.769 & 0.769 & 0.853 & 1.140 & 1.731 & 0.873 & 0.873 \\
m4\_quarterly/Q/short & 0.962 & 1.000 & 0.807 & 0.832 & 121.362 & 1.122 & 0.844 & 0.807 & 0.747 & \textbf{0.743} & 0.949 & 0.786 & 0.749 & 0.850 & 1.061 & 2.546 & 0.934 & 0.933 \\
m4\_weekly/W/short & 1.000 & 1.000 & 0.877 & 0.820 & 136.684 & 1.143 & 0.660 & 0.828 & 0.604 & \textbf{0.602} & 0.714 & 0.627 & 0.792 & 0.614 & 0.940 & 1.687 & 0.893 & 0.904 \\
m4\_yearly/A/short & 1.000 & 1.000 & 0.839 & 0.948 & 103.540 & 1.225 & 0.853 & 0.853 & 0.813 & 0.846 & 1.167 & 0.886 & \textbf{0.766} & 0.984 & 1.159 & 2.052 & 1.006 & 1.008 \\
m\_dense/D/short & 1.000 & 1.000 & 0.555 & 0.595 & 13.400 & 0.542 & 0.309 & 0.317 & \textbf{0.255} & 0.299 & 0.294 & 0.304 & 0.458 & 0.331 & 0.415 & 0.556 & 0.344 & 0.345 \\
m\_dense/H/long & 3.443 & 1.000 & 3.413 & 0.644 & \textbf{0.151} & 0.618 & 0.286 & 0.322 & 0.270 & 0.275 & 0.311 & 0.405 & 0.291 & 0.322 & 0.396 & 0.453 & 0.361 & 0.351 \\
m\_dense/H/medium & 3.186 & 1.000 & 3.206 & 0.676 & 0.352 & 0.506 & 0.337 & 0.363 & 0.311 & \textbf{0.308} & 0.340 & 0.417 & 0.326 & 0.360 & 0.480 & 0.465 & 0.406 & 0.399 \\
m\_dense/H/short & 1.960 & 1.000 & 1.998 & 1.023 & 4.585 & 0.779 & 0.630 & 0.546 & 0.467 & 0.467 & 0.485 & \textbf{0.455} & 0.509 & 0.499 & 0.877 & 0.580 & 0.595 & 0.601 \\
restaurant/D/short & 1.000 & 1.000 & 0.486 & 0.535 & 5.317 & 0.502 & 0.387 & 0.384 & \textbf{0.376} & 0.376 & 0.422 & 0.390 & 0.393 & 0.412 & 0.529 & 0.712 & 0.428 & 0.429 \\
saugeen/D/short & 1.000 & 1.000 & 1.144 & 0.964 & 0.790 & 1.048 & 0.697 & 0.715 & 0.597 & 0.611 & 0.648 & \textbf{0.578} & 0.605 & 0.738 & 0.995 & 0.909 & 0.725 & 0.681 \\
saugeen/M/short & 1.725 & 1.000 & 0.838 & 0.732 & \textbf{0.479} & 1.101 & 0.836 & 0.710 & 0.645 & 0.667 & 0.746 & 0.665 & 0.782 & 0.917 & 0.921 & 0.709 & 0.712 & 0.724 \\
saugeen/W/short & 1.000 & 1.000 & 1.000 & 0.748 & 3.460 & 0.917 & 0.659 & 0.635 & \textbf{0.473} & 0.477 & 0.553 & 0.494 & 0.576 & 0.644 & 0.821 & 0.738 & 0.596 & 0.622 \\
solar/10T/long & 2.299 & 1.000 & 9.854 & 1.166 & \textbf{0.046} & 0.868 & 0.503 & 0.638 & 0.430 & 0.476 & 0.541 & 0.657 & 1.340 & 1.110 & 0.834 & 1.871 & 0.465 & 0.461 \\
solar/10T/medium & 2.097 & 1.000 & 8.656 & 1.177 & \textbf{0.412} & 0.843 & 0.543 & 0.838 & 0.443 & 0.510 & 0.569 & 0.666 & 1.270 & 1.047 & 0.688 & 1.803 & 0.489 & 0.492 \\
solar/10T/short & 1.097 & 1.000 & 2.746 & 1.001 & 3.420 & 0.959 & 1.594 & 1.594 & \textbf{0.449} & 0.629 & 0.517 & 0.595 & 0.714 & 0.674 & 0.856 & 0.910 & 0.620 & 0.625 \\
solar/D/short & 1.000 & 1.000 & 0.512 & 0.504 & 3.810 & 0.685 & 0.513 & 1.558 & \textbf{0.477} & 0.488 & 0.579 & 0.514 & 0.528 & 0.583 & 0.767 & 0.912 & 0.591 & 0.592 \\
solar/H/long & 3.209 & 1.000 & 6.791 & 0.563 & \textbf{0.004} & 0.568 & 0.327 & 0.341 & 0.315 & 0.266 & 0.272 & 0.376 & 0.334 & 0.430 & 0.404 & 0.486 & 0.365 & 0.334 \\
solar/H/medium & 3.146 & 1.000 & 6.480 & 0.589 & \textbf{0.042} & 0.584 & 0.364 & 0.396 & 0.384 & 0.301 & 0.327 & 0.390 & 0.350 & 0.376 & 0.442 & 0.538 & 0.372 & 0.356 \\
solar/H/short & 2.247 & 1.000 & 3.938 & 1.061 & \textbf{0.352} & 0.857 & 0.575 & 0.720 & 0.585 & 0.461 & 0.555 & 0.504 & 0.571 & 0.565 & 0.701 & 0.723 & 0.579 & 0.575 \\
solar/W/short & 1.000 & 1.000 & 0.739 & 0.725 & 40.239 & 1.001 & 0.772 & 0.968 & 0.661 & 0.685 & 0.705 & \textbf{0.632} & 1.120 & 0.768 & 1.883 & 2.529 & 0.873 & 0.873 \\
sz\_taxi/15T/long & 5.613 & 1.000 & 1.470 & 0.930 & \textbf{0.285} & 0.925 & 0.657 & 0.659 & 0.470 & 0.463 & 0.517 & 0.581 & 0.488 & 0.619 & 0.638 & 0.514 & 0.489 & 0.487 \\
sz\_taxi/15T/medium & 5.105 & 1.000 & 1.395 & 0.926 & \textbf{0.512} & 0.781 & 0.580 & 0.549 & 0.533 & 0.533 & 0.601 & 0.644 & 0.556 & 0.707 & 0.707 & 0.593 & 0.569 & 0.566 \\
sz\_taxi/15T/short & 2.142 & 1.000 & 0.933 & 1.001 & 6.673 & 0.878 & 0.671 & 0.693 & 0.648 & \textbf{0.647} & 0.723 & 0.654 & 0.690 & 0.764 & 0.946 & 0.737 & 0.711 & 0.709 \\
sz\_taxi/H/short & 2.272 & 1.000 & 1.085 & 0.795 & 9.868 & 0.940 & 0.673 & 0.659 & \textbf{0.633} & 0.633 & 0.720 & 0.636 & 0.669 & 0.697 & 0.851 & 0.819 & 0.718 & 0.722 \\
temperature\_rain/D/short & 1.000 & 1.000 & 0.600 & 0.547 & 4.779 & 0.655 & 0.508 & 0.498 & 0.425 & 0.431 & 0.489 & 0.424 & \textbf{0.422} & 0.481 & 0.616 & 0.517 & 0.496 & 0.496 \\
us\_births/D/short & 1.000 & 1.000 & 0.626 & 0.617 & 0.584 & 0.344 & 0.208 & 0.223 & \textbf{0.138} & 0.192 & 0.181 & 0.217 & 0.223 & 0.185 & 0.297 & 0.580 & 0.258 & 0.252 \\
us\_births/M/short & 2.320 & 1.000 & 1.129 & 0.618 & \textbf{0.115} & 1.914 & 0.993 & 1.076 & 0.744 & 0.739 & 1.682 & 1.156 & 0.886 & 1.046 & 1.183 & 2.883 & 1.435 & 1.372 \\
us\_births/W/short & 1.000 & 1.000 & 0.948 & 0.933 & 4.571 & 1.130 & 0.751 & 1.410 & 0.591 & 0.638 & 0.899 & 0.668 & 0.897 & \textbf{0.575} & 1.073 & 1.404 & 0.791 & 0.817 \\
\midrule[1pt]
Aggregation & 1.591 & 1.000 & 1.244 & 0.912 & 1.637 & 0.846 & 0.587 & 0.620 & \textbf{0.485} & 0.488 & 0.559 & 0.574 & 0.610 & 0.652 & 0.755 & 0.880 & 0.591 & 0.573 \\
\bottomrule[1.5pt]
\end{tabular}
}
\end{table}
\begin{table}[tb!]
\centering
\caption{\textbf{Full forecasting MAPE (lower is better) on GIFT-Eval.} During aggregation, N/A values are replaced with 1.0.} \label{tab:mape_full}
\resizebox{!}{0.65\textwidth}{
\begin{tabular}{c|cccc|cccc|cccccccc|>{\columncolor{gray!30}}c>{\columncolor{gray!30}}c}
\toprule[1.5pt]
Task & \rotatebox{90}{Naive} & \rotatebox{90}{Seasonal-Naive} & \rotatebox{90}{Auto-Theta} & \rotatebox{90}{Auto-Arima} & \rotatebox{90}{Crossformer} & \rotatebox{90}{DLinear} & \rotatebox{90}{PatchTST} & \rotatebox{90}{iTransformer} & \rotatebox{90}{Chronos-2} & \rotatebox{90}{TiRex} & \rotatebox{90}{Sundial} & \rotatebox{90}{Chronos-Bolt} & \rotatebox{90}{Moirai} & \rotatebox{90}{Chronos} & \rotatebox{90}{VisionTS} & \rotatebox{90}{Lag-Llama} & \rotatebox{90}{UniTok-FM(ZeroShot)} & \rotatebox{90}{UniTok-FM(Prompt)} \\
\midrule[1pt]
bitbrains\_fast\_storage/5T/long & 0.278 & 1.000 & 0.579 & 1.000 & 1.144 & 1.343 & 0.410 & 0.579 & 0.585 & 0.716 & 0.927 & 0.415 & 0.814 & \textbf{0.229} & 0.922 & 0.773 & 0.637 & 0.716 \\
bitbrains\_fast\_storage/5T/medium & 0.714 & 1.000 & 0.848 & 1.000 & 0.974 & 0.827 & \textbf{0.548} & 0.693 & 0.598 & 0.741 & 0.941 & 0.659 & 0.735 & 0.574 & 0.943 & 1.018 & 0.559 & 0.663 \\
bitbrains\_fast\_storage/5T/short & 0.410 & 1.000 & 0.518 & 1.000 & 0.366 & 0.457 & 0.364 & 0.303 & 0.336 & \textbf{0.298} & 0.468 & 0.399 & 0.364 & 0.356 & 0.518 & 0.520 & 0.374 & 0.418 \\
bitbrains\_fast\_storage/H/short & 1.199 & 1.000 & 1.322 & 0.859 & 1.875 & 1.169 & 0.682 & \textbf{0.614} & 0.670 & 0.974 & 1.195 & 0.879 & 0.905 & 0.696 & 1.157 & 1.137 & 0.720 & 0.731 \\
bitbrains\_rnd/5T/long & 1.787 & 1.000 & 2.475 & 1.000 & 1.956 & 2.214 & 0.403 & 0.767 & \textbf{0.402} & 0.509 & 0.975 & 0.789 & 0.815 & 1.579 & 1.252 & 0.686 & 0.499 & 0.745 \\
bitbrains\_rnd/5T/medium & 0.209 & 1.000 & 0.916 & 1.000 & 2.647 & 1.309 & 0.183 & 0.465 & 0.310 & \textbf{0.117} & 0.863 & 0.161 & 0.516 & 0.160 & 1.214 & 0.539 & 0.342 & 0.339 \\
bitbrains\_rnd/5T/short & 0.538 & 1.000 & 0.775 & 1.001 & 1.254 & 0.568 & 0.440 & 0.423 & 0.337 & 0.341 & 0.411 & 0.480 & 0.345 & 0.456 & 0.794 & \textbf{0.267} & 0.382 & 0.392 \\
bitbrains\_rnd/H/short & 0.676 & 1.000 & 0.949 & 0.459 & 0.991 & 2.047 & 0.290 & 0.307 & 0.348 & 0.254 & 0.648 & 0.437 & \textbf{0.219} & 0.361 & 0.981 & 0.656 & 0.536 & 0.512 \\
bizitobs\_application/10S/long & 2.057 & 1.000 & \textbf{0.892} & 11586.047 & 2.326 & 1.307 & 1.001 & 1.083 & 0.918 & 1.054 & 1.154 & 3.149 & 4.135 & 2.735 & 1.051 & 2.134 & 1.251 & 0.989 \\
bizitobs\_application/10S/medium & 2.217 & 1.000 & \textbf{0.617} & 1.001 & 2.030 & 1.400 & 1.024 & 1.090 & 0.651 & 0.964 & 1.051 & 3.526 & 4.735 & 3.456 & 1.003 & 2.455 & 1.295 & 0.755 \\
bizitobs\_application/10S/short & 1.598 & 1.000 & 0.484 & 1.001 & 3.112 & 3.259 & 0.948 & 0.608 & \textbf{0.414} & 0.482 & 0.608 & 2.411 & 2.023 & 1.266 & 0.998 & 2.557 & 0.685 & 0.651 \\
bizitobs\_l2c/5T/long & 0.966 & 1.000 & 0.963 & 0.999 & \textbf{0.426} & 0.783 & 0.594 & 0.432 & 0.539 & 0.442 & 0.530 & 0.880 & 0.806 & 0.942 & 0.500 & 1.282 & 0.925 & 0.829 \\
bizitobs\_l2c/5T/medium & 0.667 & 1.000 & 0.729 & 1.000 & 0.442 & 0.640 & 0.599 & 0.552 & 0.472 & 0.451 & \textbf{0.414} & 0.612 & 0.581 & 0.711 & 0.538 & 1.137 & 0.579 & 0.521 \\
bizitobs\_l2c/5T/short & 0.394 & 1.000 & 0.397 & 1.000 & 0.495 & \textbf{0.303} & 0.324 & 0.361 & 0.350 & 0.349 & 0.338 & 0.337 & 0.366 & 0.401 & 0.453 & 0.931 & 0.405 & 0.370 \\
bizitobs\_l2c/H/long & 1.065 & 1.000 & 1.029 & 0.414 & 0.285 & 0.383 & 0.314 & 0.656 & 0.258 & 0.276 & 0.304 & \textbf{0.250} & 0.407 & 0.660 & 0.350 & 0.375 & 0.277 & 0.295 \\
bizitobs\_l2c/H/medium & 0.679 & 1.000 & 0.840 & 0.609 & 0.337 & 0.467 & 0.388 & 0.387 & 0.294 & 0.303 & 0.320 & 0.295 & 0.597 & 0.786 & 0.335 & 0.464 & \textbf{0.282} & 0.288 \\
bizitobs\_l2c/H/short & 0.947 & 1.000 & 0.978 & 0.750 & 0.374 & 0.494 & 0.318 & 0.331 & \textbf{0.275} & 0.355 & 0.323 & 0.290 & 0.440 & 0.701 & 0.406 & 0.526 & 0.298 & 0.294 \\
bizitobs\_service/10S/long & 1.778 & 1.000 & 1.201 & 1.000 & 14.955 & 5.523 & 1.602 & 2.767 & \textbf{0.932} & 1.039 & 1.684 & 12.871 & 13.895 & 2.685 & 1.731 & 1.733 & 1.857 & 1.135 \\
bizitobs\_service/10S/medium & 1.677 & 1.000 & 0.861 & 1.000 & 19.716 & 5.061 & 1.532 & 1.997 & \textbf{0.818} & 0.861 & 1.491 & 10.167 & 9.695 & 8.703 & 1.369 & 2.143 & 1.500 & 1.043 \\
bizitobs\_service/10S/short & 3.522 & 1.000 & 0.731 & 1.000 & 24.879 & 5.546 & 1.250 & 1.420 & \textbf{0.676} & 0.708 & 1.016 & 6.184 & 4.721 & 1.590 & 1.481 & 1.958 & 0.879 & 0.826 \\
car\_parts/M/short & 0.995 & 1.000 & \textbf{0.802} & 0.859 & 0.889 & 0.849 & 1.171 & 1.163 & 0.993 & 0.973 & 0.830 & 0.945 & 1.004 & 0.897 & 0.873 & 1.129 & 0.977 & 0.973 \\
covid\_deaths/D/short & 1.000 & 1.000 & 1.017 & \textbf{0.640} & 2230.185 & 0.744 & 0.995 & 0.737 & 0.655 & 0.661 & 1.431 & 0.858 & 0.721 & 0.889 & 0.942 & 3.413 & 1.065 & 1.060 \\
electricity/15T/long & 2.580 & 1.000 & 1.350 & 1.000 & 31.352 & 1.185 & 0.745 & 0.905 & \textbf{0.605} & 0.750 & 0.724 & 0.698 & 0.990 & 0.865 & 0.915 & 2.112 & 0.844 & 0.840 \\
electricity/15T/medium & 1.875 & 1.000 & 1.073 & 0.999 & 21.368 & 1.161 & 0.748 & 0.836 & \textbf{0.547} & 0.681 & 0.630 & 0.625 & 0.980 & 0.818 & 0.785 & 1.954 & 0.799 & 0.774 \\
electricity/15T/short & 1.169 & 1.000 & 0.714 & 0.999 & 17.331 & 0.936 & 0.850 & 0.771 & \textbf{0.495} & 0.574 & 0.519 & 0.542 & 0.741 & 0.585 & 0.953 & 1.221 & 0.599 & 0.597 \\
electricity/D/short & 1.000 & 1.000 & 1.099 & 1.114 & 17.019 & 1.155 & 0.915 & 0.884 & \textbf{0.405} & 0.764 & 0.598 & 0.613 & 0.773 & 0.838 & 0.831 & 1.255 & 0.920 & 0.922 \\
electricity/H/long & 1.509 & 1.000 & 1.348 & 1.001 & 15.090 & 1.398 & 0.929 & 0.957 & 0.806 & 0.847 & \textbf{0.764} & 0.848 & 0.857 & 0.857 & 1.132 & 0.973 & 0.939 & 0.919 \\
electricity/H/medium & 1.528 & 1.000 & 1.066 & 1.001 & 11.742 & 1.241 & 0.634 & 0.724 & 0.599 & 0.650 & 0.599 & \textbf{0.534} & 0.709 & 0.672 & 0.895 & 0.798 & 0.769 & 0.715 \\
electricity/H/short & 1.436 & 1.000 & 0.980 & 0.999 & 15.144 & 1.022 & 0.586 & 0.675 & \textbf{0.430} & 0.521 & 0.550 & 0.489 & 0.694 & 0.432 & 0.815 & 0.763 & 0.627 & 0.562 \\
electricity/W/short & 1.000 & 1.000 & 1.005 & 0.959 & 24.717 & 0.928 & 0.970 & 0.998 & \textbf{0.436} & 0.487 & 0.873 & 0.858 & 0.956 & 0.868 & 0.935 & 1.319 & 0.974 & 0.986 \\
ett1/15T/long & 1.189 & 1.000 & 1.199 & 1.000 & 0.804 & 0.820 & 0.837 & 0.846 & 0.795 & 0.788 & 0.832 & 0.915 & \textbf{0.719} & 0.860 & 0.829 & 1.060 & 0.782 & 0.787 \\
ett1/15T/medium & 1.221 & 1.000 & 0.977 & 1.000 & 0.808 & 0.831 & \textbf{0.784} & 0.817 & 0.840 & 0.823 & 0.846 & 0.873 & 0.835 & 0.971 & 1.012 & 1.161 & 0.795 & 0.815 \\
ett1/15T/short & 1.472 & 1.000 & 0.800 & 1.000 & 0.996 & 0.747 & 0.820 & 0.905 & 0.775 & \textbf{0.721} & 0.792 & 0.738 & 0.727 & 0.793 & 0.820 & 1.410 & 0.791 & 0.747 \\
ett1/D/short & 1.000 & 1.000 & \textbf{0.813} & 0.895 & 1.118 & 1.075 & 0.964 & 1.764 & 1.151 & 0.999 & 1.178 & 1.333 & 1.239 & 1.058 & 1.798 & 2.060 & 1.711 & 1.579 \\
ett1/H/long & 2.136 & 1.000 & 1.013 & 0.827 & 1.020 & 1.224 & 1.696 & 1.168 & 1.293 & 2.122 & 1.063 & \textbf{0.571} & 1.375 & 1.107 & 1.455 & 1.582 & 1.413 & 1.571 \\
ett1/H/medium & 4.147 & 1.000 & 1.083 & 1.411 & 2.707 & 2.873 & 1.807 & 3.116 & 1.572 & 1.630 & 1.415 & 1.507 & 0.993 & 0.972 & 2.848 & 1.029 & 0.383 & \textbf{0.085} \\
ett1/H/short & 1.622 & 1.000 & 1.153 & 0.912 & 1.049 & 0.816 & 0.782 & 0.810 & 0.767 & 0.804 & 0.821 & 0.811 & 0.847 & \textbf{0.756} & 0.928 & 0.925 & 0.994 & 0.925 \\
ett1/W/short & 1.000 & 1.000 & 1.041 & 1.084 & 1.707 & 1.171 & 1.013 & 1.119 & 1.194 & 1.224 & 1.281 & 1.084 & 0.992 & \textbf{0.976} & 1.301 & 1.558 & 1.079 & 1.144 \\
ett2/15T/long & 1.245 & 1.000 & 1.118 & 1.000 & 1.168 & 1.050 & 0.932 & 0.969 & 0.851 & \textbf{0.846} & 0.885 & 0.887 & 1.280 & 1.081 & 1.044 & 1.104 & 1.070 & 1.048 \\
ett2/15T/medium & 1.117 & 1.000 & 0.988 & 1.000 & 1.054 & 1.103 & 0.866 & 0.903 & \textbf{0.776} & 0.805 & 0.842 & 0.822 & 0.988 & 0.969 & 1.091 & 1.091 & 0.926 & 0.877 \\
ett2/15T/short & 1.062 & 1.000 & 0.808 & 0.997 & 0.893 & 0.847 & 0.795 & 0.782 & \textbf{0.678} & 0.710 & 0.710 & 0.727 & 0.926 & 0.795 & 0.906 & 1.000 & 0.756 & 0.716 \\
ett2/D/short & 1.000 & 1.000 & 1.035 & 0.885 & 1.076 & 1.364 & 1.020 & 0.933 & 0.886 & 0.876 & 0.978 & 0.910 & 0.863 & 0.874 & 1.023 & 0.955 & 0.647 & \textbf{0.528} \\
ett2/H/long & 1.146 & 1.000 & 1.272 & 1.109 & 1.372 & 1.235 & 1.115 & 1.104 & \textbf{0.931} & 1.061 & 1.013 & 0.934 & 1.036 & 0.983 & 1.183 & 1.187 & 1.079 & 1.007 \\
ett2/H/medium & 1.202 & 1.000 & 1.163 & 1.307 & 1.074 & 1.024 & 0.999 & 0.979 & 0.881 & 0.957 & 0.918 & \textbf{0.850} & 0.855 & 0.959 & 1.069 & 1.026 & 0.914 & 0.913 \\
ett2/H/short & 1.248 & 1.000 & 1.183 & 1.127 & 1.231 & 0.899 & 0.961 & 0.996 & \textbf{0.815} & 0.816 & 0.868 & 0.815 & 0.927 & 0.878 & 0.885 & 0.973 & 0.889 & 0.872 \\
ett2/W/short & 1.000 & 1.000 & 1.989 & 1.727 & 2.480 & 2.333 & 1.711 & 8.431 & 1.037 & 0.985 & 1.140 & 0.992 & 1.170 & \textbf{0.941} & 1.915 & 3.012 & 1.773 & 2.039 \\
hierarchical\_sales/D/short & 1.000 & 1.000 & 0.764 & 0.609 & 0.975 & 0.671 & 0.538 & 0.536 & 0.522 & 0.528 & 0.588 & 0.515 & \textbf{0.514} & 0.553 & 0.685 & 0.572 & 0.524 & 0.526 \\
hierarchical\_sales/W/short & 1.000 & 1.000 & 0.813 & 0.744 & 2.673 & 0.927 & 0.603 & 0.614 & \textbf{0.541} & 0.548 & 0.720 & 0.577 & 0.577 & 0.601 & 0.855 & 0.642 & 0.664 & 0.662 \\
hospital/M/short & 1.045 & 1.000 & 0.824 & 0.862 & 43.763 & 0.897 & 0.897 & 0.867 & 0.830 & \textbf{0.821} & 0.912 & 0.863 & 0.867 & 0.884 & 0.957 & 1.567 & 0.864 & 0.867 \\
jena\_weather/10T/long & 1.093 & 1.000 & 5.571 & 1.000 & 1.684 & 1.710 & 1.136 & 1.081 & 1.134 & 1.130 & 1.310 & \textbf{0.972} & 1.477 & 1.130 & 1.037 & 1.477 & 1.572 & 1.456 \\
jena\_weather/10T/medium & 1.055 & 1.000 & 3.025 & 1.000 & 1.357 & 1.502 & 0.866 & 0.885 & 0.844 & 0.939 & 1.085 & \textbf{0.818} & 1.146 & 2.403 & 0.992 & 1.457 & 1.060 & 1.154 \\
jena\_weather/10T/short & 0.476 & 1.000 & 0.776 & 0.999 & 2.553 & 1.035 & 0.442 & 0.550 & 0.412 & 0.380 & 0.438 & \textbf{0.376} & 0.457 & 0.414 & 0.647 & 1.229 & 0.517 & 0.502 \\
jena\_weather/D/short & 1.000 & 1.000 & 0.649 & 0.599 & 0.943 & 0.757 & 0.548 & 0.915 & 0.392 & \textbf{0.328} & 0.443 & 0.367 & 0.364 & 0.383 & 0.378 & 0.547 & 0.378 & 0.381 \\
jena\_weather/H/long & 1.272 & 1.000 & 31.402 & 1.329 & 4.775 & 5.466 & 1.918 & 1.720 & 1.355 & 1.209 & 2.005 & 1.224 & 1.517 & \textbf{0.852} & 1.827 & 2.402 & 1.308 & 1.449 \\
jena\_weather/H/medium & 1.198 & 1.000 & 17.159 & 0.987 & 2.268 & 3.444 & 2.239 & 2.515 & 1.499 & 1.131 & 2.191 & 1.153 & 1.892 & \textbf{0.729} & 1.657 & 2.555 & 1.286 & 1.248 \\
jena\_weather/H/short & 0.965 & 1.000 & 10.837 & 1.075 & 1.868 & 2.793 & 1.313 & 1.700 & 1.524 & 0.921 & 1.544 & 1.127 & 0.987 & \textbf{0.797} & 1.410 & 1.597 & 1.344 & 1.333 \\
kdd\_cup\_2018/D/short & 1.000 & 1.000 & 0.969 & 0.942 & 8.400 & 0.986 & 0.893 & 0.898 & 0.853 & 0.904 & 1.030 & 0.957 & 0.899 & \textbf{0.831} & 1.183 & 0.891 & 0.983 & 0.984 \\
kdd\_cup\_2018/H/long & 0.813 & 1.000 & 0.963 & 0.859 & 8.146 & 1.007 & 0.706 & 0.732 & 0.813 & 0.699 & 0.569 & \textbf{0.518} & 0.755 & 0.881 & 0.903 & 0.733 & 0.818 & 0.835 \\
kdd\_cup\_2018/H/medium & 0.927 & 1.000 & 0.739 & 0.977 & 4.439 & 0.702 & 0.590 & 0.629 & 0.674 & 0.535 & 0.505 & \textbf{0.423} & 0.573 & 0.824 & 0.666 & 0.598 & 0.625 & 0.602 \\
kdd\_cup\_2018/H/short & 0.949 & 1.000 & 0.837 & 0.999 & 9.987 & 0.837 & 0.837 & 0.900 & 0.659 & 0.613 & 0.533 & \textbf{0.343} & 0.580 & 0.781 & 1.062 & 0.653 & 0.728 & 0.759 \\
loop\_seattle/5T/long & 1.483 & 1.000 & 1.098 & 0.998 & 0.664 & 1.050 & 1.068 & 1.081 & 0.836 & 0.802 & 0.764 & 1.212 & \textbf{0.495} & 1.359 & 0.829 & 1.402 & 0.961 & 0.734 \\
loop\_seattle/5T/medium & 1.616 & 1.000 & 1.368 & 0.998 & 2.701 & 1.016 & 1.074 & 1.194 & 0.777 & 0.784 & 0.741 & 1.124 & \textbf{0.421} & 1.408 & 1.395 & 1.476 & 0.910 & 0.720 \\
loop\_seattle/5T/short & 1.062 & 1.000 & 0.899 & 0.999 & 2.205 & 1.122 & 1.037 & 1.068 & 0.740 & 0.765 & 0.715 & 0.828 & \textbf{0.672} & 0.976 & 1.145 & 1.452 & 0.814 & 0.738 \\
loop\_seattle/D/short & 1.000 & 1.000 & 0.801 & 0.851 & 0.821 & 0.499 & 0.516 & 0.506 & 0.491 & \textbf{0.487} & 0.496 & 0.501 & 0.500 & 0.504 & 0.529 & 1.005 & 0.518 & 0.518 \\
loop\_seattle/H/long & 1.387 & 1.000 & 1.010 & 1.282 & \textbf{0.228} & 0.609 & 0.660 & 0.707 & 0.525 & 0.518 & 0.572 & 0.665 & 0.741 & 0.622 & 0.686 & 0.870 & 0.558 & 0.562 \\
loop\_seattle/H/medium & 1.602 & 1.000 & 0.941 & 1.115 & \textbf{0.228} & 0.722 & 0.717 & 0.782 & 0.575 & 0.575 & 0.649 & 0.687 & 0.772 & 0.672 & 0.637 & 0.877 & 0.645 & 0.647 \\
loop\_seattle/H/short & 1.762 & 1.000 & 1.023 & 1.000 & \textbf{0.306} & 0.854 & 0.906 & 1.041 & 0.649 & 0.650 & 0.677 & 0.743 & 0.959 & 0.760 & 0.854 & 0.989 & 0.746 & 0.763 \\
m4\_daily/D/short & 1.000 & 1.000 & 1.029 & 0.941 & 45.523 & 1.100 & 1.015 & 1.053 & 0.894 & \textbf{0.869} & 0.930 & 0.875 & 1.207 & 0.913 & 1.140 & 2.600 & 1.062 & 1.059 \\
m4\_hourly/H/short & 2.416 & 1.000 & 1.428 & 1.038 & 114.655 & 1.659 & 1.070 & 1.127 & 0.630 & 0.551 & 0.689 & 0.579 & 0.619 & \textbf{0.544} & 0.807 & 1.018 & 0.812 & 0.815 \\
m4\_monthly/M/short & 0.989 & 1.000 & \textbf{0.801} & 0.838 & 6.607 & 0.936 & 0.874 & 0.999 & 0.808 & 0.825 & 1.022 & 0.827 & 0.864 & 0.884 & 1.061 & 1.651 & 0.882 & 0.882 \\
m4\_quarterly/Q/short & 0.929 & 1.000 & 0.838 & 0.860 & 9.935 & 0.986 & 0.937 & 0.909 & 0.798 & \textbf{0.776} & 0.910 & 0.831 & 0.817 & 0.810 & 0.937 & 2.375 & 0.937 & 0.937 \\
m4\_weekly/W/short & 1.000 & 1.000 & 0.998 & 0.925 & 22.042 & 0.943 & 0.777 & 1.006 & 0.677 & \textbf{0.673} & 0.840 & 0.701 & 0.849 & 0.684 & 0.919 & 1.668 & 1.133 & 1.133 \\
m4\_yearly/A/short & 1.000 & 1.000 & 0.936 & 1.016 & 6.849 & 1.062 & 1.005 & 1.033 & 0.931 & 0.940 & 1.160 & 0.979 & \textbf{0.930} & 0.976 & 1.090 & 2.137 & 1.063 & 1.067 \\
m\_dense/D/short & 1.000 & 1.000 & 0.845 & 0.884 & 8.450 & 0.679 & 0.517 & 0.509 & \textbf{0.437} & 0.491 & 0.473 & 0.498 & 0.778 & 0.509 & 0.557 & 0.873 & 0.555 & 0.555 \\
m\_dense/H/long & 2.190 & 1.000 & 2.290 & 0.916 & 5.263 & 1.139 & 0.583 & 0.611 & 0.560 & 0.534 & 0.634 & 0.693 & 0.526 & \textbf{0.518} & 0.675 & 0.887 & 0.722 & 0.678 \\
m\_dense/H/medium & 2.181 & 1.000 & 1.378 & 0.901 & 3.776 & 0.781 & 0.532 & 0.564 & 0.517 & \textbf{0.478} & 0.584 & 0.593 & 0.490 & 0.481 & 0.619 & 0.744 & 0.671 & 0.644 \\
m\_dense/H/short & 1.926 & 1.000 & 1.198 & 1.000 & 5.167 & 0.780 & 0.691 & 0.596 & 0.554 & 0.530 & 0.537 & 0.538 & 0.535 & \textbf{0.524} & 0.841 & 0.627 & 0.687 & 0.675 \\
restaurant/D/short & 1.000 & 1.000 & 0.924 & 0.874 & 1.442 & 0.753 & \textbf{0.691} & 0.724 & 0.708 & 0.726 & 0.795 & 0.755 & 0.704 & 0.753 & 0.830 & 0.756 & 0.758 & 0.760 \\
saugeen/D/short & 1.000 & 1.000 & 1.122 & 0.946 & 2.249 & 1.326 & 0.912 & 0.876 & \textbf{0.592} & 0.637 & 0.663 & 0.635 & 0.617 & 0.690 & 1.239 & 1.007 & 0.850 & 0.792 \\
saugeen/M/short & 1.470 & 1.000 & 1.000 & 0.766 & 0.849 & 1.197 & 0.966 & 0.903 & \textbf{0.712} & 0.745 & 0.848 & 0.740 & 0.807 & 0.772 & 0.845 & 0.946 & 0.792 & 0.789 \\
saugeen/W/short & 1.000 & 1.000 & 1.073 & 0.656 & 1.638 & 1.043 & 0.658 & 0.613 & 0.449 & 0.441 & 0.402 & 0.494 & 0.574 & \textbf{0.398} & 0.844 & 0.638 & 0.500 & 0.506 \\
solar/10T/long & 0.433 & 1.000 & 4.099 & 1.000 & N/A & 1.468 & 1.355 & 1.247 & 1.523 & 1.273 & 1.510 & 0.821 & \textbf{0.432} & 1.056 & 1.797 & 1.093 & 1.605 & 1.563 \\
solar/10T/medium & 0.889 & 1.000 & 1.298 & 1.000 & 1.214 & 1.190 & 1.121 & 0.562 & 1.124 & 1.102 & 1.091 & 0.938 & \textbf{0.339} & 1.174 & 1.009 & 0.700 & 1.203 & 1.205 \\
solar/10T/short & 0.603 & 1.000 & 1.008 & 0.999 & 0.840 & 0.892 & 0.366 & 0.352 & 1.125 & 1.395 & 1.070 & 1.231 & 0.770 & 1.471 & 0.997 & \textbf{0.347} & 1.475 & 1.483 \\
solar/D/short & 1.000 & 1.000 & 0.858 & 0.867 & \textbf{0.426} & 0.876 & 0.893 & 1.214 & 0.892 & 0.931 & 0.860 & 0.985 & 0.954 & 1.032 & 1.136 & 1.492 & 0.998 & 0.998 \\
solar/H/long & \textbf{0.188} & 1.000 & 5.291 & 0.816 & N/A & 1.499 & 1.681 & 1.150 & 0.887 & 0.837 & 0.807 & 0.491 & 0.533 & 0.612 & 0.940 & 2.061 & 1.945 & 1.768 \\
solar/H/medium & \textbf{0.191} & 1.000 & 1.827 & 0.847 & N/A & 1.327 & 1.186 & 1.003 & 0.722 & 0.699 & 0.705 & 0.488 & 0.633 & 0.816 & 0.795 & 1.886 & 1.287 & 1.253 \\
solar/H/short & \textbf{0.259} & 1.000 & 1.771 & 1.001 & N/A & 1.593 & 1.182 & 0.853 & 0.984 & 0.911 & 0.963 & 1.147 & 0.747 & 1.233 & 1.184 & 1.821 & 1.158 & 1.234 \\
solar/W/short & 1.000 & 1.000 & 0.774 & 0.758 & 1.869 & 0.758 & 0.739 & 0.875 & 0.667 & 0.681 & \textbf{0.639} & 0.649 & 1.127 & 0.765 & 1.435 & 2.236 & 0.846 & 0.847 \\
sz\_taxi/15T/long & 2.826 & 1.000 & 1.550 & 2.549 & \textbf{0.000} & 1.937 & 1.699 & 1.830 & 1.876 & 1.975 & 2.061 & 2.541 & 2.162 & 2.473 & 1.885 & 2.886 & 2.122 & 2.030 \\
sz\_taxi/15T/medium & 0.528 & 1.000 & 0.355 & 0.655 & \textbf{0.000} & 0.534 & 0.509 & 0.603 & 0.499 & 0.517 & 0.549 & 0.646 & 0.640 & 0.661 & 0.597 & 0.720 & 0.585 & 0.426 \\
sz\_taxi/15T/short & 1.000 & 1.000 & 0.929 & 0.998 & \textbf{0.000} & 1.219 & 0.899 & 0.622 & 0.887 & 0.807 & 0.769 & 0.906 & 0.899 & 0.539 & 0.937 & 0.704 & 0.454 & 0.429 \\
sz\_taxi/H/short & 0.624 & 1.000 & \textbf{0.513} & 0.857 & 2.177 & 0.683 & 0.614 & 0.683 & 0.609 & 0.683 & 0.741 & 0.574 & 0.636 & 0.773 & 0.857 & 0.736 & 0.647 & 0.608 \\
temperature\_rain/D/short & 1.000 & 1.000 & 1.134 & 0.902 & 4.112 & 1.144 & 0.606 & 0.441 & 0.205 & \textbf{0.203} & 0.530 & 0.250 & 0.321 & 0.242 & 0.855 & 0.282 & 0.325 & 0.329 \\
us\_births/D/short & 1.000 & 1.000 & 0.874 & 0.842 & 0.874 & 0.335 & 0.257 & 0.270 & \textbf{0.170} & 0.232 & 0.201 & 0.251 & 0.265 & 0.220 & 0.290 & 0.773 & 0.303 & 0.294 \\
us\_births/M/short & 1.983 & 1.000 & 1.170 & 0.621 & \textbf{0.592} & 1.529 & 1.039 & 1.092 & 0.842 & 0.799 & 1.541 & 1.219 & 0.951 & 1.034 & 0.951 & 2.839 & 1.538 & 1.383 \\
us\_births/W/short & 1.000 & 1.000 & 0.954 & 0.954 & 2.179 & 0.923 & 0.790 & 1.549 & 0.632 & 0.673 & 0.870 & 0.700 & 0.915 & \textbf{0.600} & 0.880 & 1.322 & 0.824 & 0.850 \\
\midrule[1pt]
Aggregation & 1.055 & 1.000 & 1.126 & 1.033 & 1.024 & 1.086 & 0.788 & 0.846 & \textbf{0.666} & 0.677 & 0.777 & 0.775 & 0.825 & 0.802 & 0.925 & 1.115 & 0.798 & 0.761 \\
\bottomrule[1.5pt]
\end{tabular}
}
\end{table}
\begin{table}[tb!]
\centering
\caption{\textbf{Full forecasting MASE (lower is better) on GIFT-Eval.}} \label{tab:mase_full}
\resizebox{!}{0.65\textwidth}{
\begin{tabular}{c|cccc|cccc|cccccccc|>{\columncolor{gray!30}}c>{\columncolor{gray!30}}c}
\toprule[1.5pt]
Task & \rotatebox{90}{Naive} & \rotatebox{90}{Seasonal-Naive} & \rotatebox{90}{Auto-Theta} & \rotatebox{90}{Auto-Arima} & \rotatebox{90}{Crossformer} & \rotatebox{90}{DLinear} & \rotatebox{90}{PatchTST} & \rotatebox{90}{iTransformer} & \rotatebox{90}{Chronos-2} & \rotatebox{90}{TiRex} & \rotatebox{90}{Sundial} & \rotatebox{90}{Chronos-Bolt} & \rotatebox{90}{Moirai} & \rotatebox{90}{Chronos} & \rotatebox{90}{VisionTS} & \rotatebox{90}{Lag-Llama} & \rotatebox{90}{UniTok-FM(ZeroShot)} & \rotatebox{90}{UniTok-FM(Prompt)} \\
\midrule[1pt]
bitbrains\_fast\_storage/5T/long & 1.047 & 1.000 & 1.416 & 1.003 & 2.727 & 3.053 & 1.003 & 1.047 & \textbf{0.766} & 0.805 & 0.890 & 0.834 & 0.851 & 0.889 & 1.109 & 0.949 & 0.936 & 0.926 \\
bitbrains\_fast\_storage/5T/medium & 1.042 & 1.000 & 1.164 & 1.000 & 3.589 & 2.729 & 0.983 & 1.106 & \textbf{0.786} & 0.822 & 0.908 & 0.871 & 0.860 & 0.910 & 1.033 & 1.010 & 0.965 & 0.957 \\
bitbrains\_fast\_storage/5T/short & 0.964 & 1.000 & 1.012 & 1.003 & 4.683 & 1.303 & 0.856 & 0.834 & \textbf{0.578} & 0.608 & 0.651 & 0.662 & 0.697 & 0.748 & 0.959 & 0.860 & 0.892 & 0.971 \\
bitbrains\_fast\_storage/H/short & 1.084 & 1.000 & 1.040 & 1.101 & 3.573 & 2.041 & 1.032 & 1.009 & \textbf{0.717} & 0.785 & 0.886 & 0.824 & 0.909 & 0.855 & 1.248 & 1.177 & 0.935 & 0.931 \\
bitbrains\_rnd/5T/long & 1.139 & 1.000 & 1.174 & 1.000 & 1.582 & 1.814 & 1.062 & 1.057 & \textbf{0.935} & 0.957 & 1.006 & 0.970 & 0.985 & 1.077 & 1.017 & 1.054 & 1.031 & 1.024 \\
bitbrains\_rnd/5T/medium & 1.051 & 1.000 & 1.074 & 0.999 & 2.144 & 1.559 & 1.024 & 1.048 & \textbf{0.960} & 0.971 & 1.004 & 0.979 & 0.997 & 1.013 & 1.017 & 1.044 & 1.023 & 1.021 \\
bitbrains\_rnd/5T/short & 1.043 & 1.000 & 1.050 & 1.000 & 2.263 & 1.334 & 1.005 & 1.025 & \textbf{0.821} & 0.839 & 0.870 & 0.865 & 0.923 & 0.908 & 1.060 & 1.039 & 1.027 & 1.014 \\
bitbrains\_rnd/H/short & 0.973 & 1.000 & \textbf{0.952} & 1.007 & 1.789 & 1.408 & 1.012 & 0.997 & 0.961 & 0.963 & 0.990 & 0.977 & 1.005 & 0.959 & 1.110 & 1.146 & 1.012 & 1.012 \\
bizitobs\_application/10S/long & 2.248 & 1.000 & \textbf{0.914} & 11352.490 & 2.614 & 1.325 & 0.995 & 1.073 & 0.920 & 1.059 & 1.156 & 3.270 & 4.210 & 2.885 & 1.060 & 2.321 & 1.178 & 0.995 \\
bizitobs\_application/10S/medium & 2.422 & 1.000 & \textbf{0.661} & 0.999 & 2.519 & 1.442 & 1.029 & 1.104 & 0.670 & 0.982 & 1.061 & 3.612 & 4.756 & 3.667 & 1.011 & 2.659 & 1.157 & 0.775 \\
bizitobs\_application/10S/short & 1.678 & 1.000 & 0.495 & 0.999 & 3.447 & 3.412 & 0.999 & 0.620 & \textbf{0.448} & 0.537 & 0.637 & 2.468 & 2.373 & 1.342 & 1.003 & 2.851 & 0.693 & 0.675 \\
bizitobs\_l2c/5T/long & 0.858 & 1.000 & 0.853 & 0.997 & \textbf{0.363} & 0.770 & 0.472 & 0.402 & 0.445 & 0.417 & 0.437 & 0.853 & 0.770 & 0.825 & 0.418 & 1.100 & 0.806 & 0.732 \\
bizitobs\_l2c/5T/medium & 0.664 & 1.000 & 0.698 & 0.997 & 0.478 & 0.719 & 0.633 & 0.525 & 0.468 & 0.498 & \textbf{0.426} & 0.706 & 0.686 & 0.757 & 0.519 & 1.131 & 0.598 & 0.548 \\
bizitobs\_l2c/5T/short & 0.287 & 1.000 & 0.296 & 1.000 & 0.437 & \textbf{0.246} & 0.270 & 0.275 & 0.270 & 0.283 & 0.252 & 0.282 & 0.295 & 0.305 & 0.339 & 0.781 & 0.333 & 0.298 \\
bizitobs\_l2c/H/long & 1.001 & 1.000 & 0.989 & 1.080 & 0.415 & 0.522 & 0.433 & 1.325 & 0.401 & 0.410 & 0.466 & \textbf{0.390} & 0.764 & 0.877 & 0.445 & 0.616 & 0.478 & 0.495 \\
bizitobs\_l2c/H/medium & 0.983 & 1.000 & 1.093 & 1.033 & 0.347 & 0.448 & 0.356 & 0.352 & 0.330 & 0.344 & 0.364 & \textbf{0.328} & 0.874 & 0.887 & 0.333 & 0.516 & 0.369 & 0.382 \\
bizitobs\_l2c/H/short & 0.895 & 1.000 & 0.980 & 1.030 & 0.436 & 0.527 & 0.408 & 0.350 & \textbf{0.340} & 0.407 & 0.392 & 0.356 & 0.823 & 0.815 & 0.500 & 0.595 & 0.368 & 0.384 \\
bizitobs\_service/10S/long & 2.823 & 1.000 & 1.185 & 1.002 & 2.662 & 1.675 & 1.236 & 1.265 & \textbf{0.970} & 1.093 & 1.066 & 3.875 & 4.447 & 3.087 & 1.090 & 2.976 & 2.276 & 1.114 \\
bizitobs\_service/10S/medium & 2.807 & 1.000 & 0.803 & 1.000 & 3.188 & 1.689 & 1.128 & 1.249 & \textbf{0.741} & 0.889 & 0.963 & 3.768 & 4.536 & 3.468 & 1.000 & 3.161 & 1.692 & 0.957 \\
bizitobs\_service/10S/short & 1.786 & 1.000 & 0.646 & 1.004 & 3.191 & 1.526 & 1.012 & 1.673 & \textbf{0.585} & 0.674 & 0.685 & 2.706 & 2.799 & 1.534 & 0.988 & 3.159 & 0.884 & 0.765 \\
car\_parts/M/short & 1.009 & 1.000 & 1.024 & 0.797 & 0.941 & 0.830 & \textbf{0.663} & 0.666 & 0.696 & 0.705 & 0.796 & 0.712 & 0.695 & 0.756 & 0.932 & 0.686 & 0.728 & 0.729 \\
covid\_deaths/D/short & 1.000 & 1.000 & 0.968 & \textbf{0.669} & 1166.003 & 0.708 & 0.804 & 0.725 & 0.694 & 0.727 & 1.287 & 0.828 & 0.738 & 0.910 & 0.968 & 2.479 & 1.051 & 1.047 \\
electricity/15T/long & 2.933 & 1.000 & 1.289 & 0.997 & 30.253 & 1.066 & 0.825 & 0.894 & \textbf{0.735} & 0.801 & 0.779 & 0.801 & 1.134 & 0.868 & 0.877 & 2.468 & 0.886 & 0.857 \\
electricity/15T/medium & 3.004 & 1.000 & 1.243 & 0.999 & 19.639 & 1.138 & 0.849 & 0.866 & \textbf{0.699} & 0.771 & 0.742 & 0.749 & 1.156 & 0.860 & 0.829 & 2.369 & 0.848 & 0.827 \\
electricity/15T/short & 1.431 & 1.000 & 0.786 & 1.002 & 15.084 & 0.955 & 0.856 & 0.815 & 0.537 & 0.617 & \textbf{0.521} & 0.545 & 0.897 & 0.612 & 0.897 & 1.399 & 0.648 & 0.646 \\
electricity/D/short & 1.000 & 1.000 & 0.946 & 0.916 & 38.399 & 1.792 & 0.931 & 0.840 & 0.712 & \textbf{0.710} & 0.733 & 0.729 & 0.755 & 0.785 & 0.851 & 1.129 & 0.876 & 0.872 \\
electricity/H/long & 2.635 & 1.000 & 1.345 & 0.997 & 22.505 & 1.450 & 0.912 & 0.873 & 0.782 & 0.810 & \textbf{0.706} & 0.812 & 0.827 & 0.787 & 0.912 & 0.966 & 0.978 & 0.956 \\
electricity/H/medium & 2.937 & 1.000 & 1.278 & 0.998 & 21.329 & 1.623 & 0.833 & 0.891 & 0.768 & 0.782 & \textbf{0.713} & 0.774 & 0.855 & 0.761 & 0.912 & 0.941 & 0.989 & 0.967 \\
electricity/H/short & 2.889 & 1.000 & 1.282 & 1.002 & 22.611 & 0.965 & 0.795 & 0.766 & 0.720 & 0.669 & 0.686 & \textbf{0.643} & 0.803 & 0.664 & 0.825 & 0.762 & 0.944 & 0.939 \\
electricity/W/short & 1.000 & 1.000 & 1.024 & 1.000 & 53.118 & 0.881 & 0.938 & 1.000 & \textbf{0.663} & 0.690 & 0.773 & 0.707 & 0.919 & 0.713 & 0.890 & 1.939 & 1.038 & 1.038 \\
ett1/15T/long & 1.717 & 1.000 & 1.478 & 0.999 & 1.092 & 0.999 & 0.924 & 0.932 & \textbf{0.851} & 0.857 & 0.914 & 0.954 & 0.940 & 1.134 & 0.907 & 1.428 & 0.986 & 0.958 \\
ett1/15T/medium & 1.534 & 1.000 & 1.052 & 1.001 & 1.010 & 1.010 & 0.909 & 0.926 & \textbf{0.830} & 0.854 & 0.898 & 0.893 & 1.044 & 1.111 & 1.069 & 1.444 & 0.936 & 0.889 \\
ett1/15T/short & 2.108 & 1.000 & 0.924 & 1.000 & 1.008 & 0.861 & 0.894 & 0.865 & 0.735 & 0.759 & 0.760 & \textbf{0.728} & 0.883 & 0.857 & 0.895 & 1.627 & 0.784 & 0.788 \\
ett1/D/short & 1.000 & 1.000 & 0.984 & 1.040 & 1.153 & 1.113 & 0.945 & 1.344 & \textbf{0.922} & 0.944 & 1.070 & 0.940 & 0.978 & 1.068 & 1.057 & 1.072 & 1.011 & 1.012 \\
ett1/H/long & 1.537 & 1.000 & 1.697 & 1.116 & 1.434 & 0.987 & 0.994 & 0.967 & 0.934 & \textbf{0.893} & 0.950 & 0.916 & 0.933 & 0.967 & 0.960 & 1.070 & 0.925 & 1.025 \\
ett1/H/medium & 1.297 & 1.000 & 1.174 & 1.001 & 0.982 & 1.059 & 0.887 & 0.918 & 0.812 & \textbf{0.788} & 0.821 & 0.877 & 0.861 & 0.874 & 0.893 & 0.942 & 0.851 & 0.871 \\
ett1/H/short & 1.867 & 1.000 & 1.310 & 1.018 & 1.054 & 0.967 & 0.914 & 0.914 & \textbf{0.805} & 0.836 & 0.849 & 0.847 & 0.906 & 0.859 & 0.870 & 1.010 & 0.935 & 0.938 \\
ett1/W/short & 1.000 & 1.000 & 1.068 & 1.125 & 1.498 & 1.221 & 1.068 & \textbf{0.820} & 0.899 & 0.904 & 1.042 & 0.959 & 0.871 & 0.938 & 0.978 & 1.250 & 1.028 & 1.073 \\
ett2/15T/long & 1.285 & 1.000 & 1.086 & 0.997 & 1.244 & 1.086 & 0.949 & 0.997 & \textbf{0.879} & 0.884 & 0.910 & 0.928 & 1.284 & 1.126 & 1.047 & 1.138 & 1.134 & 1.134 \\
ett2/15T/medium & 1.158 & 1.000 & 0.989 & 0.999 & 1.056 & 1.151 & 0.888 & 0.888 & \textbf{0.807} & 0.833 & 0.863 & 0.877 & 1.046 & 1.008 & 1.103 & 1.095 & 0.967 & 0.935 \\
ett2/15T/short & 1.166 & 1.000 & 0.780 & 1.003 & 0.889 & 0.878 & 0.824 & 0.766 & \textbf{0.680} & 0.706 & 0.700 & 0.718 & 0.899 & 0.803 & 0.892 & 0.962 & 0.732 & 0.722 \\
ett2/D/short & 1.000 & 1.000 & 1.331 & 1.043 & 1.439 & 2.338 & 1.561 & 1.324 & 0.929 & 0.943 & 1.084 & 0.951 & 0.942 & \textbf{0.906} & 1.194 & 1.982 & 1.205 & 1.234 \\
ett2/H/long & 1.172 & 1.000 & 1.294 & 1.134 & 1.365 & 1.400 & 1.267 & 1.179 & 0.929 & 1.050 & 1.010 & \textbf{0.918} & 0.993 & 0.993 & 1.143 & 1.221 & 1.095 & 1.034 \\
ett2/H/medium & 1.153 & 1.000 & 1.050 & 1.179 & 1.082 & 1.098 & 1.025 & 1.009 & 0.880 & 0.916 & 0.900 & \textbf{0.830} & 0.832 & 0.929 & 1.050 & 1.019 & 0.914 & 0.931 \\
ett2/H/short & 1.178 & 1.000 & 1.105 & 1.031 & 1.105 & 0.885 & 0.929 & 0.945 & 0.798 & \textbf{0.784} & 0.835 & 0.794 & 0.874 & 0.846 & 0.864 & 0.934 & 0.874 & 0.873 \\
ett2/W/short & 1.000 & 1.000 & 1.811 & 1.451 & 2.351 & 2.479 & 1.914 & 9.595 & 0.991 & 0.983 & 1.203 & \textbf{0.949} & 1.093 & 0.962 & 1.927 & 2.851 & 1.848 & 1.868 \\
hierarchical\_sales/D/short & 1.000 & 1.000 & 0.821 & 0.716 & 1.269 & 0.758 & 0.666 & 0.676 & 0.656 & 0.657 & 0.696 & \textbf{0.655} & 0.657 & 0.682 & 0.773 & 0.695 & 0.681 & 0.682 \\
hierarchical\_sales/W/short & 1.000 & 1.000 & 0.828 & 0.829 & 2.751 & 0.969 & 0.752 & 0.778 & \textbf{0.692} & 0.704 & 0.733 & 0.715 & 0.729 & 0.745 & 0.841 & 0.906 & 0.766 & 0.764 \\
hospital/M/short & 1.051 & 1.000 & 0.827 & 0.897 & 36.175 & 0.881 & 0.891 & 0.833 & \textbf{0.804} & 0.821 & 0.910 & 0.860 & 0.842 & 0.886 & 0.955 & 1.543 & 0.886 & 0.887 \\
jena\_weather/10T/long & 0.989 & 1.000 & 1.300 & 0.999 & 1.229 & 1.198 & 1.405 & 0.906 & 0.909 & \textbf{0.852} & 0.892 & 0.862 & 1.001 & 1.053 & 0.926 & 0.979 & 1.091 & 1.039 \\
jena\_weather/10T/medium & 1.062 & 1.000 & 1.126 & 1.000 & 1.690 & 1.634 & 1.317 & 0.929 & \textbf{0.846} & \textbf{0.846} & 0.892 & 0.852 & 0.994 & 1.008 & 0.926 & 1.045 & 1.055 & 1.034 \\
jena\_weather/10T/short & 0.490 & 1.000 & 0.495 & 1.000 & 37.691 & 2.638 & 0.743 & 0.501 & \textbf{0.357} & 0.377 & 0.400 & 0.411 & 0.471 & 0.493 & 0.534 & 0.864 & 0.430 & 0.420 \\
jena\_weather/D/short & 1.000 & 1.000 & 1.017 & 0.922 & 1.220 & 1.017 & 0.883 & 1.233 & 0.688 & 0.645 & \textbf{0.592} & 0.668 & 0.731 & 0.712 & 1.023 & 1.360 & 0.719 & 0.726 \\
jena\_weather/H/long & 1.156 & 1.000 & 2.082 & 1.562 & 1.191 & 1.499 & 1.033 & 0.970 & 0.795 & \textbf{0.777} & 0.858 & 0.811 & 0.836 & 0.876 & 0.868 & 0.887 & 0.916 & 0.798 \\
jena\_weather/H/medium & 0.982 & 1.000 & 1.530 & 1.632 & 1.328 & 1.122 & 1.227 & 1.084 & 0.884 & 0.853 & 0.977 & \textbf{0.841} & 0.919 & 0.994 & 0.946 & 1.168 & 0.995 & 0.947 \\
jena\_weather/H/short & 0.896 & 1.000 & 1.215 & 1.494 & 1.130 & 1.358 & 0.887 & 0.988 & \textbf{0.723} & 0.724 & 0.745 & 0.741 & 0.766 & 0.784 & 0.865 & 0.964 & 0.821 & 0.820 \\
kdd\_cup\_2018/D/short & 1.000 & 1.000 & 0.922 & 0.788 & 8.016 & 0.822 & 0.815 & 0.815 & 0.800 & 0.802 & \textbf{0.784} & 0.799 & 0.802 & 0.915 & 0.982 & 0.924 & 0.827 & 0.829 \\
kdd\_cup\_2018/H/long & 0.884 & 1.000 & 1.026 & 0.884 & 4.830 & 0.816 & 0.764 & 0.771 & 0.784 & 0.756 & 0.580 & \textbf{0.512} & 0.719 & 0.854 & 0.779 & 0.839 & 0.841 & 0.814 \\
kdd\_cup\_2018/H/medium & 1.012 & 1.000 & 0.931 & 0.994 & 8.817 & 0.784 & 0.735 & 0.735 & 0.728 & 0.725 & 0.588 & \textbf{0.490} & 0.735 & 0.868 & 0.742 & 0.807 & 0.823 & 0.803 \\
kdd\_cup\_2018/H/short & 0.955 & 1.000 & 0.947 & 1.000 & 12.160 & 0.843 & 0.836 & 0.850 & 0.706 & 0.701 & 0.597 & \textbf{0.448} & 0.704 & 0.776 & 0.895 & 0.776 & 0.792 & 0.787 \\
loop\_seattle/5T/long & 1.126 & 1.000 & 1.151 & 0.999 & 0.872 & 0.935 & 0.848 & 0.824 & 0.729 & 0.761 & 0.714 & 0.990 & \textbf{0.473} & 1.047 & 0.864 & 1.032 & 0.793 & 0.696 \\
loop\_seattle/5T/medium & 1.723 & 1.000 & 1.786 & 0.997 & 1.752 & 0.971 & 0.910 & 0.893 & 0.733 & 0.763 & 0.711 & 0.985 & \textbf{0.453} & 1.396 & 1.145 & 1.133 & 0.801 & 0.710 \\
loop\_seattle/5T/short & 1.171 & 1.000 & 1.023 & 1.000 & 2.099 & 1.174 & 0.976 & 0.964 & 0.709 & 0.735 & 0.711 & 0.823 & \textbf{0.703} & 1.002 & 1.036 & 1.169 & 0.784 & 0.800 \\
loop\_seattle/D/short & 1.000 & 1.000 & 0.802 & 0.860 & 0.935 & 0.520 & 0.539 & 0.537 & 0.512 & \textbf{0.505} & 0.519 & 0.521 & 0.521 & 0.526 & 0.548 & 0.927 & 0.539 & 0.539 \\
loop\_seattle/H/long & 1.133 & 1.000 & 1.307 & 1.675 & \textbf{0.488} & 0.666 & 0.633 & 0.647 & 0.565 & 0.565 & 0.638 & 0.644 & 0.744 & 0.653 & 0.660 & 0.785 & 0.635 & 0.628 \\
loop\_seattle/H/medium & 1.200 & 1.000 & 1.087 & 1.351 & \textbf{0.450} & 0.811 & 0.696 & 0.675 & 0.613 & 0.620 & 0.685 & 0.688 & 0.770 & 0.709 & 0.709 & 0.784 & 0.698 & 0.691 \\
loop\_seattle/H/short & 1.231 & 1.000 & 1.083 & 0.998 & \textbf{0.570} & 0.874 & 0.828 & 0.820 & 0.638 & 0.638 & 0.681 & 0.696 & 0.820 & 0.716 & 0.859 & 0.827 & 0.745 & 0.748 \\
m4\_daily/D/short & 1.000 & 1.000 & 1.019 & 0.994 & 60.395 & 1.043 & 0.982 & 0.994 & 1.043 & 0.986 & 1.133 & 0.976 & 1.638 & \textbf{0.970} & 1.223 & 2.933 & 1.151 & 1.164 \\
m4\_hourly/H/short & 9.728 & 1.000 & 2.062 & 0.863 & 137.444 & 1.416 & 1.173 & 1.215 & 0.661 & 0.605 & 0.729 & 0.701 & 0.814 & \textbf{0.581} & 0.775 & 2.940 & 1.501 & 1.504 \\
m4\_monthly/M/short & 0.957 & 1.000 & 0.767 & 0.775 & 16.115 & 0.897 & 0.841 & 1.278 & \textbf{0.717} & 0.731 & 0.873 & 0.753 & 0.757 & 0.772 & 0.921 & 2.482 & 0.870 & 0.870 \\
m4\_quarterly/Q/short & 0.922 & 1.000 & 0.743 & 0.799 & 15.853 & 0.911 & 0.824 & 0.768 & 0.720 & 0.715 & 0.910 & 0.764 & \textbf{0.712} & 0.768 & 0.849 & 3.724 & 0.891 & 0.891 \\
m4\_weekly/W/short & 1.000 & 1.000 & 0.958 & 0.850 & 37.807 & 1.671 & 0.843 & 0.929 & 0.730 & \textbf{0.716} & 0.863 & 0.748 & 1.012 & 0.749 & 0.911 & 3.151 & 1.219 & 1.222 \\
m4\_yearly/A/short & 1.000 & 1.000 & 0.784 & 0.935 & 7.917 & 1.049 & 0.830 & 0.850 & 0.816 & 0.848 & 1.092 & 0.884 & \textbf{0.759} & 0.918 & 0.966 & 2.412 & 0.992 & 0.993 \\
m\_dense/D/short & 1.000 & 1.000 & 0.731 & 0.803 & 8.446 & 0.605 & 0.438 & 0.436 & \textbf{0.368} & 0.405 & 0.408 & 0.429 & 0.659 & 0.427 & 0.472 & 0.771 & 0.466 & 0.465 \\
m\_dense/H/long & 1.932 & 1.000 & 1.549 & 0.819 & 2.957 & 0.839 & 0.499 & 0.564 & \textbf{0.462} & 0.470 & 0.522 & 0.635 & 0.497 & 0.523 & 0.556 & 0.722 & 0.591 & 0.574 \\
m\_dense/H/medium & 1.742 & 1.000 & 1.108 & 0.809 & 2.471 & 0.592 & 0.482 & 0.533 & \textbf{0.444} & 0.445 & 0.483 & 0.561 & 0.468 & 0.482 & 0.568 & 0.634 & 0.553 & 0.546 \\
m\_dense/H/short & 1.824 & 1.000 & 1.136 & 1.002 & 3.146 & 0.699 & 0.692 & 0.601 & 0.526 & 0.526 & 0.532 & \textbf{0.521} & 0.563 & 0.538 & 0.766 & 0.629 & 0.622 & 0.627 \\
restaurant/D/short & 1.000 & 1.000 & 0.838 & 0.923 & 1.968 & 0.702 & 0.686 & 0.684 & \textbf{0.674} & \textbf{0.674} & 0.700 & 0.696 & 0.700 & 0.724 & 0.739 & 1.164 & 0.732 & 0.732 \\
saugeen/D/short & 1.000 & 1.000 & 1.055 & 1.096 & 1.392 & 1.231 & 0.961 & 0.982 & 0.874 & 0.879 & \textbf{0.815} & 0.832 & 0.853 & 0.964 & 1.169 & 1.224 & 1.009 & 0.952 \\
saugeen/M/short & 1.260 & 1.000 & 0.934 & 0.743 & \textbf{0.729} & 0.978 & 0.915 & 0.803 & 0.741 & 0.767 & 0.771 & 0.757 & 0.854 & 0.875 & 0.819 & 0.791 & 0.774 & 0.763 \\
saugeen/W/short & 1.000 & 1.000 & 1.065 & 0.779 & 1.371 & 0.909 & 0.779 & 0.748 & 0.597 & \textbf{0.588} & 0.602 & 0.611 & 0.708 & 0.678 & 0.814 & 0.823 & 0.702 & 0.710 \\
solar/10T/long & 2.317 & 1.000 & 5.201 & 1.000 & \textbf{0.458} & 1.355 & 1.047 & 1.332 & 0.912 & 0.954 & 1.090 & 1.229 & 2.319 & 1.906 & 1.309 & 3.267 & 0.979 & 0.995 \\
solar/10T/medium & 2.766 & 1.000 & 2.902 & 1.000 & 2.179 & 1.165 & 0.985 & 1.489 & \textbf{0.852} & 0.924 & 1.015 & 1.108 & 2.039 & 1.661 & 0.960 & 2.814 & 0.911 & 0.920 \\
solar/10T/short & 1.311 & 1.000 & 1.628 & 1.004 & 1.547 & 1.121 & 1.990 & 2.053 & \textbf{0.704} & 0.984 & 0.757 & 0.896 & 0.995 & 1.004 & 1.004 & 1.276 & 0.900 & 0.909 \\
solar/D/short & 1.000 & 1.000 & 0.908 & 0.874 & 1.306 & 0.891 & 0.832 & 2.708 & 0.831 & \textbf{0.827} & 0.935 & 0.849 & 0.882 & 0.874 & 1.004 & 1.504 & 0.989 & 0.992 \\
solar/H/long & 2.057 & 1.000 & 4.892 & 0.929 & \textbf{0.066} & 1.260 & 0.913 & 1.008 & 0.929 & 0.771 & 0.700 & 0.964 & 0.999 & 0.999 & 0.892 & 1.342 & 0.998 & 0.942 \\
solar/H/medium & 2.268 & 1.000 & 3.070 & 0.907 & \textbf{0.394} & 1.251 & 1.032 & 1.144 & 1.144 & 0.838 & 0.820 & 0.996 & 0.954 & 0.862 & 0.951 & 1.424 & 1.002 & 0.980 \\
solar/H/short & 2.199 & 1.000 & 2.154 & 1.000 & \textbf{0.188} & 1.114 & 1.002 & 1.187 & 1.034 & 0.782 & 0.827 & 0.854 & 0.938 & 0.869 & 0.915 & 1.143 & 0.976 & 0.960 \\
solar/W/short & 1.000 & 1.000 & 0.782 & 0.762 & 2.299 & 0.769 & 0.748 & 0.918 & 0.689 & 0.707 & 0.667 & \textbf{0.666} & 1.129 & 0.782 & 1.455 & 2.308 & 0.864 & 0.867 \\
sz\_taxi/15T/long & 1.074 & 1.000 & 1.098 & 0.865 & 2.591 & 1.217 & 1.101 & 1.133 & 0.738 & \textbf{0.737} & 0.777 & 0.789 & 0.777 & 0.821 & 0.805 & 0.824 & 0.759 & 0.756 \\
sz\_taxi/15T/medium & 1.092 & 1.000 & 1.004 & 0.886 & 2.719 & 0.882 & 0.824 & 0.768 & \textbf{0.749} & 0.753 & 0.789 & 0.784 & 0.782 & 0.837 & 0.786 & 0.829 & 0.775 & 0.773 \\
sz\_taxi/15T/short & 1.028 & 1.000 & 0.849 & 0.999 & 2.420 & 0.761 & 0.733 & 0.750 & \textbf{0.710} & \textbf{0.710} & 0.724 & 0.717 & 0.754 & 0.771 & 0.819 & 0.798 & 0.754 & 0.752 \\
sz\_taxi/H/short & 1.131 & 1.000 & 0.936 & 0.845 & 4.186 & 0.904 & 0.801 & 0.780 & \textbf{0.759} & 0.763 & 0.788 & 0.762 & 0.797 & 0.780 & 0.805 & 0.966 & 0.828 & 0.831 \\
temperature\_rain/D/short & 1.000 & 1.000 & 0.959 & 0.850 & 13.124 & 0.910 & 0.751 & 0.716 & 0.651 & 0.660 & 0.711 & \textbf{0.648} & 0.651 & 0.701 & 0.795 & 0.749 & 0.720 & 0.720 \\
us\_births/D/short & 1.000 & 1.000 & 0.874 & 0.847 & 0.885 & 0.346 & 0.261 & 0.280 & \textbf{0.175} & 0.237 & 0.208 & 0.260 & 0.273 & 0.225 & 0.299 & 0.761 & 0.308 & 0.303 \\
us\_births/M/short & 1.972 & 1.000 & 1.161 & 0.613 & \textbf{0.559} & 1.538 & 1.028 & 1.080 & 0.832 & 0.791 & 1.523 & 1.215 & 0.951 & 1.023 & 0.956 & 2.877 & 1.512 & 1.360 \\
us\_births/W/short & 1.000 & 1.000 & 0.953 & 0.947 & 2.156 & 0.934 & 0.787 & 1.516 & 0.626 & 0.669 & 0.871 & 0.696 & 0.921 & \textbf{0.596} & 0.889 & 1.331 & 0.820 & 0.845 \\
\midrule[1pt]
Aggregation & 1.270 & 1.000 & 1.090 & 1.074 & 2.574 & 1.061 & 0.849 & 0.893 & \textbf{0.698} & 0.716 & 0.750 & 0.808 & 0.901 & 0.876 & 0.863 & 1.228 & 0.851 & 0.824 \\
\bottomrule[1.5pt]
\end{tabular}
}
\end{table}

\begin{table}[tb!]
\centering
\caption{\textbf{Full generation performance on Stocks, ETTh, Energy and fMRI.} \#Train denotes the number of training examples for baseline models, while \#Prompt denotes the number of in-context prompt examples for UniTok-FM. The best-performing model on each dataset is highlighted in bold.} \label{tab:generation_full}
\resizebox{0.98\textwidth}{!}{
\begin{tabular}{c|ccc|ccc|ccc|ccc|ccc|>{\columncolor{gray!30}}c}
\toprule[1.5pt]
Methods & \multicolumn{3}{c|}{Diffusion-TS} & \multicolumn{3}{c|}{SDformer} & \multicolumn{3}{c|}{TimeGAN} & \multicolumn{3}{c|}{TimeVAE} & \multicolumn{3}{c|}{Cot-GAN} & UniTok-FM \\
\midrule[1pt]
\#Train / \#Prompt & 5 & 200 & 1000 & 5 & 200 & 1000 & 5 & 200 & 1000 & 5 & 200 & 1000 & 5 & 200 & 1000 & 5 \\
\midrule[1pt]
\multicolumn{17}{c}{Predictive Score$\uparrow$} \\
\midrule[1pt]
Stocks & 0.796 & 0.997 & 0.995 & 0.820 & 0.998 & \textbf{0.999} & 0.524 & 0.988 & 0.952 & 0.728 & 0.997 & 0.997 & 0.962 & 0.998 & 0.997 & 0.998 \\
ETTh & 0.347 & 0.822 & 0.874 & 0.229 & 0.786 & 0.867 & 0.298 & -0.700 & -0.622 & 0.626 & 0.866 & \textbf{0.891} & 0.775 & 0.700 & 0.824 & 0.870 \\
Energy & 0.162 & 0.451 & 0.466 & 0.166 & 0.123 & 0.342 & 0.294 & 0.502 & -0.077 & 0.341 & 0.515 & \textbf{0.548} & 0.323 & 0.517 & 0.539 & 0.432 \\
fMRI & -0.134 & \textbf{0.114} & -0.007 & -0.256 & -0.558 & 0.086 & -1.389 & -0.709 & -0.051 & -1.068 & -0.003 & 0.012 & 0.020 & 0.078 & 0.049 & 0.103 \\
\midrule[1pt]
Average & 0.293 & 0.596 & 0.582 & 0.239 & 0.337 & 0.573 & -0.068 & 0.020 & 0.051 & 0.157 & 0.594 & \textbf{0.612} & 0.520 & 0.573 & 0.602 & 0.601 \\
\midrule[1pt]
\multicolumn{17}{c}{Discriminative Score$\downarrow$}\\
\midrule[1pt]
Stocks & 0.493 & 0.181 & 0.243 & 0.471 & 0.040 & \textbf{0.021} & 0.500 & 0.500 & 0.500 & 0.457 & 0.103 & 0.256 & 0.376 & 0.207 & 0.447 & 0.260 \\
ETTh & 0.499 & 0.201 & 0.260 & 0.493 & 0.144 & \textbf{0.070} & 0.500 & 0.499 & 0.500 & 0.481 & 0.363 & 0.439 & 0.473 & 0.500 & 0.499 & 0.467 \\
Energy & 0.497 & 0.366 & 0.457 & 0.486 & 0.176 & \textbf{0.083} & 0.500 & 0.500 & 0.500 & 0.495 & 0.494 & 0.498 & 0.494 & 0.499 & 0.500 & 0.455 \\
fMRI & 0.499 & 0.496 & 0.445 & 0.499 & 0.228 & \textbf{0.212} & 0.500 & 0.500 & 0.500 & 0.497 & 0.500 & 0.500 & 0.461 & 0.390 & 0.497 & 0.499 \\
\midrule[1pt]
Average & 0.497 & 0.311 & 0.351 & 0.487 & 0.147 & \textbf{0.096} & 0.500 & 0.500 & 0.500 & 0.483 & 0.365 & 0.423 & 0.451 & 0.399 & 0.486 & 0.420 \\
\bottomrule[1.5pt]
\end{tabular}
}
\end{table}

\begin{table}[tb!]
\centering
\caption{\textbf{Full classification accuracy on UCR-FewShot.} \#Train per Class denotes the number of training samples per class. The best-performing model on each dataset is highlighted in bold. Results of Manits, UniTS and Moment are from \cite{feofanov2025mantis}. ``/'' indicates out-of-memory errors encountered during UniTS fine-tuning due to long sequence lengths, for which results are unavailable.} \label{tab:classify_full}
\resizebox{0.98\textwidth}{!}{
\begin{tabular}{c|c|ccc|ccc|c|cc|>{\columncolor{gray!30}}c}
\toprule[1.5pt]
Dataset & \#Train per Class & KNN & TStree & RDST & FCN & LITE & Inception & Mantis & UniTS & Moment & UniTok-FM \\
\midrule[1pt]
Fungi & 1 & 0.839 & 0.737 & 0.570 & 0.199 & 0.167 & 0.081 & 0.778 & 0.627 & \textbf{0.996} & 0.715 \\
PigAirwayPressure & 2 & 0.091 & 0.072 & 0.072 & 0.111 & 0.149 & 0.077 & 0.484 & / & 0.119 & \textbf{0.764} \\
PigArtPressure & 2 & 0.288 & 0.106 & 0.471 & 0.038 & 0.067 & 0.058 & 0.910 & / & 0.611 & \textbf{0.966} \\
PigCVP & 2 & 0.139 & 0.072 & 0.365 & 0.072 & 0.130 & 0.144 & 0.784 & / & 0.609 & \textbf{0.904} \\
DiatomSizeReduction & 4 & 0.935 & 0.964 & 0.859 & 0.301 & 0.307 & 0.105 & \textbf{0.968} & 0.912 & 0.887 & 0.863 \\
Symbols & 4.2 & 0.899 & 0.835 & 0.859 & 0.165 & 0.174 & 0.338 & \textbf{0.989} & 0.821 & 0.938 & 0.928 \\
PickupGestureWiimoteZ & 5 & 0.300 & 0.500 & 0.480 & 0.160 & 0.260 & 0.060 & \textbf{0.760} & 0.560 & 0.680 & 0.520 \\
Rock & 5 & 0.640 & 0.580 & 0.580 & 0.240 & 0.340 & 0.300 & 0.713 & 0.613 & \textbf{0.840} & 0.740 \\
ShakeGestureWiimoteZ & 5 & 0.420 & 0.580 & 0.720 & 0.220 & 0.500 & 0.100 & \textbf{0.920} & 0.607 & 0.853 & 0.820 \\
Phoneme & 5.5 & 0.109 & 0.073 & 0.203 & 0.249 & 0.276 & 0.297 & \textbf{0.321} & 0.143 & 0.275 & 0.201 \\
InsectEPGSmallTrain & 5.7 & \textbf{1.000} & \textbf{1.000} & 0.920 & 0.526 & 0.831 & \textbf{1.000} & \textbf{1.000} & 0.667 & 0.850 & \textbf{1.000} \\
Beef & 6 & 0.667 & 0.600 & 0.667 & 0.200 & 0.333 & 0.333 & 0.700 & 0.667 & \textbf{0.744} & 0.633 \\
FaceFour & 6 & 0.784 & 0.830 & 0.784 & 0.455 & 0.159 & 0.500 & \textbf{0.977} & 0.652 & 0.777 & 0.841 \\
Mallat & 6.9 & 0.914 & 0.889 & 0.716 & 0.125 & 0.371 & 0.261 & \textbf{0.940} & 0.874 & 0.859 & 0.812 \\
OliveOil & 7.5 & 0.867 & \textbf{0.900} & 0.867 & 0.400 & 0.400 & 0.300 & 0.889 & 0.478 & \textbf{0.900} & 0.667 \\
GestureMidAirD1 & 8 & 0.292 & 0.354 & 0.538 & 0.146 & 0.308 & 0.323 & \textbf{0.769} & 0.526 & 0.674 & 0.338 \\
GestureMidAirD2 & 8 & 0.223 & 0.354 & 0.385 & 0.069 & 0.269 & 0.100 & \textbf{0.674} & 0.405 & 0.574 & 0.346 \\
GestureMidAirD3 & 8 & 0.108 & 0.162 & 0.285 & 0.046 & 0.069 & 0.031 & \textbf{0.408} & 0.295 & 0.359 & 0.138 \\
FiftyWords & 9 & 0.631 & 0.415 & 0.404 & 0.299 & 0.549 & 0.767 & \textbf{0.814} & 0.623 & 0.678 & 0.653 \\
ACSF1 & 10 & 0.540 & 0.610 & 0.560 & 0.620 & 0.440 & 0.660 & 0.743 & 0.693 & \textbf{0.750} & 0.440 \\
BME & 10 & 0.827 & 0.987 & 0.753 & 0.333 & 0.433 & 0.333 & \textbf{0.996} & 0.887 & 0.976 & 0.847 \\
BeetleFly & 10 & 0.750 & 0.800 & 0.750 & 0.500 & 0.500 & 0.500 & 0.883 & 0.717 & \textbf{0.950} & 0.700 \\
BirdChicken & 10 & 0.550 & 0.850 & 0.650 & 0.500 & 0.500 & 0.500 & \textbf{0.900} & 0.617 & 0.850 & 0.850 \\
CBF & 10 & 0.852 & 0.800 & 0.767 & 0.426 & 0.589 & \textbf{0.993} & 0.985 & 0.851 & 0.941 & 0.960 \\
Chinatown & 10 & 0.945 & 0.977 & 0.971 & 0.927 & \textbf{0.985} & 0.974 & 0.972 & 0.978 & 0.981 & 0.974 \\
CinCECGTorso & 10 & \textbf{0.897} & 0.554 & 0.810 & 0.248 & 0.348 & 0.253 & 0.781 & 0.585 & 0.700 & 0.819 \\
DodgerLoopGame & 10 & \textbf{0.899} & 0.630 & 0.681 & 0.486 & 0.486 & 0.486 & 0.884 & 0.831 & 0.812 & 0.826 \\
DodgerLoopWeekend & 10 & \textbf{0.986} & 0.899 & 0.899 & 0.746 & 0.746 & 0.746 & 0.978 & 0.964 & 0.961 & 0.949 \\
Lightning7 & 10 & 0.575 & 0.575 & 0.575 & 0.329 & 0.699 & \textbf{0.808} & 0.740 & 0.557 & 0.676 & 0.712 \\
MoteStrain & 10 & 0.879 & 0.797 & 0.859 & 0.466 & 0.539 & 0.788 & 0.907 & 0.834 & 0.891 & \textbf{0.924} \\
ShapeletSim & 10 & 0.539 & 0.472 & 0.561 & 0.500 & 0.500 & 0.772 & \textbf{1.000} & 0.698 & 0.965 & 0.750 \\
ShapesAll & 10 & 0.752 & 0.568 & 0.515 & 0.172 & 0.725 & 0.827 & \textbf{0.891} & 0.670 & 0.824 & 0.860 \\
SonyAIBORobotSurface1 & 10 & 0.696 & 0.794 & 0.800 & 0.429 & 0.429 & 0.429 & 0.814 & 0.721 & 0.809 & \textbf{0.835} \\
Adiac & 10.5 & 0.611 & 0.501 & 0.430 & 0.256 & 0.414 & 0.327 & 0.777 & 0.577 & \textbf{0.789} & 0.629 \\
WordSynonyms & 10.7 & 0.618 & 0.368 & 0.448 & 0.332 & 0.478 & 0.621 & \textbf{0.743} & 0.523 & 0.600 & 0.558 \\
DodgerLoopDay & 11.1 & 0.563 & 0.463 & 0.388 & 0.238 & 0.400 & 0.413 & 0.625 & 0.450 & 0.454 & \textbf{0.663} \\
ECGFiveDays & 11.5 & 0.797 & 0.834 & 0.988 & 0.503 & \textbf{1.000} & 0.999 & 0.915 & 0.846 & 0.856 & 0.930 \\
TwoLeadECG & 11.5 & 0.747 & 0.741 & 0.963 & 0.500 & 0.500 & 0.665 & \textbf{0.994} & 0.755 & 0.965 & 0.873 \\
ArrowHead & 12 & 0.800 & 0.646 & 0.640 & 0.303 & 0.423 & 0.303 & 0.821 & 0.690 & 0.808 & \textbf{0.834} \\
UMD & 12 & 0.806 & 0.486 & 0.694 & 0.160 & 0.458 & 0.333 & \textbf{0.991} & 0.806 & 0.970 & 0.938 \\
SonyAIBORobotSurface2 & 13.5 & 0.859 & 0.723 & 0.804 & 0.652 & 0.617 & \textbf{0.927} & 0.900 & 0.829 & 0.835 & 0.880 \\
Coffee & 14 & \textbf{1.000} & 0.893 & 0.964 & 0.536 & 0.536 & 0.536 & \textbf{1.000} & \textbf{1.000} & 0.893 & 0.857 \\
FreezerSmallTrain & 14 & 0.676 & 0.845 & 0.952 & 0.500 & 0.500 & 0.492 & \textbf{0.967} & 0.671 & 0.776 & 0.854 \\
FacesUCR & 14.3 & 0.769 & 0.552 & 0.640 & 0.789 & 0.939 & \textbf{0.969} & 0.915 & 0.720 & 0.795 & 0.811 \\
InlineSkate & 14.3 & 0.342 & 0.280 & 0.235 & 0.173 & 0.165 & 0.184 & \textbf{0.363} & / & 0.318 & 0.296 \\
Car & 15 & 0.733 & 0.550 & 0.817 & 0.233 & 0.367 & 0.283 & \textbf{0.872} & 0.628 & 0.794 & 0.683 \\
Plane & 15 & 0.962 & 0.981 & 0.990 & 0.448 & \textbf{1.000} & \textbf{1.000} & \textbf{1.000} & 0.952 & 0.984 & \textbf{1.000} \\
ToeSegmentation2 & 18 & 0.808 & 0.592 & 0.846 & 0.185 & 0.185 & 0.315 & 0.923 & 0.797 & 0.846 & \textbf{0.931} \\
HouseTwenty & 20 & 0.681 & 0.697 & 0.899 & 0.832 & 0.941 & 0.849 & \textbf{0.980} & 0.835 & 0.936 & 0.882 \\
InsectWingbeatSound & 20 & 0.562 & 0.519 & 0.479 & 0.317 & 0.615 & 0.602 & 0.596 & 0.573 & \textbf{0.616} & 0.604 \\
Meat & 20 & 0.933 & 0.917 & 0.917 & 0.333 & 0.333 & 0.333 & 0.933 & 0.911 & \textbf{0.944} & 0.750 \\
MixedShapesSmallTrain & 20 & 0.835 & 0.725 & 0.897 & 0.415 & 0.842 & 0.708 & \textbf{0.953} & 0.728 & 0.839 & 0.839 \\
ToeSegmentation1 & 20 & 0.680 & 0.640 & 0.768 & 0.526 & 0.711 & 0.917 & \textbf{0.966} & 0.801 & 0.924 & 0.934 \\
\midrule[1pt]
Average & 10.2 & 0.672 & 0.628 & 0.673 & 0.357 & 0.472 & 0.491 & \textbf{0.840} & 0.697 & 0.778 & 0.755 \\
\#Win & N/A & 5 & 2 & 0 & 0 & 3 & 6 & \textbf{27} & 1 & 9 & 10 \\
\bottomrule[1.5pt]
\end{tabular}
}
\end{table}

\section{Broader Impacts} \label{sec:social}
UniTok and UniTok-FM advance time series modeling by enabling unified, efficient processing via next-token prediction, supporting training-free, in-context analysis across domains such as finance and energy. However, its generative capabilities could be misused to create fake information or manipulate temporal data. Furthermore, incorrect results in areas like healthcare or power grid management could lead to systemic failures.  

%%%%%%%%%%%%%%%%%%%%%%%%%%%%%%%%%%%%%%%%%%%%%%%%%%%%%%%%%%%%

\end{document}